\documentclass[runningheads]{llncs}
\usepackage{graphicx}
\usepackage[breaklinks,colorlinks]{hyperref}
\hypersetup{
    colorlinks=true, 
    linkcolor=blue,
    filecolor=blue,      
    urlcolor=blue,
    citecolor=blue,
}
\usepackage{gensymb} 
\usepackage{amsmath} 
\usepackage{amssymb} 
\usepackage[capitalise]{cleveref} 
\usepackage{algpseudocode}
\usepackage{algorithm} 
\usepackage{booktabs}
\usepackage{multirow}
\usepackage{float} 
\usepackage{caption}
\usepackage{subcaption}
\usepackage[misc]{ifsym}
\newcommand{\sectionBefore}{\vspace{-2ex}} 
\newcommand{\sectionAfter}{\vspace{-1.2ex}} 
\newcommand{\subsectionBefore}{\vspace{-1.0ex}} 
\newcommand{\subsectionAfter}{\vspace{-0.8ex}} 
\newcommand{\subsubsectionBefore}{\vspace{-2.5ex}} 
\algrenewcommand\algorithmicrequire{\textbf{Input:}}
\algrenewcommand\algorithmicensure{\textbf{Output:}}
\begin{document}
\title{JVLDLoc: a Joint Optimization of 
\\ Visual-LiDAR Constraints and
Direction Priors for Localization in Driving Scenario}
\titlerunning{JVLDLoc}
%
\author{Longrui Dong  \and 
Gang Zeng\textsuperscript{(\Letter)}  }
\authorrunning{L. Dong and G. Zeng}
%
\institute{Key Lab. of Machine Perception (MoE), \\
School of Intelligence Science and Technology, \\
Beijing 100871, China \\
\email{\{lrdong,zeng\}@pku.edu.cn}}
\maketitle              
\begin{abstract}

The ability for a moving agent to localize itself in environment is the basic demand for emerging applications, such as autonomous driving, etc.
Many existing methods based on multiple sensors still suffer from drift.
We propose a scheme that fuses map prior and vanishing points from images, which can establish an energy term that is only constrained on rotation, called the direction projection error.  
Then we embed these direction priors into a visual-LiDAR SLAM system that integrates camera and LiDAR measurements in a tightly-coupled way at backend.
Specifically, our method generates visual reprojection error and point to Implicit Moving Least Square(IMLS) surface of scan constraints, and solves them jointly along with direction projection error at global optimization.
Experiments on KITTI, KITTI-360 and Oxford Radar Robotcar show that we achieve lower localization error or Absolute Pose Error (APE) than prior map, which validates our method is effective. 

\keywords{Visual-LiDAR SLAM  \and Sensor Fusion \and Structure from Motion \and Vanishing Point}
\end{abstract}
\section{Introduction}\label{sec:intro}
\sectionAfter
In the fields of computer vision and robotics, Simultaneous localization and mapping (SLAM) is an active research topic, it also plays a vital role in many real world applications.
For example, by knowing their precise location, vehicles are able to navigate safely.
To fully perceive surrounding scenes, agents are usually equipped with several sensors: camera, LiDAR, GPS/IMU, etc. 
As we all know, relatively cheaper cameras have been widely set on various kinds of platforms, which makes visual odometry(VO)/visual SLAM(V-SLAM) the main force in research\cite{campos2021orb,liu2020visual}.
Due to the characteristics of camera and algorithms' reliance on low-level visual features, V-SLAM tends to fail in challenge conditions: abrupt motion, textureless region, too much occlusion.
Light Detection And Ranging sensor(LiDAR), which can cover 360\degree FoV information, has an advantage over camera that the noise associated with each distance measurement is independent of the distance and the lighting conditions\cite{deschaud2018imls}. 
But LiDAR does not offer texture and is often sparse due to the physical spacing between laser fibers. 

In order to take advantage of those two types of sensors, fusion of visual and LiDAR data during localization is quite reasonable, which is called visual-LiDAR odometry/SLAM.
Some recent work\cite{Grter2018LIMOLV,Huang2020LidarMonocularVO,Shin2020DVLSLAMSD} have explored the way of fusing those two sensors. 
However, they nearly all fuse visual-LiDAR at a level of loosely coupling.
\cite{shan2020lio,zhao2021super} use pose graph to fuse different sensors. 
Although they are very efficient, rich geometric constraints from data are thrown away in optimization.

\begin{figure}[t]
	\centering

	 \includegraphics[width=0.60\textwidth]{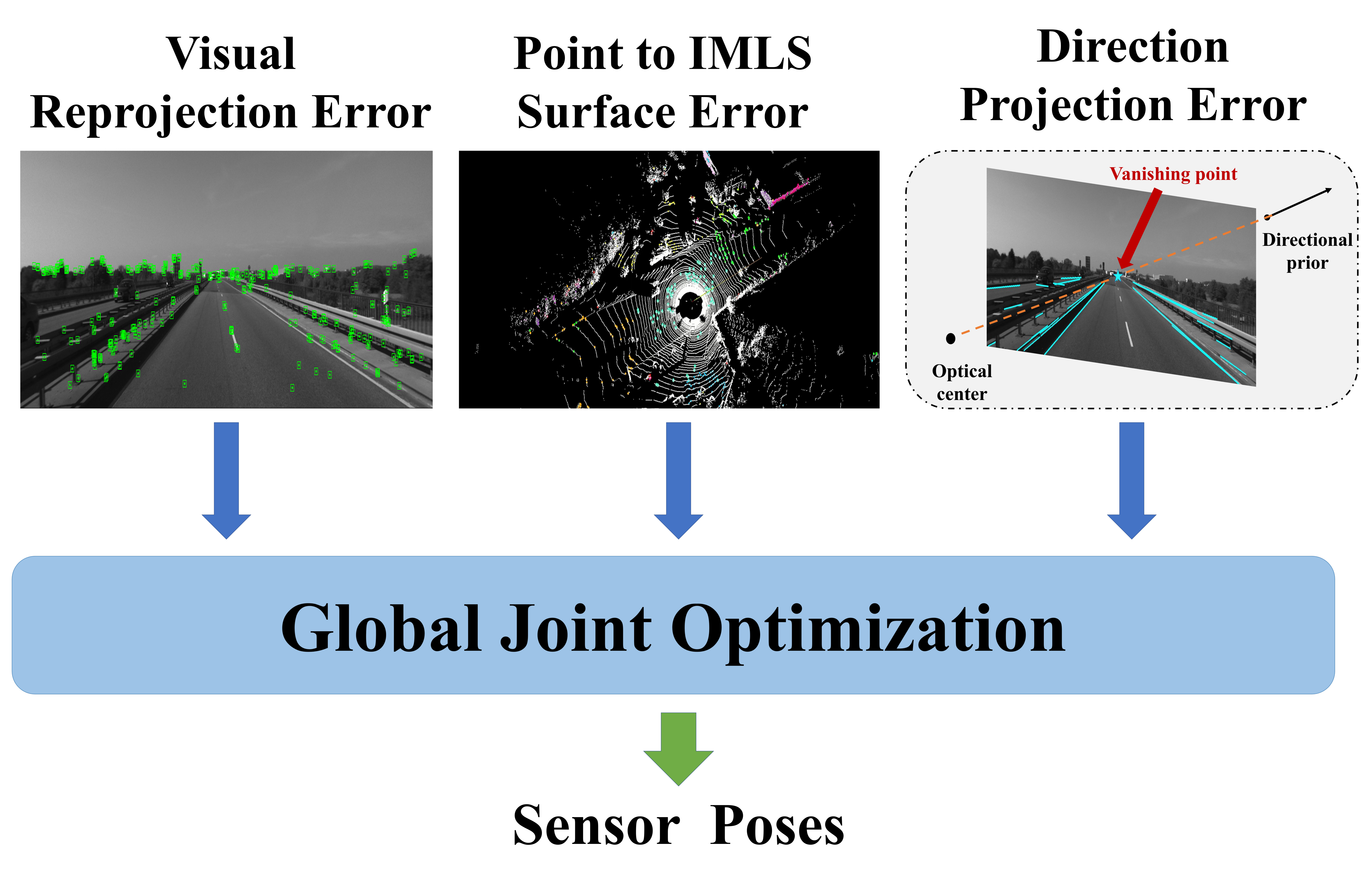} 
	 \caption{\textbf{JVLDLoc}: a tight-coupling of visual reprojection error, point to Implicit Moving Least Square(IMLS) surface error 
	 and direction projection error at global joint optimization.
	 }
	 \label{fig:teaser}
\end{figure}
Unlike \cite{zhao2021super} which just takes odometry as relative pose factor between adjacent keyframes, we use input trajectory to construct direction priors. 
They contain scene structure information from images' vanishing points. 
Since the direction is independent of distance, for the vehicle, as long as there is no road turn, the direction is always in view.
As a result, we can establish longer range correspondences among much more keyframes than traditional visual point features\cite{mur2017orb}.
Ideally, the motion of a vehicle on the road plane can be approximated as a two-dimensional movement, in which the rotating part is almost only a twist around an axis perpendicular to the ground. 
A more accurate rotation estimate can reduce the overall drift of the trajectory\cite{sturm2012benchmark}.
Motivated by them, we extract at most one vanishing point each image, whose corresponding direction is forward.
Actually this direction provides only two degrees of freedom constraints to the vehicle's orientation\cite{andrew2001multiple}.
The main contributions are as follows:

\vspace{-6pt}

\begin{itemize}
	\item We come up with a scheme that fuses map prior and image-detected vanishing points, which can establish an energy term that only constrains rotation.
	That constraint, called the direction projection error, finally helps to mitigate the drift.
	
    \item We propose JVLDLoc for localization in driving scenario. 
    Based on graph optimization, it integrates visual reprojection error, point-to-IMLS surface error and direction projection error into one energy function,
    and directly uses geometric constraints to optimize pose.
	
\end{itemize} 
\vspace{-2pt}
Experiments on KITTI, KITTI-360 and Oxford Radar Robotcar show that we achieve lower localization error or APE than prior map, which validates the effectiveness of our method.

\sectionBefore
\section{Related Work}
\label{sec:relaw}
\vspace{-0.8ex} 
\subsection{Visual-LiDAR SLAM}
\subsectionAfter

First we briefly review some of the major visual or LiDAR SLAM in literature.
ORB-SLAM\cite{campos2021orb,mur2017orb} completely uses ORB features, realizing a full V-SLAM system including loop detection, relocalization, and map fusion.
IMLS-SLAM\cite{deschaud2018imls} estimates pose by minimizing the distance from the current point to the surface represented by the implicit moving least square method (IMLS).
Then we categorize visual-LiDAR SLAM into two major genres according to whether or not to use both multi-modal residuals in the optimization stage.
\subsubsectionBefore
\subsubsection{Loosely Coupled Method}
Methods of this type usually extract depth from LiDAR to enhance visual odometry.
DEMO\cite{zhang2017real} integrates the depth information of point clouds into visual SLAM for the first time.
LIMO-PL\cite{Huang2020LidarMonocularVO} utilizes line features with depths from LiDAR.
\cite{yu2020monocular} uses 3D lines from prior LiDAR map to associate with 2D line features from images.
DVL-SLAM\cite{Shin2020DVLSLAMSD} requires no feature points when using sparse depth from point cloud.
However, incorrect pixel-depth pair might damage accuracy.
\subsubsectionBefore
\subsubsection{Tightly Coupled Method}
ViLiVO\cite{Xiang2019ViLiVOVL} adds the image feature reprojection error and the point cloud matching error to Bundle adjustment(BA).
Similiar methods appear in \cite{Zhen2020LiDARenhancedS} but they further introduce observation error between image feature and point cloud.
TVL-SLAM\cite{chou2021efficient} also incorporates visual and lidar measurements in backend optimization. 
There are also filter based methods for tightly fusion different data\cite{lin2021r2live}.
LIO-SAM\cite{shan2020lio} and LVI-SAM\cite{shan2021lvi} achieve sensor-fusion atop the pose graph instead of including any gemometric constraints.

\subsectionBefore
\subsection{Using Vanishing Points as Direction Constraint} 
\subsectionAfter
The common directions of parallel 3-D lines are known as vanishing points (VPs).
Because the VPs offer direction information independent of the present robot proposition\cite{bosse2003vanishing}, 
Many academics have researched VPs to help improving orientation. 
\cite{liu2020visual} apply the orthogonal constraint between multiple VPs, known as the Manhattan world constraint,
to calcute the full rotation. 
\cite{lim2022uv} on longer need Manhattan world assumption but they will extract all valid VPs from an image while our method indeed only use one VP.
\cite{wang2021vanishing} also combine VP error in a pose estimator, but it have to generate VPs of three directions before refinement. 

\begin{figure}[t]
	\centering
	\includegraphics[width=0.9\textwidth]{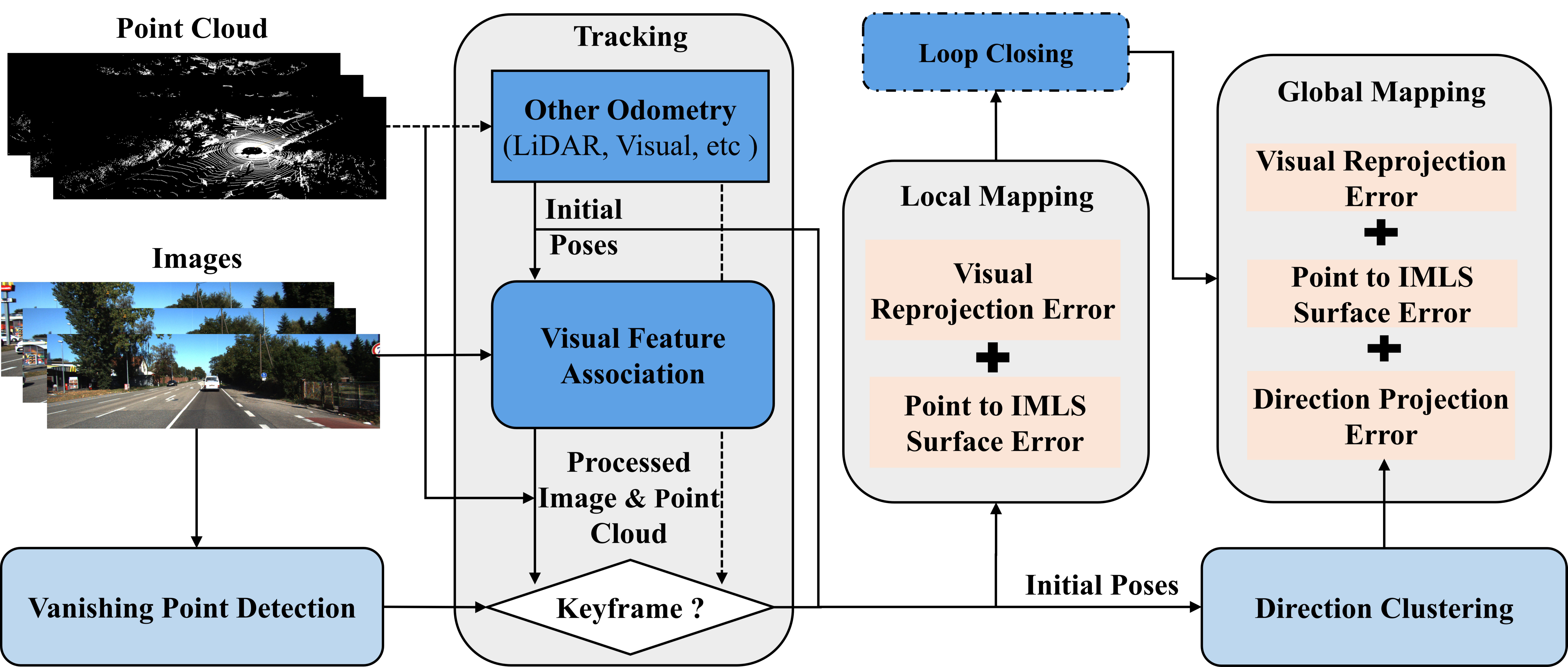}
	\caption{The Overall diagram of the proposed \textbf{JVLDLoc}.Our approach inputs image and point cloud stream and estimates 6DoF poses.
	}
	\label{fig:workflow}
\end{figure}
\sectionBefore
\section{Notation}
\label{sec:notation}
\sectionAfter

$\mathcal{I}_n$, $\mathcal{S}_n$ stands for an image and a point cloud at time $t_n$.
A frame $\mathcal{F}_n$ is the data structure that includes ORB features of $\mathcal{I}_n$ and corresponding processed scan $\mathcal{S}_n$,
while $\mathcal{K}_k$ refers to a keyframe.
Every 3D point $\textbf{\textrm{X}}=(X,Y,Z,1)^T\in \mathbb{R}^4$ in a homogeneous coordinate from visual map,
can be observed in the image plane
as a pixel coordinate $\mathbf{u}=(u,v,1)^T \in \mathbb{R}^3$ via a projection function: 
\begin{equation}
  \pi(\textbf{X}) = \\
  \frac{1}{Z}\textbf{K} [ \textbf{I}|\textbf{0}] \textbf{X} = \\
  \frac{1}{Z} \\
  \begin{bmatrix}
  f_x & 0 & c_x & 0\\
  0 & f_y & c_y & 0\\
  0 & 0 & 1 & 0  
  \end{bmatrix} \\
  \begin{bmatrix}
  X\\
  Y\\
  Z\\
  1
  \end{bmatrix} = \\
  \begin{bmatrix}
	f_x\frac{X}{Z}+c_x\\
	f_y\frac{Y}{Z}+c_y \\
	1 
  \end{bmatrix} = \\
  \begin{bmatrix}
	u\\
	v \\
	1 
  \end{bmatrix}
  \label{eq:pinholeproj}
\end{equation}
where $K\in\mathbb{R}^{3\times 3}$ is camera intrinsic matrix, $f_x$, $f_y$, $c_x$ and $c_y$ are intrinsic parameters.   
We use $T_{wn}\in SE(3)$ to express the transformation from the $n_{th}$ frame's camera coordinate to the world coordinate,
$T_{nw}$ is the inverse of $T_{wn}$.
$\textrm{R}_{wn} \in SO(3)$ and $\textrm{t}_{wn} \in \mathbb{R}^3$ are rotation and translation part of $T_{wn}$. 
\sectionBefore
\section{Method}
\label{sec:method}
\sectionAfter

The dataflow of the JVLDLoc is in \cref{fig:workflow}.
The whole system consists of four modules: tracking(\cref{subsec:tracking}), local mapping(\cref{subsec:localmapping}), visual based loop closing(see supplemental material for detail) and global mapping(\cref{subsec:globalopt}).
\subsectionBefore
\subsection{Tracking}
\label{subsec:tracking}
\subsectionAfter

We process visual and LiDAR data and compute their features and decide whether to generate a new keyframe.
Note that we need to transform point cloud to camera coordinate by given extrinsic: $T_{CL}$.
If there is no initial pose from other odometry, we can still handle it by launching a LiDAR odometry.
It is modified from the approach in IMLS-SLAM\cite{deschaud2018imls}.
The main changes compared to \cite{deschaud2018imls} can be found in supplemental material.

\subsubsectionBefore
\subsubsection{Visual Feature Association}
\label{subsubsec:visualpart}

Unlike \cite{mur2017orb}, we no longer launch motion-only BA.
Instead, we just use odometry poses 
to construct 2D feature-3D point associations within the visual map.
Finally, we incorporate processed image keypoints and point cloud into a frame $\mathcal{F}_n$.
Tracking module will treat current $\mathcal{F}_n$ as a new keyframe $\mathcal{K}_n$ if 4 conditions in \cite{mur2017orb} are satisfied \textbf{OR} $\mathcal{F}_n$
has valid supporting direction constraints(described in \cref{subsubsec:makedirection}).

\begin{figure}[t] 
\centering
\includegraphics[width=0.73\textwidth]{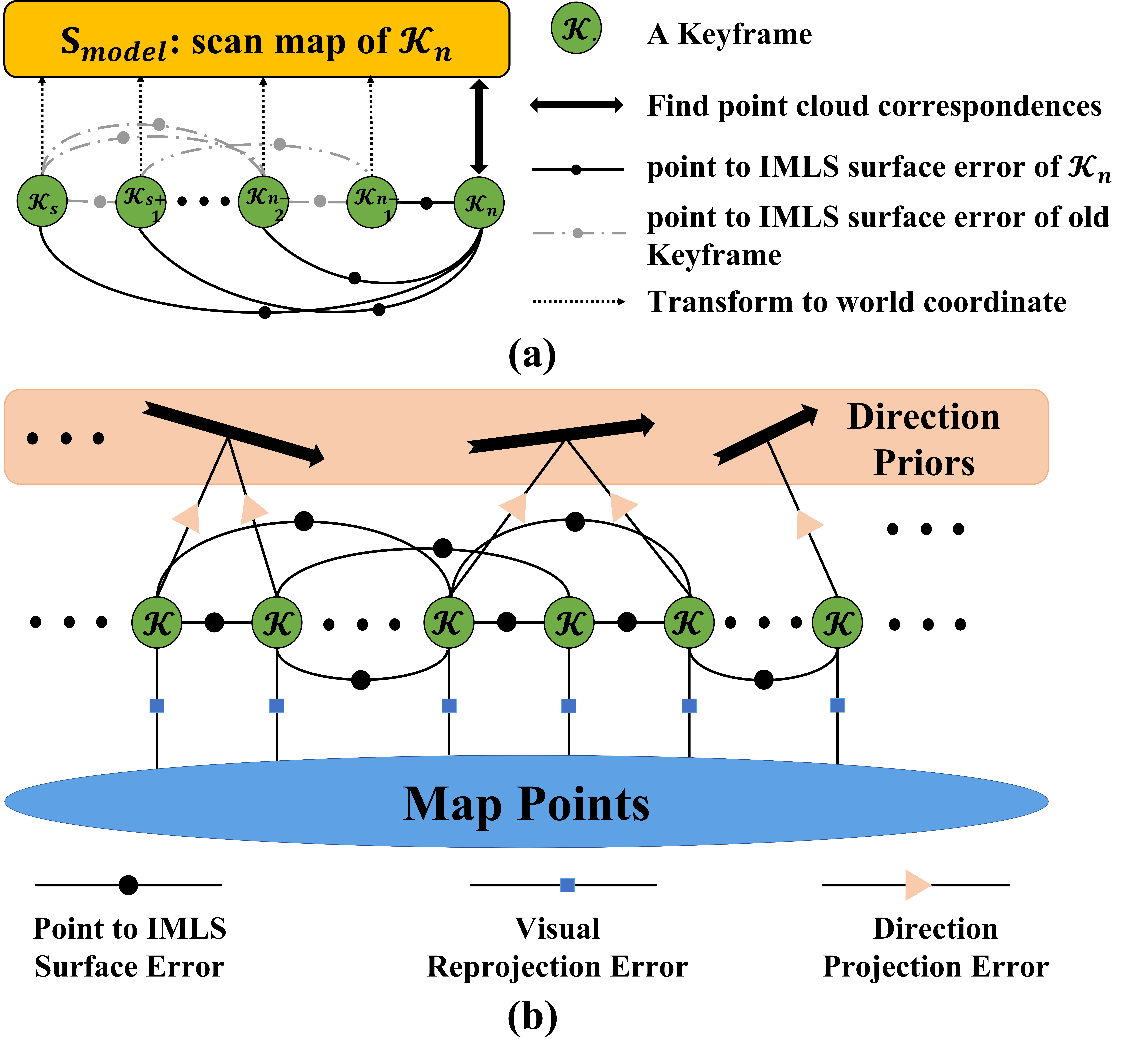}
\caption{(a)  Scan to map association of one Keyframe in Local and global mapping.
We build point to IMLS surface error between $\mathcal{K}_n$ and the keyframe to which the closest point $p_{c}$ belongs.
(b)  Global optimization graph structure.
}
\label{fig:buildimsandglobalgraph}
\end{figure}
\subsectionBefore
\subsection{Local Mapping}
\label{subsec:localmapping}
\subsectionAfter

A new keyframe $\mathcal{K}_n$ will trigger a visual-LiDAR graph optimization.
The corresponding joint energy function consists of visual reprojection error
and point to IMLS surface error.

Our backend maintains a covisiblity graph where keyframes that observe same 3D point are connected. 
Denote keyframe launching local mapping as $\mathcal{K}_n$, sort all the keyframes connected to $\mathcal{K}_n$ at covisiblity graph (abbreviated as $\mathbf{K}_{n}^{cov}$) including itself by their timestamps:
$\mathbf{K}_{local} = \left \{ ...\mathcal{K}_i...\mathcal{K}_n \right \}$. 
$\mathcal{P}_{local}$ refers to all 3D visual map points seen by $\mathbf{K}_{local}$.
Defining $\mathcal{X}_i= \{(\textrm{x}^{j}, \textbf{\textrm{X}}^j) | j= 0,1...\}$ as the set of correspondences between feature points at keyframe $\mathcal{K}_i$ and 3D visual map points from $\mathcal{P}_{local}$.
Then do the following steps:

\begin{itemize}
\label{itms:localpipe}
\item [1)] For all keyframes in $\mathbf{K}_{local}$, according to visual correspondences from $\mathcal{X}_k$, we can get visual reprojection error term like \cite{mur2017orb}:
		$\textbf{e}_{V}(i, j) = \| \textrm{x}^{j}_{(\cdot)} - \pi_{(\cdot)} ( \textrm{R}_{iw} \textbf{\textrm{X}}^j + \textrm{t}_{iw} ) \|^{2}_{\sum}$.

\item [2)] \label{itm:step2}
		For current keyframe $\mathcal{K}_n$ which has a global frame index $f_n$, we find the first keyframe whose frame index met $f_n-f_i \leq \textbf{N}$, where $\textbf{N}$ is the size of local scan map of each keyframe.
		Denote this keyframe as $\mathcal{K}_{start}$.
		Now list all corresponding scan from $\mathcal{K}_{start}$ to $\mathcal{K}_{n-1}$ : $\left \{ \mathcal{S}_{start},...,\mathcal{S}_{n-1} \right \}$.
		We use each keyframe's current pose to transform these scans to world frame and accumulate to be a whole point cloud model named $\mathbf{S}_{model}$(See \cref{fig:buildimsandglobalgraph} (a)).
\item [3)] \label{itm:step3}
		Sort every point in $\mathcal{S}_{n}$ by their computed nine feature values.
		Then transform $\mathcal{S}_{n}$ to world coordinate to find correspondences in its $\mathbf{S}_{model}$.
		we use the same scan sampling strategy as Section V in \cite{deschaud2018imls} to get sampled point cloud $\widetilde{\mathcal{S}}_n$.
		we compute for each point the distance to the IMLS surface in $\mathbf{S}_{model}$:
		\begin{equation}
		I^{n}(\textbf{p}_x) = \\
		\frac{\sum_{\textbf{p}_i \in \mathbf{S}_{model}} W_i(\textbf{p}_x)((\textbf{p}_x - \textbf{p}_i) \cdot \vec{n_c})}{\sum_{\textbf{p}_i \in \mathbf{S}_{model}} W_i(\textbf{p}_x) }
		\label{eq:imlsdist}
		\end{equation} 
		where $\textbf{p}_x\in\widetilde{\mathcal{S}}_n$ but in world coordinate, $\textbf{p}_i$ is the neighbor of $\textbf{p}_x$ from model scan and $\vec{n_c}$ is the normal of $\textbf{p}_c$ that is nearest neighbor of $\textbf{p}_x$.
		And $\textbf{p}_c$ is from $\mathcal{S}_{c}$ of $\mathcal{K}_c$.
		$W_i(\textbf{p}_x)=e^{-\|\textbf{p}_x - \textbf{p}_i \|^2/h^2}$ is the same weight function as \cite{deschaud2018imls}.
		Next we can obtain the projection of $\textbf{p}_x$ at IMLS surface, $\textbf{p}_y$: $\textbf{p}_y = \textbf{p}_x - I^{n}(\textbf{p}_x)\vec{n_c}$.
\item [4)] \label{itm:step4}
		Now we have found a association between $\mathcal{K}_{n}$ and $\mathcal{K}_{c}$.
		Denote $^{c}\textbf{p}_y = T_{cw}\textbf{p}_y$. 
		Finally, we can get point to IMLS surface error:
		\begin{equation}
		\textbf{e}_{L}(n, j) = \| (T_{cw}T_{nw}^{-1}\textbf{p}_x^j - ^{c}\textbf{p}_y^j ) \cdot \vec{n_c}^j\|^{2}
		\label{eq:imlsicp}
		\end{equation} 
		where $\textbf{p}_x^j$ is the $j_{th}$ point from $\widetilde{\mathcal{S}}_n$.

		All the correspondences $(\textbf{p}_x^j, ^{c}\textbf{p}_y^j, \vec{n_c}^j)$ for each $\textbf{p}_x^j$ and keyframe index $n$ and $c$ in \cref{eq:imlsicp} will be saved in $\mathbf{L} = \{ (n, c, \textbf{p}_x^j, ^{c}\textbf{p}_y^j, \vec{n_c}^j) \}$. 
		The previous saved correspondences in $\mathbf{L}$ will be taken out to build more costraints within $\mathbf{K}_{local}$, see supplemental material for detail. 
		

\item[5)] Our proposed local joint energy function is weighted sum of above two types of constraints:
		\begin{equation}
		\begin{split}
		\textbf{E}_{local} = \sum_{i\in\mathbf{K}_{local}}\alpha\sum_{j \in \mathcal{X}_i} \rho(\textbf{e}_{V}(i,j)) + 
		\beta \sum_{j \in \widetilde{\mathcal{S}}_n} \textbf{e}_{L}(n,j)
		\end{split}
		\label{eq:localjoint}
		\end{equation}
		where $\rho$ is the robust Huber cost function, $\alpha$ and $\beta$ are weight hyper parameters to balance those two kinds of measurements.
		The poses of keyframes in $\mathbf{K}_{local}$ are obtained by minimizing \cref{eq:localjoint}.
\end{itemize}

\begin{figure}[t] 
	\centering
	\includegraphics[width=0.8\textwidth]{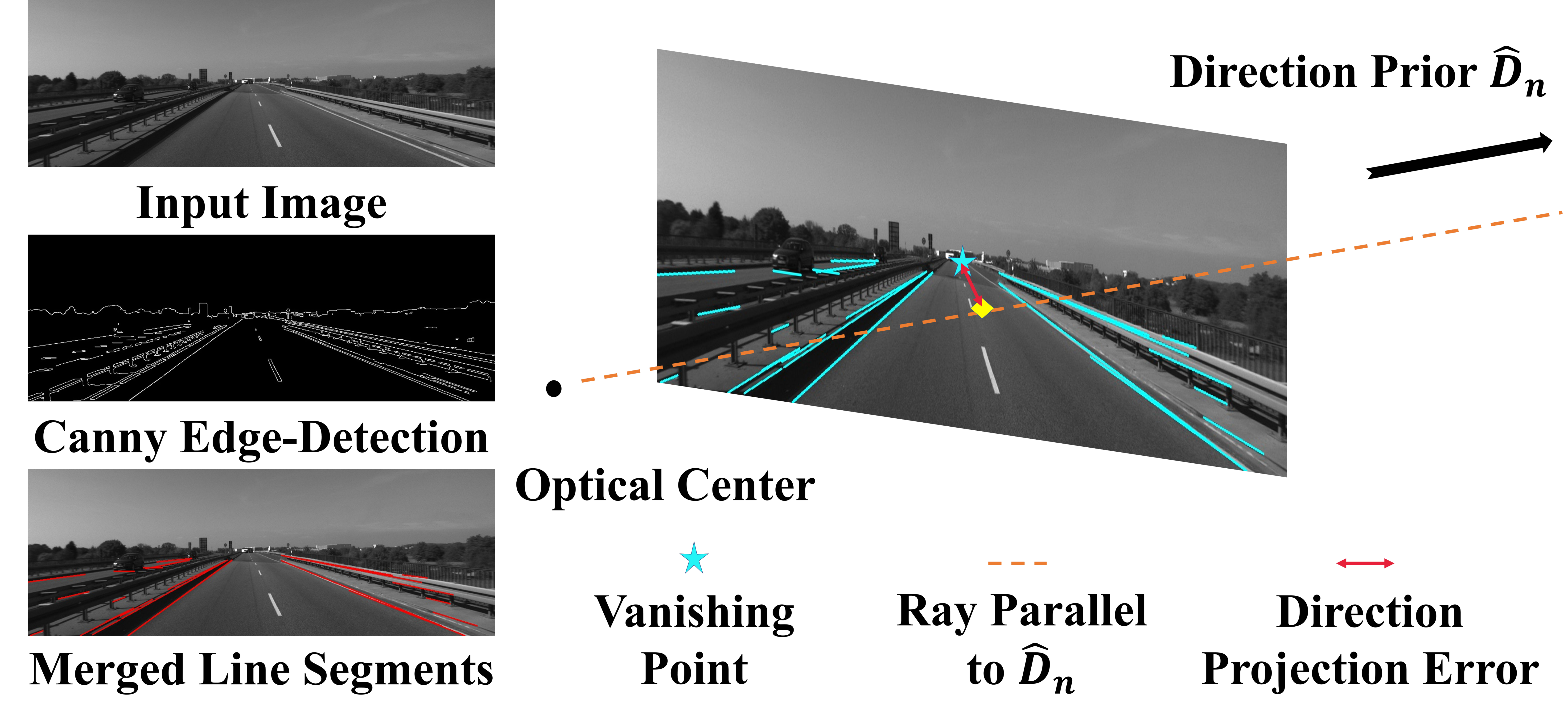}
	\caption{Construction of direction projection error.
	The left figure show the process of vanishing point detection.
	A ray parallel to the direction prior passing through the camera optical center intersects the image plane at $\tilde{\textit{v}}$ as \cref{eq:linetovp}.  
	}
	\label{fig:vp&directionerr}
\end{figure}

\subsectionBefore
\subsection{Global Mapping using Direction Priors} 
\label{subsec:globalopt}
\subsectionAfter

After local mapping of the final keyframe, we will launch the global mapping.

\subsubsectionBefore
\subsubsection{Association of Direction Priors and VPs}
\label{subsubsec:makedirection}

We detect the vanishing point $\textbf{v}$ in each image and get corresponding direction via $\textbf{d} = \textbf{\textrm{R}}_{wc}\textbf{K}^{-1}\textbf{v} / \| \textbf{\textrm{R}}_{wc}\textbf{K}^{-1}\textbf{v} \|$\cite{andrew2001multiple},
where $\textbf{d} \in \mathbb{R}^3$.
Detailed steps and bi-level mean-shift clustering on directions are in supplemental material.
After clustering, there is at most one VP per image.
Each VP corresponds to a cluster's center which is direction prior $\hat{\textbf{d}}_w$.
\subsubsectionBefore
\subsubsection{Global Joint Optimization}
\label{subsubsec:globalopt}



One can regard global optimization as a special case of local mapping: ALL keyframes in the map are included in $\mathbf{K}_{global}$.
We traverse all keyframes in $\mathbf{K}_{global}$ to construct \cref{eq:imlsicp}.
The global optimization graph is shown in \cref{fig:buildimsandglobalgraph} (b).
We expresse global direction prior $\hat{\textbf{D}}_w \in \mathbb{R}^4$ in a homogeneous coordinate as follows \cite{andrew2001multiple}:
$\hat{\textbf{D}}_w = [ \hat{\textbf{d}}_w, 0 ]^T $.

The direction prior $\hat{\textbf{D}}_w$ at world coordinate is transformed by current estimated rotation 
to camera's coordinate: $\hat{\textbf{D}}_n$. This direction can be written as a 3D line through a homogeneous point $\mathbf{A}$:
$\mathbf{Z}(\lambda) = \mathbf{A} + \lambda\hat{\textbf{D}}_n$, where $\lambda \in (0,\infty )$.
The observation on image of that direction is then equal to the projection of a point at infinity on that 3D line as follows:
\begin{equation}
  \tilde{\textbf{v}} = \lim_{\lambda \to \infty}\pi(\mathbf{Z}(\lambda))
  = \lim_{\lambda \to \infty}\pi((\mathbf{A} + \lambda\hat{\textbf{D}}_n)) \\
  = \pi(\hat{\textbf{D}}_n)
  \label{eq:linetovp}
\end{equation}

The direction projection error is just distance between $\tilde{\textbf{v}}$ and actual detected VP $\textbf{v}$ (See \cref{fig:vp&directionerr}):
\begin{equation}
  \textbf{e}_D(n) = \| \textbf{v} - \tilde{\textbf{v}} \|^{2}_{\sum} = \| \textbf{v} - \pi(\hat{\textbf{D}}_n) \|^{2}_{\sum} \\
  = \| \textbf{v} - \pi([\textbf{R}_{nw}|\textbf{I}]\hat{\textbf{D}}_w) \|^{2}_{\sum}
  \label{eq:adirectionerr}
\end{equation} 

Finally, we take out all keyframes $\mathbf{K}_{global}$ from the global map, the global joint energy function can be written as:

\begin{equation}
\textbf{E}_{global} = \sum_{n \in \mathbf{K}_{global}} \{ \alpha \sum_{j \in \mathcal{X}_n} \rho(\textbf{e}_{V}(n, j)) + \\
  \beta \sum_{j \in \widetilde{\mathcal{S}}_n} \textbf{e}_{L}(n, j) + \gamma \textbf{e}_D(n)\}
  \label{eq:globalloss}
\end{equation}
where $\alpha$, $\beta$, $\gamma$ are weights factors to balance these three types of residuals.
In practice, we set $\alpha$ to be 1 and set the other two according to the corresponding number of residuals.  
The poses are optimized by  : 
\begin{equation}
  \{\textbf{R}_n, \textbf{t}_n | n\in\mathbf{K}_{global} \} = \arg\min_{\textbf{R}_n, \textbf{t}_n} \textbf{E}_{global}
  \label{eq:solveglobal}
\end{equation}

Limited to the length of the paper, please refer to our supplemental material for the jacobin of direction projection error.

\begin{figure}[t]
	\centering
	\begin{subfigure}{0.21\textwidth} 
		\includegraphics[width=\textwidth, interpolate]{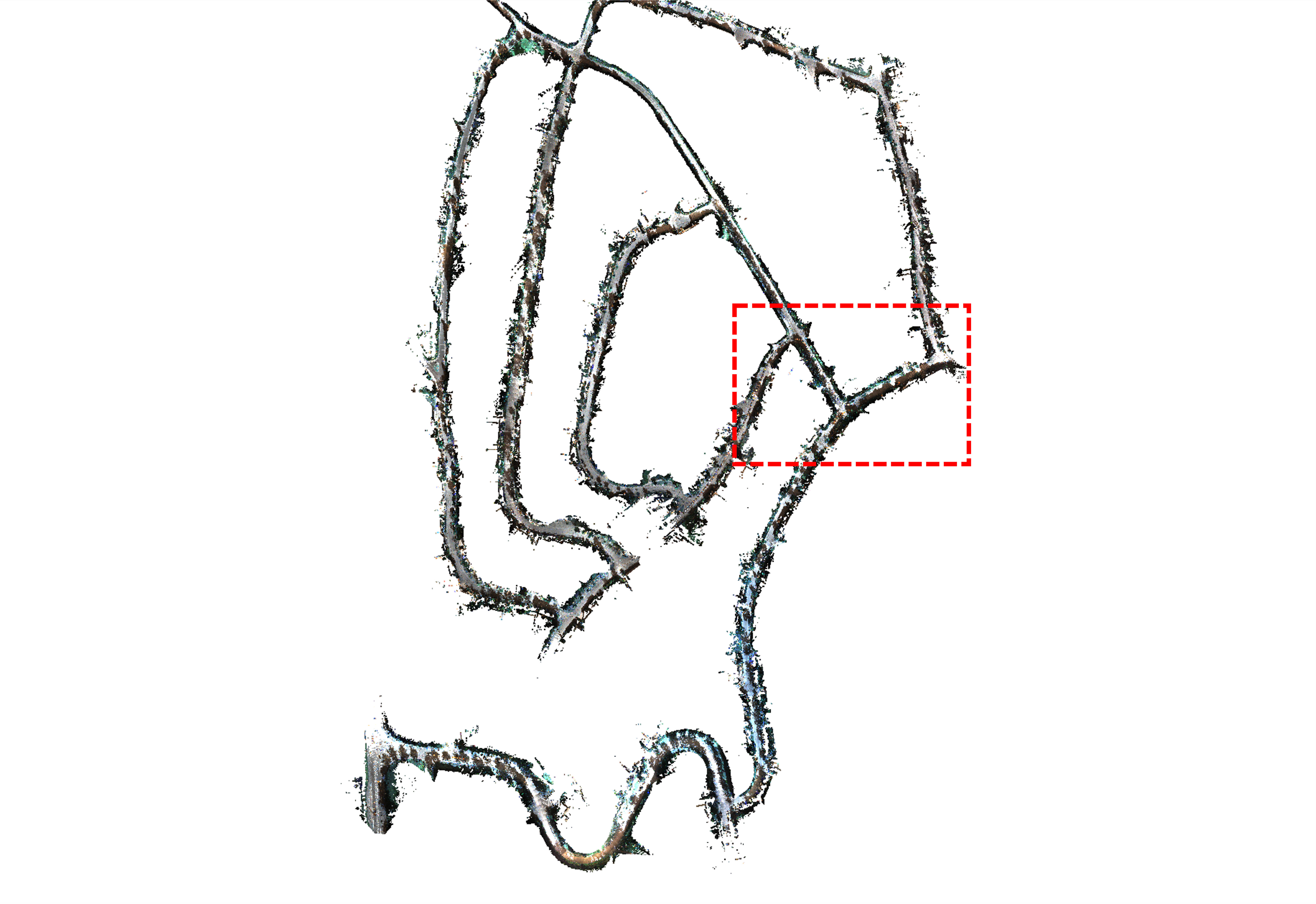}
		\caption{Full view}
	\end{subfigure}
	\begin{subfigure}{0.24\textwidth}
		\includegraphics[width=\textwidth, interpolate]{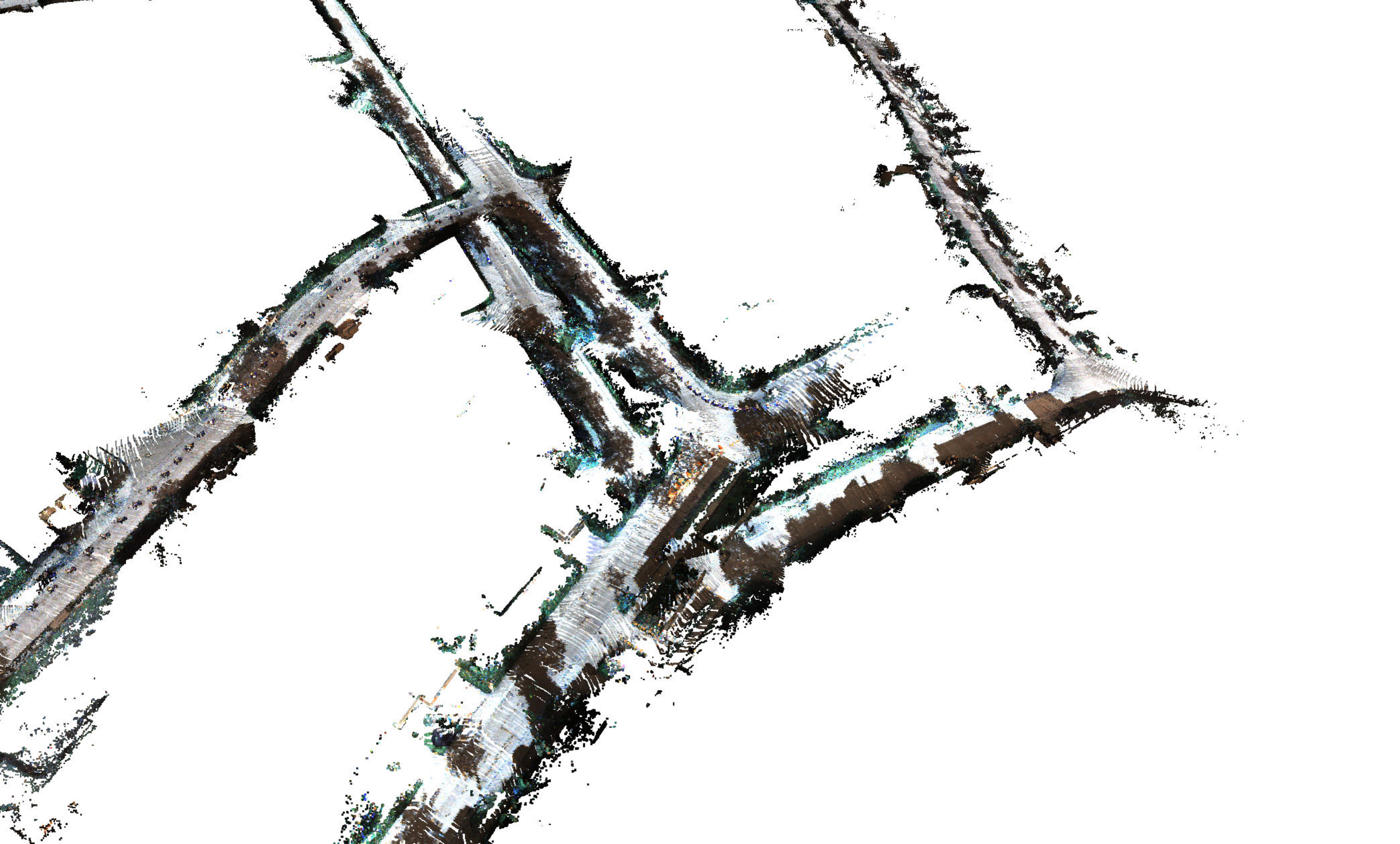}
		\caption{LiDAR odo}
	\end{subfigure}
	\begin{subfigure}{0.24\textwidth} 
		\centering
		\includegraphics[width=\textwidth, interpolate]{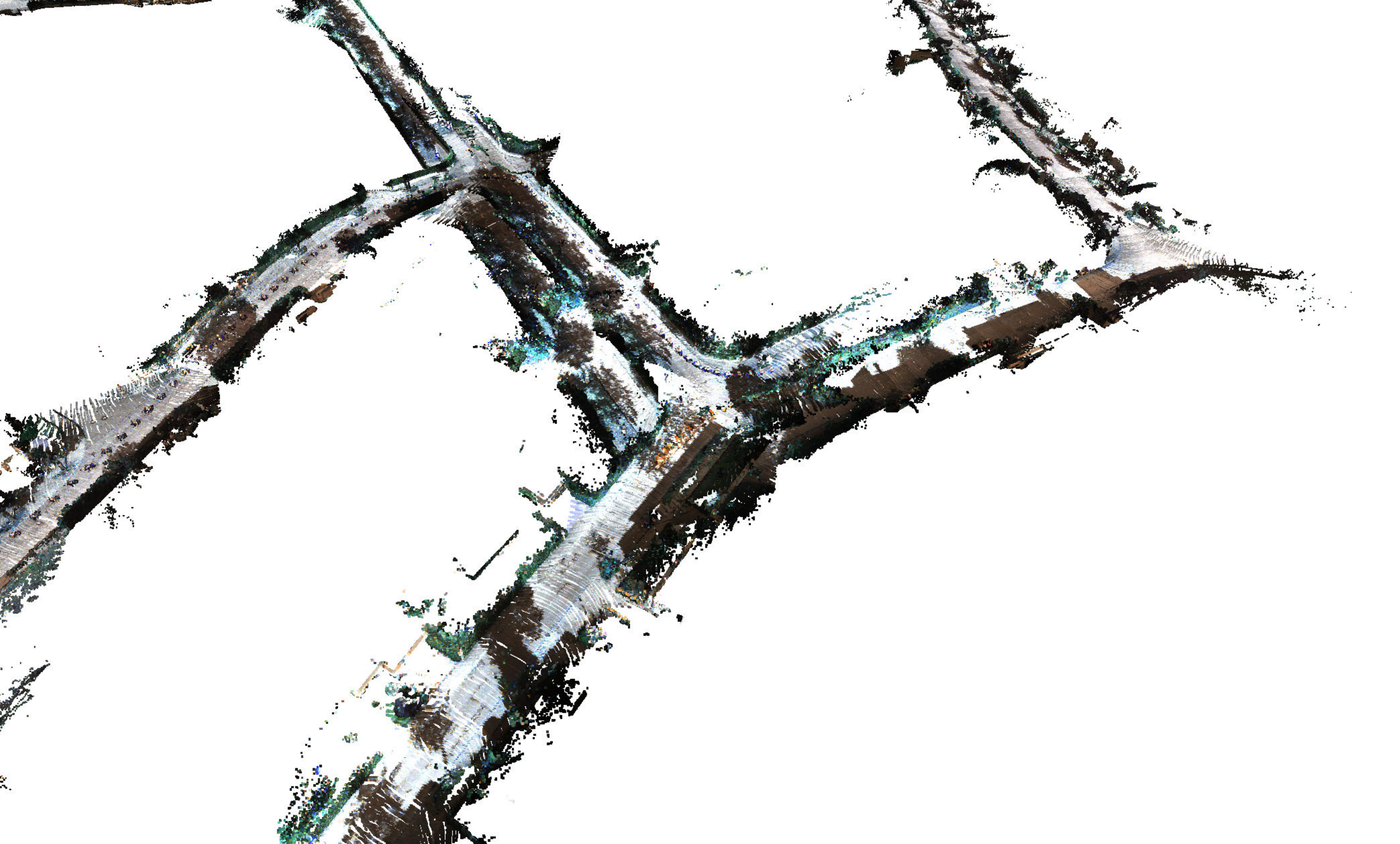}
		\caption{IMLS-SLAM}
	\end{subfigure}
	\begin{subfigure}{0.24\textwidth} 
		\centering
		\includegraphics[width=\textwidth, interpolate]{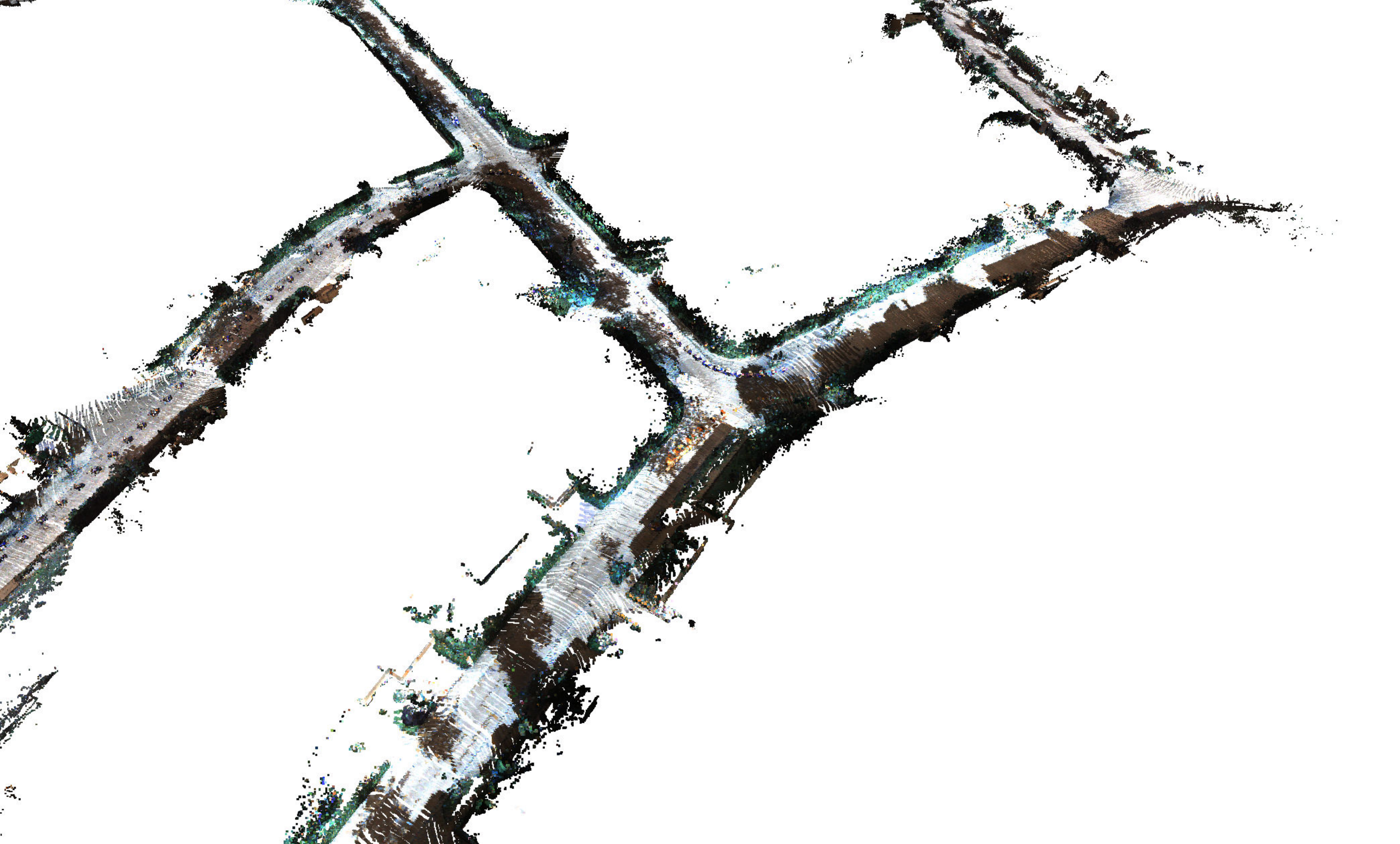}
		\caption{JVLDLoc}
	\end{subfigure}
	\caption{
	Mapping result on KITTI 02.
	(a) is the full view output from JVLDLoc. (b), (c), (d) are the enlargement of the red box region in (a) output from different methods respectively.
	LiDAR odo means LiDAR Odometry with $\textbf{N}=1$.
	}
	\label{fig:lidarmap}
\end{figure}

\sectionBefore
\section{Experiments}
\label{sec:experiment}
\sectionAfter


This method is implemented in C++ and Python.
All the non-linear optimization mentioned above are Levenberg-Marquardt method in $g^{2}o$ library\cite{kummerle2011g}.
Our experiments involve the following 3 driving datasets which contain camera and LiDAR sensors:

\begin{table}[t] 
	\caption{RMSE(m) of APE for KITTI 00-10.
	For each sequence, we report RMSE of input pose, JVLDLoc without direction priors and full JVLDLoc respectively.}
	\label{tab:bettertoprior}
	\centering
		\resizebox{1.0\textwidth}{!}
		{
			\begin{tabular}{l||c|c|c|c|c|c|c|c|c|c|c|c}
				\toprule
				{KITTI Odometry}& {00} & {01} & {02} & {03} & {04} & {05} & {06} & {07} & {08} & {09} & {10} & {AVG} \\
				\midrule
				\hline 
				\noalign{\smallskip}

				ORB-SLAM2 \cite{mur2017orb}    
				&1.354
				&10.16
				&6.899
				&0.693
				&0.178
				&0.783
				&0.714
				&0.536
				&3.542
				&1.643
				&1.156
				& 2.515
				\\
				w.o. direction priors
				&1.367
				&3.289
				&4.669
				&0.614
				&0.169
				&0.705
				&0.485
				&0.473
				&3.354
				&2.433
				&0.990
				& 1.686
				\\
				w. direction priors
				&1.205	  
				&1.999 
				&3.544 
				&0.530	  
				&0.151
				&0.593
				&0.300
				&0.441
				&2.862
				&1.498
				&0.818
				&\textbf{1.267}
				\\ 
				\hline
				\noalign{\smallskip}
				LiDAR odometry $\textbf{N}$=1    
				&6.849	  
				&35.03
				&12.74
				&0.839	  
				&0.265
				&3.379
				&2.109	  
				&0.769 
				&8.123 
				&3.917	  
				&1.084
				&6.828
				\\
				w.o. direction priors
				&2.027	  
				&5.932
				&4.414
				&0.764	  
				&0.186
				&1.286
				&0.374	  
				&0.945 
				&4.602 
				&3.347	  
				&1.065
				&2.268
				\\
				w. direction priors
				&1.616	  
				&3.097 
				&3.015 
				&0.585	  
				&0.175
				&0.802
				& 0.302
				& 0.521
				& 3.131
				&1.828
				&0.933
				&\textbf{1.455}
				\\ \bottomrule
			\end{tabular}
		}
\end{table}


\begin{figure}[t]
	\centering
	\begin{subfigure}{0.32\textwidth}
		\includegraphics[width=\textwidth, interpolate]{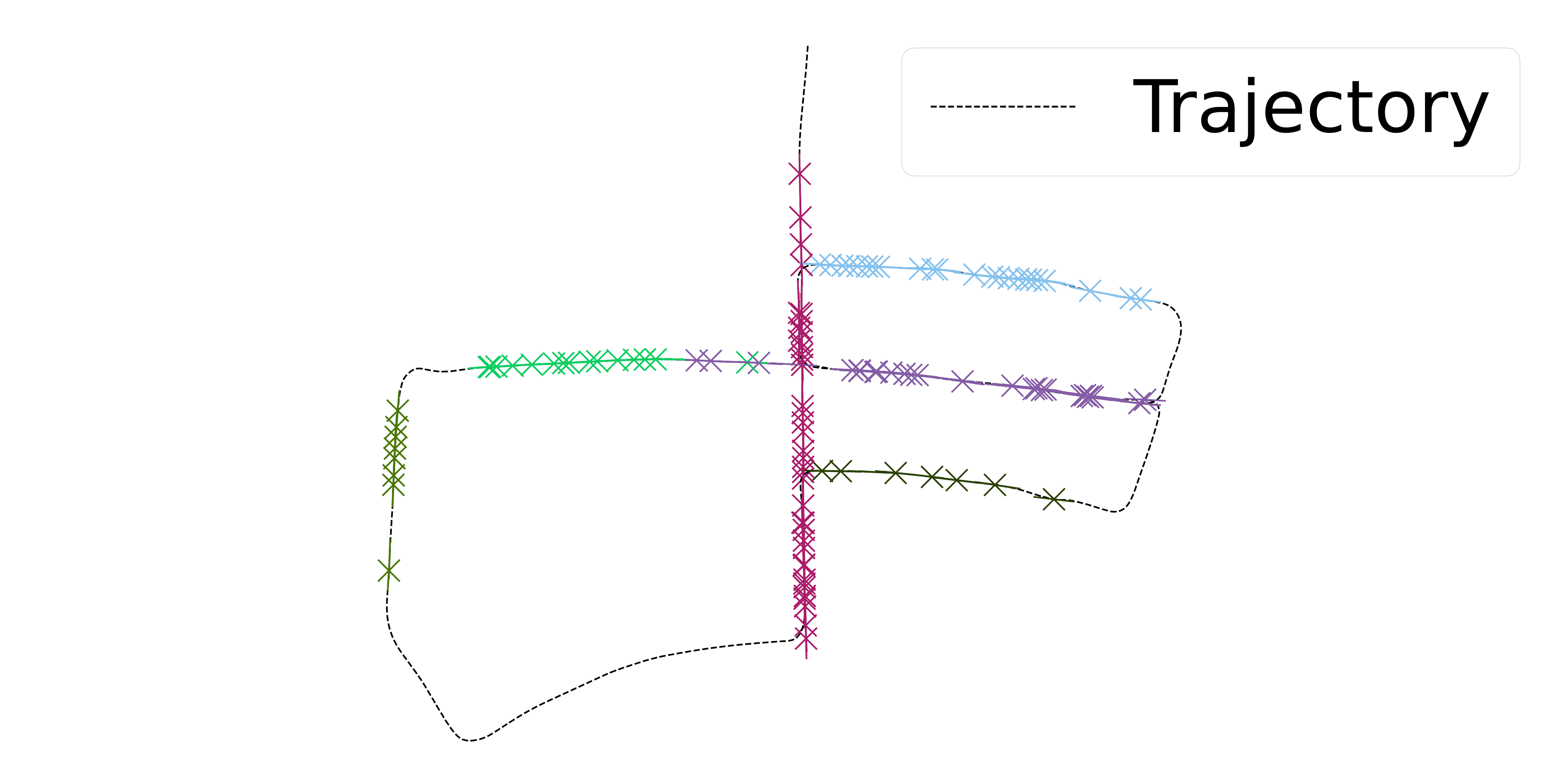}
	\end{subfigure}
	\begin{subfigure}{0.32\textwidth} 
		\centering
		\includegraphics[width=\textwidth, interpolate]{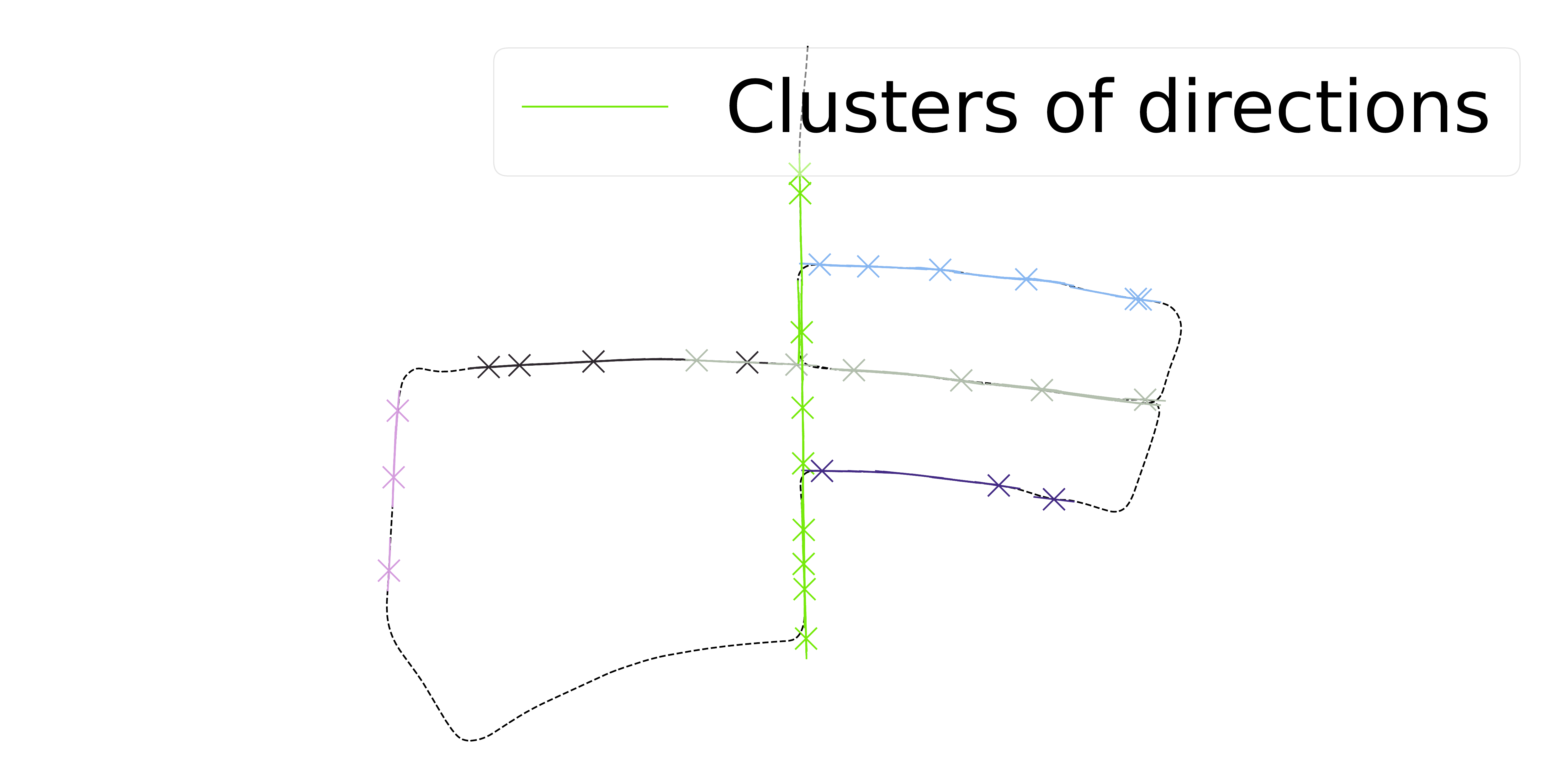}
	\end{subfigure}
	\begin{subfigure}{0.32\textwidth} 
		\centering
		\includegraphics[width=\textwidth, interpolate]{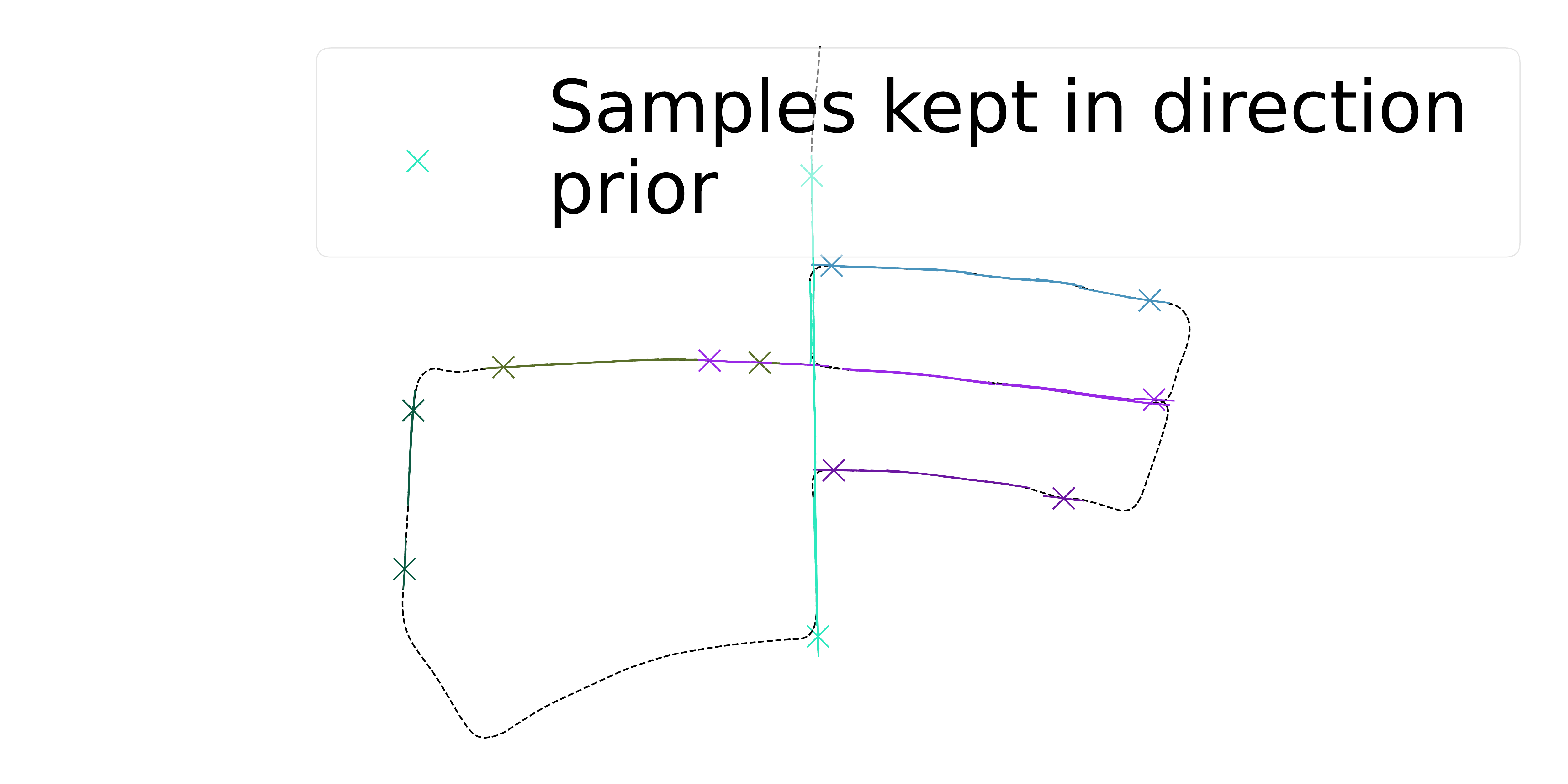}
	\end{subfigure}
	\caption{
	A visualization of direction priors of KITTI 05 when downsampled under different intervals $V_{interval}$: 2, 15, 300(from left to right).
	The number of total direction projectoin error is reduced from 117 to 12.}
	\label{fig:showlessVP}
\end{figure}

\begin{table}[t]
	\caption{APE for KITTI-360 and Oxford Radar RobotCar.
	$MIN$ and $MAX$ are the minimum and maximum APE over the entire trajectory.}	
	\label{tab:bettertoprior4}
	\begin{minipage}[t]{0.58\textwidth}
	 \centering
	 \resizebox{\textwidth}{!}
		{
		\begin{tabular}{l||ccc|ccc}
			\toprule
			&  \multicolumn{3}{c|}{12}  &\multicolumn{3}{c}{20}  \\ 
			\multirow{-2}{*}{\begin{tabular}[c]{@{}c@{}}KITTI-360 \end{tabular}}
			&  $MIN$  & $RMSE$ & $MAX$  &  $MIN$  & $RMSE$ & $MAX$ \\ 
			\midrule
			\hline
			\noalign{\smallskip}

			ORB-SLAM2 \cite{mur2017orb}    
			&0.538 &1.904 &3.289
			&0.385&2.253 &4.812
			\\
			w.o. direction priors
			&0.196 &1.367 &3.934
			&0.164 &0.989 &2.952
			\\
			w. direction priors
			&\textbf{0.057} &\textbf{0.891} &	\textbf{2.828}  
			&\textbf{0.129} &\textbf{0.786} & \textbf{1.980}
			\\ \bottomrule
		\end{tabular}
		}
	\end{minipage}
	\begin{minipage}[t]{0.42\textwidth}
		\centering
		\resizebox{\textwidth}{!}
		{
			\begin{tabular}{l||ccc}
				\toprule
				&  \multicolumn{3}{c}{1}  \\ 
				\multirow{-2}{*}{\begin{tabular}[c]{@{}c@{}}Oxford \end{tabular}}
				&  $MIN$  & $RMSE$ & $MAX$   \\ 
				\midrule
				\hline
				\noalign{\smallskip}
				Lidar odometry $\textbf{N}$=10    
				&2.592 &26.25 & 53.43
				\\
				w.o. direction priors
				&2.158 &11.09 & \textbf{19.01}
				\\
				w. direction priors
				&\textbf{0.437} &\textbf{7.982} & 27.08
				\\ \bottomrule
			\end{tabular}
		}
	\end{minipage}
\end{table}




\begin{figure}[t]
	\centering
	\begin{subfigure}{0.305\textwidth} 
		\centering
		\includegraphics[width=\textwidth, interpolate]{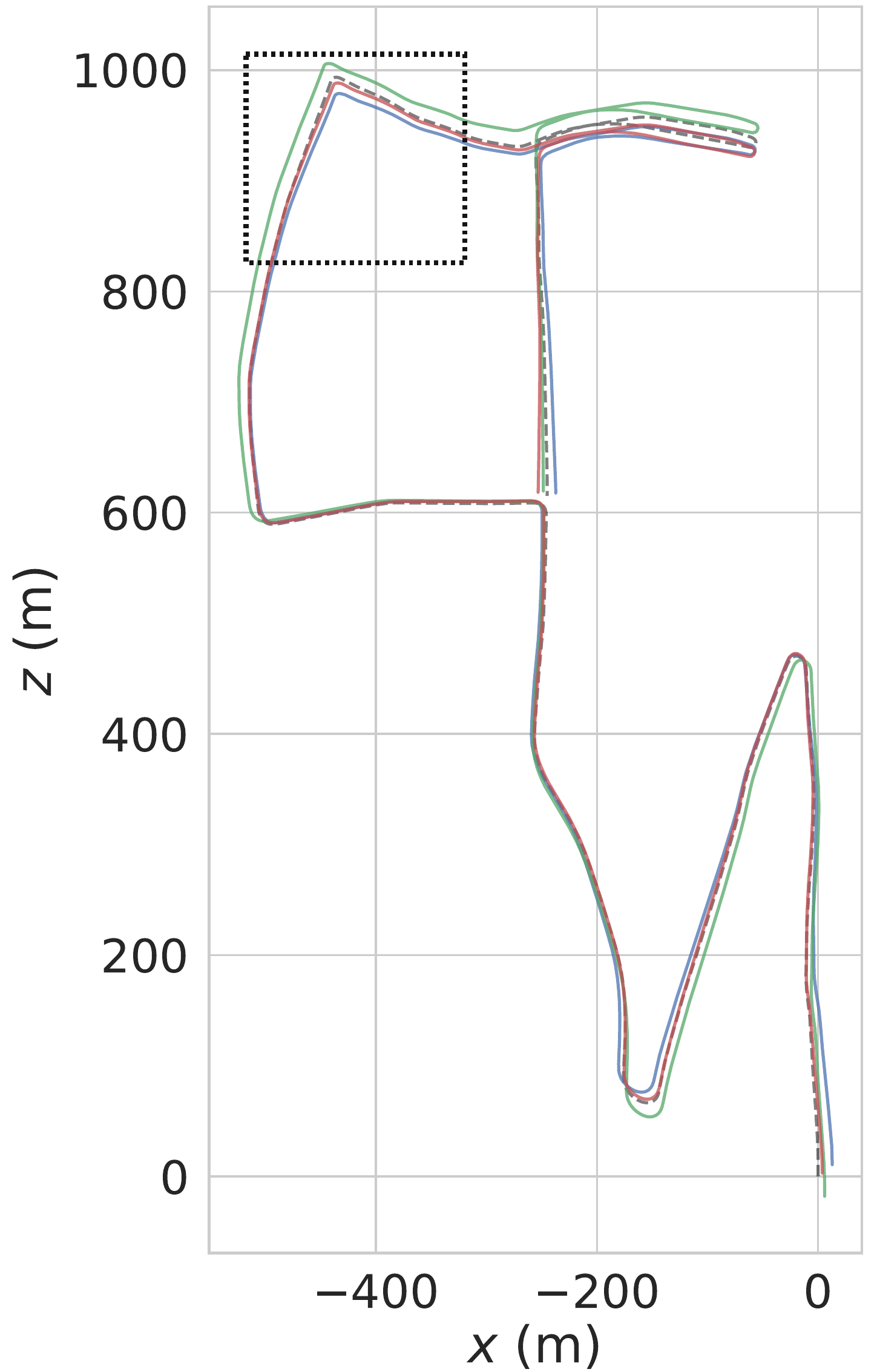} 
	\end{subfigure}
	\begin{subfigure}{0.45\textwidth} 
		\includegraphics[width=\textwidth, interpolate]{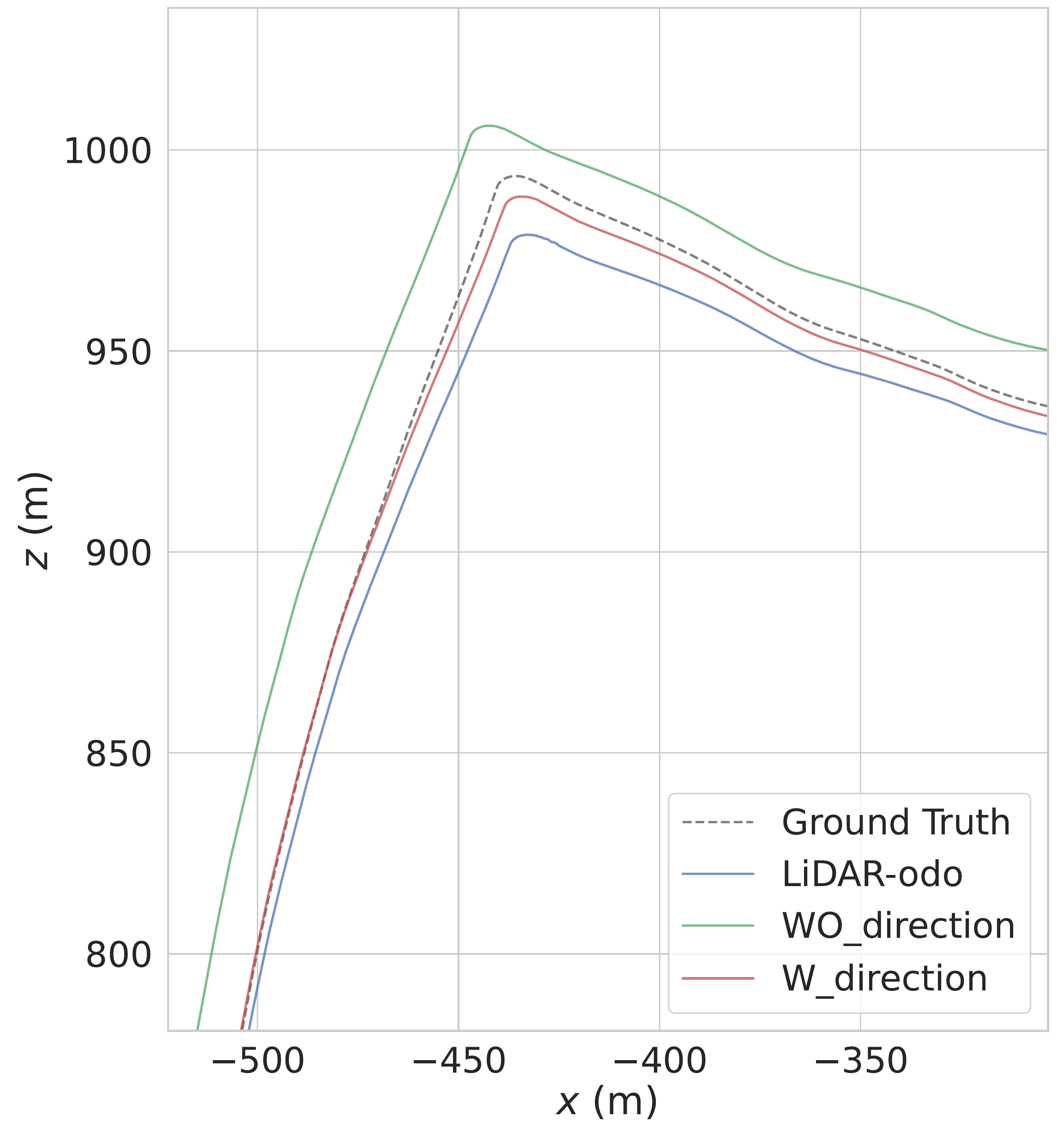}
	\end{subfigure}
	\caption{Trajectory of JVLDLoc(red), JVLDLoc without direction priors(green), LiDAR odometry(blue) on Oxford. The right image is the enlargement of the dotted box on the left image.}
	\label{fig:trajcompwod2}
\end{figure}

\subsubsectionBefore
\subsubsection{KITTI Odometry Dataset} 
\cite{geiger2013vision} may be the most popular dataset for SLAM field, which covers urban city, rural road, highways, roads with vegetation, etc.
We compare the algorithm with other state-of-the-art method on 00-10 because they are widely used in other papers.
\vspace{-6pt}
\subsubsectionBefore
\subsubsection{KITTI-360 Dataset} \cite{Liao2021ARXIV} 
is still under development, the ground truth does not fully cover each frame of raw data, because they discard some frames collected at very low moving speed.
For the convenience of evaluation, we clip two sequences from raw data as test data.
\subsubsectionBefore
\subsubsection{Oxford Radar RobotCar Dataset}\cite{RadarRobotCarDatasetICRA2020} 
is an another fresh dataset.
The car is driven to travel 32 times in same scene.
We selsect a sequence recorded on datetime 2019-01-10-14-36-48. 

\subsectionBefore
\subsection{Improvements over Prior Map} 
\subsectionAfter
To prove that JVLDLoc will reduce global drift, 
we visualize the output point cloud map with color in \cref{fig:lidarmap}. 
Qualitively, our method greatly improves the global consistency of the map.
For more mapping visualization, please turn to supplemental material.
We use RMSE of APE\cite{grupp2017evo} to quantify the drift.


\cref{tab:bettertoprior} includes results test on KITTI odometry when fed with diferent prior map. 
We have set four baselines as input odometry including ORB-SLAM2\cite{mur2017orb}, our LiDAR odometry under different $\textbf{N}$ which is the number of scans kept as local map.
The other two groups of result is shown in supplemental material.   
Under different quality prior maps, the average accuracy of these 11 sequences can be significantly improved.
Especially when input LiDAR odometry with $N=1$, the mean APE is reduced by about $79\%$.
\cref{tab:bettertoprior4} is for the other two datasets, which also prove our method's effect.
That is because JVLDLoc directly uses differnent kinds of geometric constraints to optimize pose.

\subsectionBefore
\subsection{Effects of Direction Priors}
\subsectionAfter
To demonstrate the effects of direction priors itself, we further compare accuracy with and without using the proposed direction priors.
APE are still shown in \cref{tab:bettertoprior}, \cref{tab:bettertoprior4}. 
It can be seen the direction priors are able to further
improve localization due to smaller average RMSE on KITTI.
For other two datasets, with help of direction priors, maximum of APE strongly decrease on KITTI-360 from the level when there is none direction constraint;
The minimum and RMSE have again significantly reduced by $80\%$ and $28\%$ on Oxford Radar RobotCar.

In a word, although the direction priors are not very accurate as they are based on input poses, they still contribute to better
localization.
We believe the reason is that direction priors impose longer term geometric constraints as the direction is an infinite landmark. 
\cref{fig:trajcompwod2} presents intuitive trajectory comparisons.
Apparently, we can get less drift after long distance under direction priors.

\begin{table}[t] 
	\centering
	\caption{Comparison to other methods on KITTI 00-10.
	$t_{rel}$ and $r_{rel}$ are relative translational error (\%) and rotational error ($^{\circ}$/100m).
	Due to page limitations, results on 05-10 are in supplemental material.
	(NA : Not Available).}	
	\label{tab:compwithsota}
		\resizebox{1.0\textwidth}{!}
		{
			\begin{tabular}{l||ccc|ccc|ccc|ccc|ccc|ccc}
				\toprule
				&  \multicolumn{3}{c|}{00}  &\multicolumn{3}{c|}{01}      & \multicolumn{3}{c|}{02} & \multicolumn{3}{c|}{03} &  \multicolumn{3}{c|}{04} & \multicolumn{3}{c}{AVG on 00-10} \\ 
				
				\multirow{-2}{*}{\begin{tabular}[c]{@{}c@{}}Method \end{tabular}}
				&  $t_{rel}$  & $r_{rel}$ & $APE$  & $t_{rel}$ & $r_{rel}$ & $APE$  & $t_{rel}$ & $r_{rel}$   & $APE$   & $t_{rel}$ & $r_{rel}$ & $APE$   & $t_{rel}$  & $r_{rel}$ & $APE$   & $t_{rel}$      & $r_{rel}$  & $APE$ \\
				\midrule
				\hline
				\noalign{\smallskip}

				IMLS-SLAM \cite{deschaud2018imls}    
                &0.50 &0.18 &	3.90
                &0.82 &0.10 &	2.41
                &0.53 & 0.14 &7.16
                &0.68 &0.22 &	0.65
                &0.33 & 0.12 &0.18
                & \textbf{0.52} & 0.14 &1.96
                \\ 
				LIMO-PL \cite{Huang2020LidarMonocularVO} 
				&0.99  &NA&	NA	  
				& 1.87 &NA&	NA
				& 1.38 &NA&	NA 
				& 0.65 &NA&	NA	  
				& 0.42 &NA&	NA
				& 0.94 &NA&NA
				\\ 
				DVL-SLAM \cite{Shin2020DVLSLAMSD}    
				&0.93  &NA&	NA	  
				& 1.47 &NA&	NA
				& 1.11 &NA&	NA 
				& 0.92 &NA&	NA	  
				& 0.67 &NA&	NA
				& 0.98 &NA&NA
				\\ 
				TVL-SLAM-No calib \cite{chou2021efficient}    
				&0.59  &NA&	0.84	  
				& 1.08 &NA	&	6.56
				& 0.74 & NA&	2.16 
				& 0.71 &NA&	0.75	  
				& 0.49 &NA&	0.18
				& 0.62 & NA  &	1.50
				\\ 
				TVL-SLAM-VL calib \cite{chou2021efficient}     
				&0.57  &NA&	0.88	  
				& 0.86 &NA	&	4.40
				& 0.67 & NA&	1.87 
				& 0.71 &NA&	0.74	  
				& 0.45 &NA&	0.22
				& 0.57 & NA  &	1.19
				\\ 
				LIO-SAM    \cite{shan2020lio} 
				&253  &31.0&	956	  
				& 2.94 &0.60	&	19.8
				& 268 & 87.6&	212
				& NA &NA&	NA	  
				& 0.92 &0.47&	0.30
				& 1.08 &  0.40&	4.16
				\\
				JVLDLoc input IMLS    
				&0.63  &0.15&	1.31	  
				& 0.88 &0.06	&	1.77 
				& 0.65 & 0.11&	2.62 
				& 0.70 &0.14&	0.53	  
				& 0.29 &0.10&	0.15
				& 0.54 & \textbf{0.12}   &	\textbf{1.13}
				\\ \bottomrule
			\end{tabular}
		}
\end{table}

\subsectionBefore
\subsection{Comparison to Other Methods on KITTI Odometry Dataset}
\subsectionAfter
Besides APE, \cref{tab:compwithsota} also calculates relative transition error and rotation error for 100, 200,... and 800 meter distances with KITTI's official evaluation code.


Compared to other visual-LiDAR methods, JVLDLoc achieve best relative translation error and APE, which indicates the superiority of our method over others using the same modality of data.
LIMO-PL\cite{Huang2020LidarMonocularVO} and DVL-SLAM\cite{Shin2020DVLSLAMSD} are loosely-coupled methods, which cause quite large translation error;
TVL-SLAM\cite{chou2021efficient} is a SOTA visual-LiDAR SLAM on KITTI, which also incorporates visual and lidar measurements in backend optimization.
TVL-SLAM VL calib refers to their online camera-lidar extrinsics calibration version.
Our JVLDLoc can outperform TVL-SLAM VL calib though using raw extrinsics.
LIO-SAM\cite{shan2020lio} is representative method of Lidar-inertial SLAM. 
We run its open source code with given parameters \footnote{\url{https://github.com/TixiaoShan/LIO-SAM\#other-notes}}.
Due to the unknown intrinsics of the IMU, LIO-SAM failed on 00, 02, 08, but we can get better acuracy on three metrics for all left sequences. 

\subsectionBefore
\subsection{Ablation Study}
\subsectionAfter


\begin{figure}[t]
	\centering
	\begin{subfigure}{0.48\textwidth}
		\centering
		\includegraphics[width=\textwidth, interpolate]{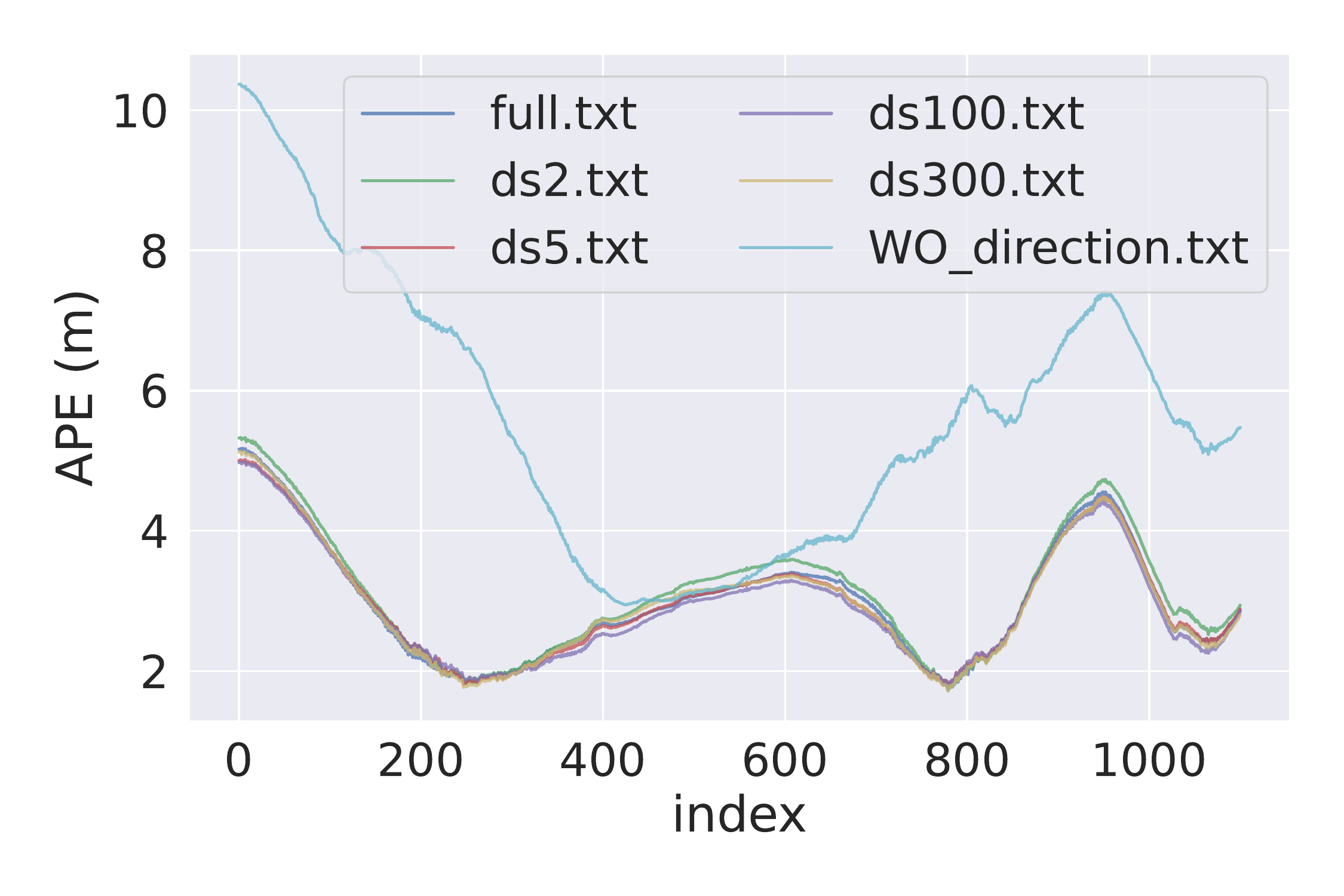} 
	\end{subfigure}
	\begin{subfigure}{0.48\textwidth}
		\includegraphics[width=\textwidth, interpolate]{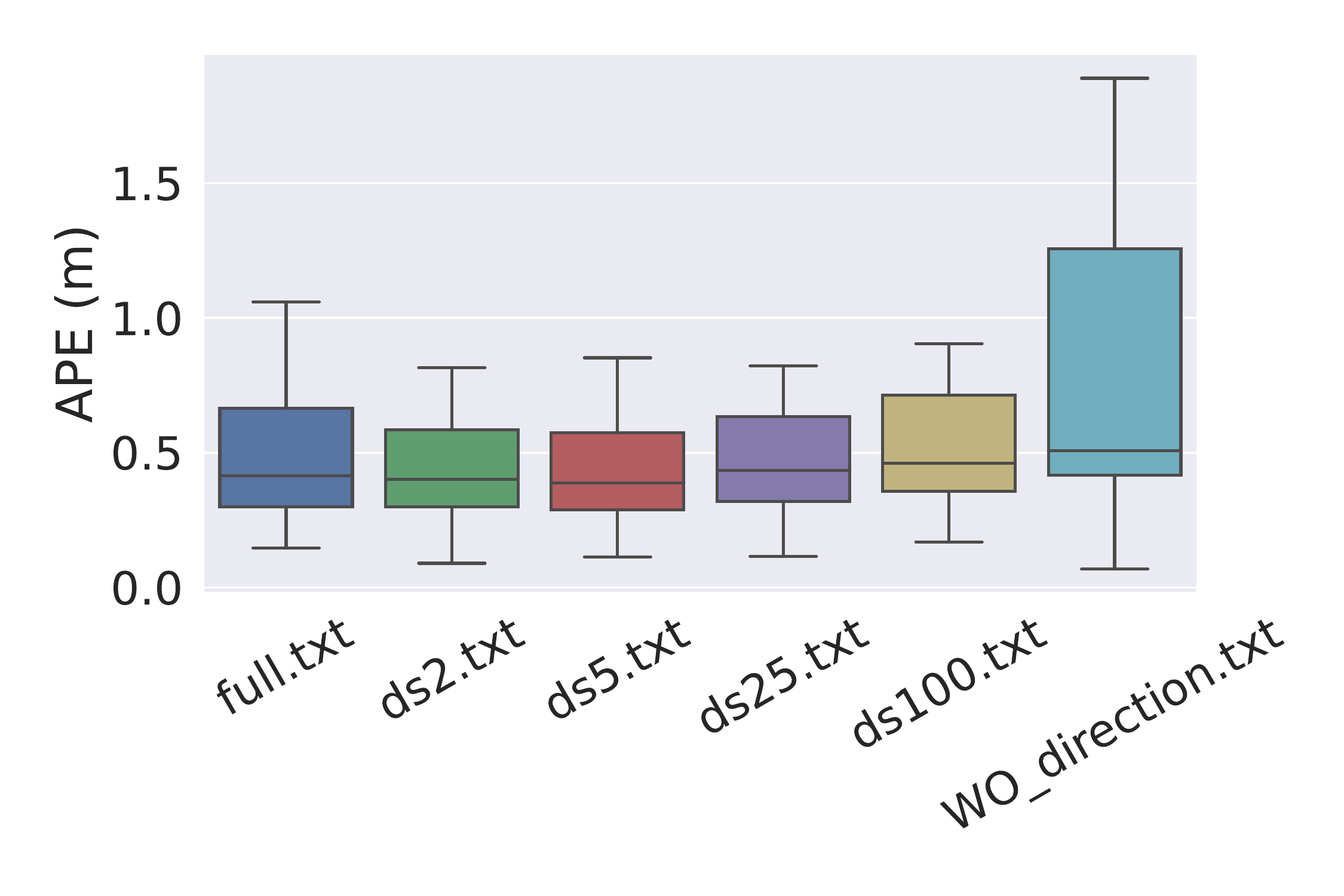} 
	\end{subfigure}
	\caption{
		left: Comparison of APE on KITTI 01.The x-coordinate is frame index. 
		right: Box plots of APE on KITTI 07. 
		$full$ refers to no downsampling, $WO direction$ refers to no direction priors.
		Other lines respectively corresponds to different $V_{interval}$.
		Take $ds2.txt$ as an example, it is the APE when $V_{interval}=2$.
	}
	\label{fig:apeboxdsVP}
\end{figure}

\subsubsection{Different Number of Direction Priors}
Here we attempt to relax the strength of direction priors by reducing the number of keyframes involved in direction priors.
For each direction cluster, downsample at a distance $V_{interval}$ from the distribution of existing VPs.
\cref{fig:showlessVP} is a visualization of different downsampled direction priors.
%
\cref{fig:apeboxdsVP} shows the comparison of APE under different $V_{interval}$. 

From those figures, we can concluded that the APE do not suffer drastic
expansion due to less direction constraints. 
Even though there is only about 10 direction constraints retained, we still get quite better accuracy than no direction priors at all.
Because the left direction priors can still play some roles. 
So we argue that as long as there is a few direction constraints over a long distance, the drift will be reduced at a certain level.
For each KITTI sequence's relative rotation error and APE RMSE under differnent $V_{interval}$, see supplementary material.

We also explore the influence of different scan map size $\textbf{N}$ of each keyframe.
The plot and analysis of that result are in supplemental material (page limit).

\sectionBefore
\section{Conclusion}
\sectionAfter
This paper present a strategy that combines map prior and VPs from images to create direction projection error, which is an energy term solely bound on rotation.
Then, we implement these direction priors into a visual-LiDAR localization system that tightly couples camera and LiDAR measurements.
Experiments on three driving datasets prove that our method achieve lower localization drift.
Even though our method is effective and outperforms many related work, There are still two flaws.
One thing is time efficiency.
Our current implementation can not run in real-time.
It is a trade-off between performance and time cost. 
The second thing is that our method relies on an assumption that agent locally moves on an approximate plane.
So compared to UAV or handheld devices, the driving scenes are more likely to meet the above assumptions.

\subsubsection{Acknowledgements.}
This work is supported by the National Key Research and Development Program of China (2017YFB1002601), National
Natural Science Foundation of China (61632003, 61375022,
61403005), Beijing Advanced Innovation Center for Intelligent Robots and Systems (2018IRS11), and PEK-SenseTime
Joint Laboratory of Machine Vision.

%
%
%
\bibliographystyle{splncs04} 
\bibliography{mybibliography}
\end{document}


\title{Supplemental Material to\\ 
	JVLDLoc: a Joint Optimization of 
\\ Visual-LiDAR Constraints and
Direction Priors for Localization in Driving Scenario}
\titlerunning{Supplemental Material of JVLDLoc}
%
\author{Longrui Dong  \and 
Gang Zeng\textsuperscript{(\Letter)}  }
\authorrunning{L. Dong and G. Zeng}
%
\institute{Key Lab. of Machine Perception (MoE), \\
School of Intelligence Science and Technology, \\
Beijing 100871, China \\
\email{\{lrdong,zeng\}@pku.edu.cn}}
\maketitle              

\section{Supplement to Method}

\subsection{LiDAR Odometry in Tracking}
Our JVLDLoc contains a LiDAR odometry in tracking in case an input without prior poses is encountered.
It is modified from the approach in IMLS-SLAM\cite{deschaud2018imls}, 
which runs scan-to-model matching with Implicit Moving Least Square(IMLS) surface representation.
For each input point cloud frame, $\textbf{N}$ last scans are kept as local map.
The technique details can be found in \cite{deschaud2018imls} except the following main differences:
\begin{itemize} 
  \item For each input scan, we discard points of which distance to local sensor's origin is less than 5m. 
  \item Before small object removal, We apply RadiusOutlierRemoval to pointcloud and extract ground plane by SACSegmentation\cite{Rusu_ICRA2011_PCL} instead of voxel growing from original paper. 
\end{itemize}

After the above LiDAR odometry, we send initial poses to do visual feature association and direction clustering.

\subsection{Scan to Map Association in Local Mapping}
In Section 4.2 of our submmited paper, we only construct IMLS constraint between current keyframe $\mathcal{K}_{n}$ and its point cloud map at Step 4).
To build enough dense 3D data associations in current $\mathbf{K}_{local}$, we take out all previous saved correspondences (see gray dotted line in Fig.3 (a) of submmited paper)
that $\textbf{p}_x^j$, $^{c}\textbf{p}_y^j$ and $\vec{n_c}^j$ are within $\mathbf{K}_{local}$:

\begin{equation}
	\textbf{e}_{L_{old}}(m, j) = \| (T_{wc}^{-1}T_{wm}\textbf{p}_x^j - ^{c}\textbf{p}_y^j ) \cdot \vec{n_c}^j\|^{2}
	\label{eq:oldicp}
\end{equation} 

where $(\textbf{p}_x^j, ^{c}\textbf{p}_y^j, \vec{n_c}^j) \in \mathbf{L}$.
Apparently above equation is nearly the same as Eq.(3) in main paper except that $m<n$.
In this way, we can give the poses of the keyframes in $\mathbf{K}_{local}$ more constraints from the point cloud data. 

Note that the pipline constructing point to IMLS surface residual (including Eq.(2), Eq.(3) in main paper) are all borrowed from \cite{deschaud2018imls} 
and modified for integration into our local mapping module.

\subsection{Loop Closing} %
\label{subsec:loopclosing}
To weaken cumulative localization error, we also detect loop candidate for each keyframe\cite{mur2017orb}.
Specific loop candidates detection procedures are the same as \cite{mur2017orb}, which is based on visual bags of words place recognition.
Once we estimates the correction transformation $\hat{T}_{nw}$ of current accepted keyframe $\mathcal{K}_n$, this correction is distributed over all keyframes
from $\mathcal{K}_l$ to $\mathcal{K}_n$.
The pseudocode of distributing \textit{SE}(3) error, inspired by \cite{yin2020caelo}, is in \cref{alg:distriberr}.
To simplify the description, we use $T_{i}$ to represent ${T}_{iw}$ in the following algorithm flow, where $i \in [i,n)$.
\begin{algorithm} 
\caption{Distribute \textit{SE}(3) error over keyframes}\label{alg:distriberr}
\begin{algorithmic}[1]
	\Require{$T = \{T_l,T_{l+1},...,T_{n}\}$, $\hat{T}_{n}$}
	\Ensure{$T_{correted} = \{\hat{T}_l,\hat{T}_{l+1},...,\hat{T}_{n}\}$} \Comment{$\hat{(\cdot)}$ means corrected pose, $T_i$ refers to pose of keyframe $\mathcal{K}_i$}
	\State $\Delta T_n \gets \hat{T}_{n}T_{n}^{-1}$ 
	\Comment{the correction transformation for $\mathcal{K}_n$}
	\State $\Delta R_n, \Delta t_n \gets  Rot\_part(\Delta T_n), Tran\_part(\Delta T_n)$
	\State $\Delta \phi_n \gets \textbf{log} (\Delta R_n) $ \Comment{Logarithmic map of \textit{SO}(3) Lie group}
	\State $i \gets (l+1)$
	\While{$i<n$}
	\State $\Delta t_n^i \gets \frac{i-l}{n-l}\Delta t_n$
	\State $\Delta \phi_n^i \gets \frac{i-l}{n-l}\Delta \phi_n$
	\State $\Delta R_n^i \gets \textbf{Exp}(\phi_n^i)$ \Comment{Exponential map of \textit{so}(3) Lie algebra}
	\State $\Delta T_n^i \gets (\Delta R_n^i, \Delta t_n^i)$
	\State $\hat{T}_i \gets \Delta T_n^i T_i$
	\EndWhile\label{kfwhile}
\end{algorithmic}
\end{algorithm}

\subsection{Association of Direction Priors and VPs}
\label{subsec:makedirection}

The direction of a line, not its position, determines the vanishing point\cite{andrew2001multiple}.
So only the rotation part of provided prior poses are utilized to generate global direction at world coordinate.
Detailed steps are as follows:
\begin{itemize}
  \item [1)] Detect at most 3 vanishing points from each (left) image.
  \item [2)] Denoting obtained image plane coordinates of vanishing points as $\textbf{v}$.
  Then we use prior orientation $\textbf{\textrm{R}}_{wc}$ from other odometry to obtain corresponding direction at world coordinates: 
  $\textbf{d} = \textbf{\textrm{R}}_{wc}\textbf{K}^{-1}\textbf{v} / \| \textbf{\textrm{R}}_{wc}\textbf{K}^{-1}\textbf{v} \|$,
  where unit direction $\textbf{d} \in \mathbb{R}^3 $.
  \item [3)] Run bi-level mean-shift clustering(see \cref{alg:directioncluster}) to all 3D vanishing directions from last step.
\end{itemize}
It is worth mentioning that these direction priors are mainly distributed in the approximate straight path of the trajectory. 

\begin{algorithm}
	\caption{Bi-level Mean-shift clustering for 3D vanishing directions}\label{alg:directioncluster}
	\begin{algorithmic}[1]
	  \Require{All raw frame index-direction pairs: $\mathbf{PD} = \{ (f_i, \textbf{d}_i)\}$, 2D translations from prior poses: $\mathbf{t}=\{t_i\}$, where $i\in|\mathbf{PD}|$}
	  \Ensure{Supporting directions $\hat{\textbf{d}}_w$ represented by Cluster centers and corresponding frame index}
	  \State $i \gets 0$
	  \While{$i<|\mathbf{PD}|$}
		\State $ \textit{fd}_i \gets \Delta t_i$  \Comment{Ground plane projection of the forward direction,$\textbf{d}_i^{proj}$ is the same}
		\If{$\arccos (\textbf{d}_i^{proj}\cdot{\textit{fd}_i}/\| \textbf{d}_i^{proj}\cdot{\textit{fd}_i} \|) > Th_{f} $}
		  \State discard $(f_i, \textbf{d}_i)$
		\EndIf 
		\State $i=i+1$
	  \EndWhile \Comment{new set : $\mathbf{PD}_1$}
	  \State $\textbf{d} \gets \{(\textbf{d}_i)| \textbf{d}_i \in \mathbf{PD}_1\}$
	  \State $(^{m}\textbf{d}, \mathbf{cluster\_id}) \gets \textbf{Mean\_shift}(\textbf{d}, bandwith1 )$ \Comment{Raw cluster center and cluster IDs for each $\textbf{d}_i \in \textbf{d}$}
	  \While{$i<|^{m}\textbf{d}| $}
		\State $\delta_i \gets $  standard deviation of each cluster
		\State $s_i \gets $  slope of each cluster center's projection $^{m}\textbf{d}_i^{proj}$
		\State insert $\delta_i, s_i$ into set $\Delta, \mathbf{s}$
		\State $i=i+1$
		\EndWhile
	  \State sort all clusters by slopes and devide into $n$ subsets
	  \State $i \gets 0, \mathbf{clustd} \gets \{ \} $
	  \While{$i<n$}
		\State sort all clusters in subset $i$ by size
		\State keep a cluster $^{j}cd$ in $\mathbf{clustd}$ when it meets:
		\State        a) $\delta_j \leq Th_{\delta}$
		\State        b) size of $^{j}cd \geq Th_{n}$
		\State $i=i+1$
	  \EndWhile
	  \State $\mathbf{clustd}_1 \gets \{ \} $
	  \State Keep $Th_{m}$ clusters into $\mathbf{clustd}_1$ every $Th_{l}$ m on the map
	  \State $\mathbf{clustd}_2 \gets \{ \} $
	  \State $j \gets 0$
	  \While{$j<|\mathbf{clustd}_1|$}
	  \State for a cluster $^{j}cd$ in $\mathbf{clustd}_1$, take out 2D translations of all samples in $^{j}cd$: $\mathbf{t}_j$
	  \State $(^{q}\textbf{t}_j, \mathbf{cluster\_id}) \gets \textbf{Mean\_shift}(\textbf{t}_j, bandwith2 )$ \Comment{ Divide a cluster $^{j}cd$ into q sub-clusters}
	  \State save corresponding direction cluster in $\mathbf{clustd}_2$
	  \State $j=j+1$
	  \EndWhile
	\end{algorithmic}
\end{algorithm}

\section{Supplement to Jacobians}
\label{sec:jacob} 
Visual reprojection error has already been applied widely in ORB-SLAM\cite{campos2021orb,mur2015orb,mur2017orb}, relevant codes are public at \footnote{\url{https://github.com/raulmur/ORB_SLAM2}}, 
so we only explain point to IMLS surface error and direction projection error.

\subsection{Point to IMLS surface error}
\label{subsec:imlserr}
In order to facilitate the following derivation of the formulas and jacobians, let:
\begin{equation}
    T_{nw} = \\
	\begin{bmatrix}
		\textrm{R}_{nw}	& \textrm{t}_{nw} \\
		\textbf{0}      & 1
	\end{bmatrix}
    \label{eq:fullSE3}
\end{equation} 
where $n$ is the index of $\mathcal{K}_n$. At Step 3) in Sec. 4.2, and as is shown in Fig. 3(a) in main paper, 
we have denoted $\mathcal{K}_c$ as the keyframe to which the closest point of $p_x^{j}$ belongs.
And $\;^{c}(\cdot )$ means that $(\cdot )$ is at $\mathcal{K}_c$ 's coordinate.

There is no ready-made class for point to IMLS surface error in $g^{2}o$\cite{kummerle2011g}, 
so we modified the raw Generalized ICP (GICP) \cite{segal2009generalized} class 
According to \cite{segal2009generalized}, different application of GICP depends on the information matrix. 
So here we further derive Eq.(3) in our paper: 

\begin{equation}
	\textbf{e}_{L}(n, j) = \| (T_{cw}T_{nw}^{-1}\textbf{p}_x^j - ^{c}\textbf{p}_y^j ) \cdot \vec{n_c}^j\|^{2}
	\label{eq:imlsicp}
\end{equation}


\begin{equation}
	\Longleftrightarrow  \\
	e_{GICP} = \;^{c}d_{j}^{T}\cdot \textbf{Q}_{j}^{-1}\cdot \;^{c}d_{j}
    \label{eq:togicp}
\end{equation}
where: 


\begin{align}
	^{c}d_{j} &= T_{cw}T_{nw}^{-1}\textbf{p}_x^j - ^{c}\textbf{p}_y^j  
	\label{eq:cdj}
\end{align}

\begin{equation}
	\textbf{Q}_{j} = \textrm{R}_{\vec{n_c}}^{T}\cdot
	\begin{bmatrix}
		\epsilon & 0 & 0 \\
		0 & \epsilon  & 0 \\
		0 & 0   & 1 
	\end{bmatrix} \cdot \textrm{R}_{\vec{n_c}} \label{eq:Q}
\end{equation}


$\textrm{R}_{\vec{n_c}}$ is rotation matrix relevant to normal $\vec{n_c}$ (See \cite{segal2009generalized} for details).
$\epsilon$ is a small constant much less than 1, we set it to 0 in the code such that \cref{eq:togicp} totally equivalent to \cref{eq:imlsicp}. 


Jacobian matrix of \cref{eq:cdj} with respect to the pose of the two keyframes are: 


\begin{align}
    \textrm{J}_c &= \frac{\partial \;^{c}d_{j}}{\partial T_{cw}} \nonumber \\
				 &= \frac{\partial (T_{cw}\,T_{nw}^{-1} p_x^j - \;^{c}p_y^j)}{\partial T_{cw}} \nonumber \\
				 &= \frac{\partial (T_{cw}\,T_{nw}^{-1} p_x^j)}{\partial T_{cw}} \nonumber \\
				 &= \begin{bmatrix}
					-\;(T_{cw}\,T_{nw}^{-1} p_x^j)^{\wedge} & \textbf{I} 
					\end{bmatrix} \nonumber \\ 
				 &= \begin{bmatrix}
					-\;(^{c}p_{x}^{j})^{\wedge} & \textbf{I} 
					\end{bmatrix} 
	\label{eq:jacobT0}
\end{align} 

\begin{align}
    \textrm{J}_n &= \frac{\partial \;^{c}d_{j}}{\partial T_{nw}} \nonumber \\
				 &= \frac{\partial (T_{cw}\,T_{nw}^{-1} p_x^j - \;^{c}p_y^j)}{\partial T_{nw}} \nonumber \\
				 &= \frac{\partial (T_{cw}\,T_{nw}^{-1} p_x^j)}{\partial T_{nw}} \nonumber \\
				 &= \frac{\partial (T_{cw}\;^{w}p_x^j)}{\partial ^{w}p_x^j} \ast \frac{\partial ^{w}p_x^j}{\partial T_{nw}} \nonumber \\
				 &= \textrm{R}_{cw} \ast \begin{bmatrix}
					\textrm{R}_{nw}^{T}(p_{x}^{j})^{\wedge} & -\textrm{R}_{nw}^{T} 
					\end{bmatrix}  \nonumber \\
				 &= \begin{bmatrix}
					\textrm{R}_{cw}\textrm{R}_{nw}^{T}(p_{x}^{j})^{\wedge} & -\textrm{R}_{cw}\textrm{R}_{nw}^{T} 
					\end{bmatrix}  \nonumber \\ 
				 &= \begin{bmatrix}
					\textrm{R}_{cn}(p_{x}^{j})^{\wedge} & -\textrm{R}_{cn}
					\end{bmatrix} 
	\label{eq:jacobT1}
\end{align} 

These jacobians are on \textrm{SE}(3) manifold, and is $3\times6$. 
The first three colomns are gradients of the rotation and the last three colomns belong to the translation.


\subsection{Direction projection error}
\label{subsec:directionerr}

We modified the original \textit{EdgeSE3ProjectXYZ} in $g^{2}o$ to implemente our proposed direction projection error.
Recall the error function : $\textbf{e}_D(n) = \| \textbf{v} - \pi([\textbf{R}_{nw}|\textbf{I}]\hat{\textbf{D}}_w) \|^{2}_{\sum}$.
For the convenience of notation, we rewrite this formular:

\begin{align}
	\textbf{e}_D(n)  &=  \| \textbf{v} - d_{wz}^{-1}\textbf{K}\textbf{R}_{nw}\hat{\textbf{d}}_w \|^{2}_{\sum} \nonumber \\ 
	&= \| \textbf{v} - d_{nz}^{-1}\textbf{K}\hat{\textbf{d}}_n \|^{2}_{\sum} 
	\label{eq:newdpe}
\end{align} 

where:
\begin{equation}
	\hat{\textbf{d}}_n = \textbf{R}_{nw}\hat{\textbf{d}}_w = \begin{bmatrix}
		d_{nx} \\
		d_{ny} \\
		d_{nz}
	\end{bmatrix}
    \label{eq:directionlocal}
\end{equation} 
The jacobian with respect to $\textrm{R}_{cw}$ on
\textrm{SO}(3) manifold is a $3\times3$ matrix :

\begin{align}
    \textrm{J}_n &= \frac{\partial (\textbf{v} - d_{nz}^{-1}\textbf{K}\hat{\textbf{d}}_n)}{\partial \textbf{R}_{nw}} \nonumber \\
				 &= \frac{\partial (\textbf{v} - d_{nz}^{-1}\textbf{K}\hat{\textbf{d}}_n)}{\partial \hat{\textbf{d}}_n} \ast \frac{\partial \hat{\textbf{d}}_n}{\partial \textbf{R}_{nw}} \nonumber \\
				 &= \begin{bmatrix}
					-\frac{f_x}{d_{nz}} & 0 & \frac{f_{x}d_{nx}}{d_{nz}^2} \\
					0 & -\frac{f_y}{d_{nz}} & \frac{f_{y}d_{ny}}{d_{nz}^2} \\
					0&0&0	
					\end{bmatrix} \ast (-(\textbf{R}_{nw}\hat{\textbf{d}}_w)^{\wedge}) \nonumber \\
				 &= \begin{bmatrix}
					-\frac{f_x}{d_{nz}} & 0 & \frac{f_{x}d_{nx}}{d_{nz}^2} \\
					0 & -\frac{f_y}{d_{nz}} & \frac{f_{y}d_{ny}}{d_{nz}^2} \\
					0&0&0	
					\end{bmatrix} \ast (-(\hat{\textbf{d}}_n)^{\wedge}) \nonumber \\
				 &= \begin{bmatrix}
					-\frac{f_x}{d_{nz}} & 0 & \frac{f_{x}d_{nx}}{d_{nz}^2} \\
					0 & -\frac{f_y}{d_{nz}} & \frac{f_{y}d_{ny}}{d_{nz}^2}\\
					0&0&0	
					\end{bmatrix} \ast \begin{bmatrix}
						0 & d_{nz} & -d_{ny} \\
						-d_{nz} & 0 & d_{nx} \\
						d_{ny} & -d_{nx} & 0
						\end{bmatrix} 
					\nonumber \\ 
				&= \begin{bmatrix}
					\frac{f_x d_{nx} d_{ny}}{d_{nz}^{2}} & -f_x (1+\frac{d_{nx}^2}{d_{nz}^{2}}) & \frac{f_{x}d_{nx}}{d_{nz}} \\
					f_y (1+\frac{d_{ny}^2}{d_{nz}^2}) & -\frac{f_y d_{nx} d_{ny}}{d_{nz}^2} & -\frac{f_{y}d_{nx}}{d_{nz}}	\\
					0&0&0
				\end{bmatrix}
	\label{eq:jacobR}
\end{align} 
Where $(\cdot )^{\wedge}$ means skew-symmetric matrix of $(\cdot )$. 
It is obvious that the jacobians of global joint energy function is just weighted sum of all above jacobians.

\begin{table}[H] 
	\caption{RMSE(m) of APE for KITTI 00-10.
	For each sequence, we report RMSE of input pose, our JVLDLoc without direction priors and our full JVLDLoc respectively.
	As for LiDAR odometry, we test under different $\textbf{N}$ which is the number of scans kept as the local map.}	
	\label{tab:bettertoprior}
	\centering
			\begin{tabular}{l||c|c|c|c|c|c}
				\toprule
				{KITTI Odometry}& {00} & {01} & {02} & {03} & {04} & {05} \\
				\midrule
				\hline 
				\noalign{\smallskip}

				ORB-SLAM2 \cite{mur2017orb}    
				&1.354
				&10.16
				&6.899
				&0.693
				&0.178
				&0.783
				\\
				w.o. direction priors
				&1.367
				&3.289
				&4.669
				&0.614
				&0.169
				&0.705
				\\
				w. direction priors
				&1.205	  
				&1.999 
				&3.544 
				&0.530	  
				&0.151
				&0.593
				\\ 
				\hline
				\noalign{\smallskip}
				IMLS-SLAM \cite{deschaud2018imls}    
				&3.895	  
				&2.412
				&7.159 
				&0.650	  
				&0.180
				&1.789
				\\
				w.o. direction priors
				&1.687	  
				&2.032
				&2.769 
				&0.573	  
				&0.181
				&0.926
				\\ 
				w. direction priors
				&1.307	  
				&1.766 
				&2.622 
				&0.534	  
				&0.145
				&0.706
				\\ 
				\hline
				\noalign{\smallskip}
				LiDAR odometry $\textbf{N}$=5    
				&3.407	  
				&3.141
				&7.758 
				&0.600	  
				&0.217
				&2.165
				\\
				w.o. direction priors
				&1.724	  
				&2.901
				&3.482 
				&0.597	  
				&0.201
				&1.010
				\\ 
				w. direction priors
				&1.324	  
				&1.788 
				&2.360 
				&0.542	  
				&0.160
				&0.693
				\\ 
				\hline
				\noalign{\smallskip}
				LiDAR odometry $\textbf{N}$=1    
				&6.849	  
				&35.03
				&12.74
				&0.839	  
				&0.265
				&3.379
				\\
				w.o. direction priors
				&2.027	  
				&5.932
				&4.414
				&0.764	  
				&0.186
				&1.286
				\\
				w. direction priors
				&1.616	  
				&3.097 
				&3.015 
				&0.585	  
				&0.175
				&0.802
				\\ \bottomrule
				\noalign{\smallskip}
				{}& {06} & {07} & {08} & {09} & {10} & {AVG} \\
				\midrule
				\hline 
				\noalign{\smallskip}
				ORB-SLAM2 \cite{mur2017orb}    
				&0.714
				&0.536
				&3.542
				&1.643
				&1.156
				& 2.515
				\\
				w.o. direction priors
				&0.485
				&0.473
				&3.354
				&2.433
				&0.990
				& 1.686
				\\
				w. direction priors
				&0.300
				&0.441
				&2.862
				&1.498
				&0.818
				&1.267
				\\ 
				\hline
				\noalign{\smallskip}
				IMLS-SLAM \cite{deschaud2018imls}    
				&0.457
				&0.461
				&2.426
				&1.388
				&0.765
				&1.962
				\\
				w.o. direction priors
				&0.397
				&0.388
				&3.083
				&1.960
				&0.860
				&1.350
				\\ 
				w. direction priors
				&0.295
				&0.332
				&2.520
				&1.383
				&0.775
				&1.126
				\\ 
				\hline
				\noalign{\smallskip}
				LiDAR odometry $\textbf{N}$=5    
				&0.497
				&0.471
				&2.882
				&1.192
				&0.738
				&2.097
				\\
				w.o. direction priors
				&0.338
				&0.466
				&3.042
				&2.217
				&0.908
				&1.535
				\\ 
				w. direction priors
				&0.312
				&0.437
				&2.710
				&1.771
				&0.897
				&1.181
				\\ 
				\hline 
				\noalign{\smallskip}
				LiDAR odometry $\textbf{N}$=1    
				&2.109	  
				&0.769 
				&8.123 
				&3.917	  
				&1.084
				&6.828
				\\
				w.o. direction priors
				&0.374	  
				&0.945 
				&4.602 
				&3.347	  
				&1.065
				&2.268
				\\
				w. direction priors
				& 0.302
				& 0.521
				& 3.131
				&1.828
				&0.933
				&1.455
				\\ \bottomrule
			\end{tabular}
\end{table}

\sectionBefore
\section{Supplement to Experiments}
\label{sec:experiment}
\sectionAfter

Limited by the number of pages of formal submission, we put the remaining figures and tables of experiments here. 
The following content is still divided like the corresponding part of our main paper.


\begin{figure}[t]
	\centering
	\begin{subfigure}{0.49\textwidth}
		\centering
		\includegraphics[width=\textwidth, interpolate]{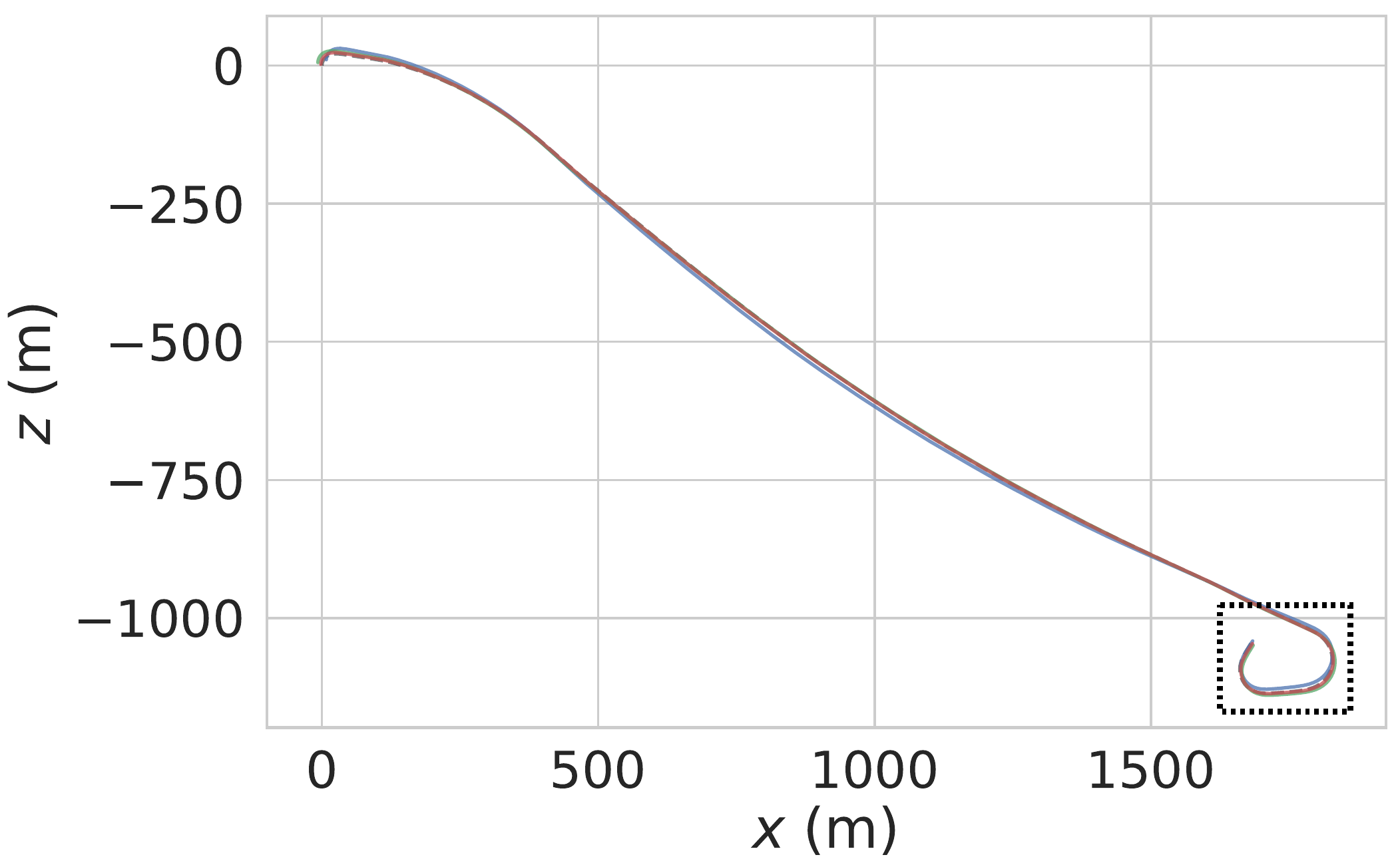}
	\end{subfigure}
	\begin{subfigure}{0.49\textwidth}
		\includegraphics[width=\textwidth, interpolate]{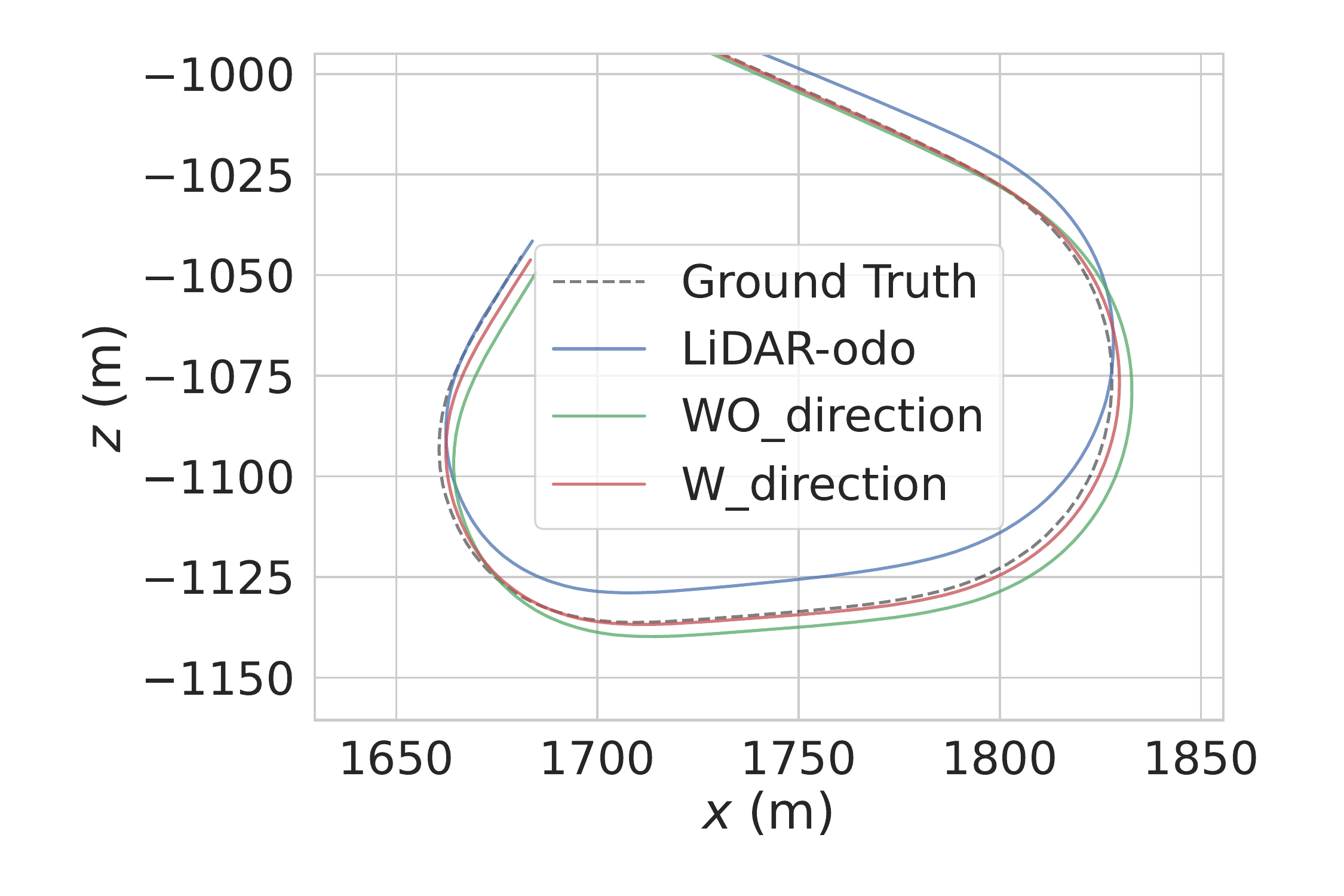}
	\end{subfigure}
	\caption{Trajectory of JVLDLoc(red), JVLDLoc without direction priors(green), LiDAR odometry(blue) tested on KITTI-01.
	 The right image is the enlargement of the dotted box on the left image}
	\label{fig:trajcompwod1}
\end{figure}

\subsectionBefore
\subsection{Improvements over Prior Map} 
\subsectionAfter
We have set another two baselines as input prior poses including IMLS-SLAM\cite{deschaud2018imls} and our LiDAR odometry.
The RMSE of trajectory are listed in \cref{tab:bettertoprior}, which is the full version of Table 1 in main paper.
Regardless of the quality of those four prior map, JVLDLoc all get lower drift on the average RMSE of KITTI 00-10.

\begin{figure}[t]
	\centering
	\begin{subfigure}{0.48\textwidth} 
		\centering
		\includegraphics[width=\textwidth, interpolate]{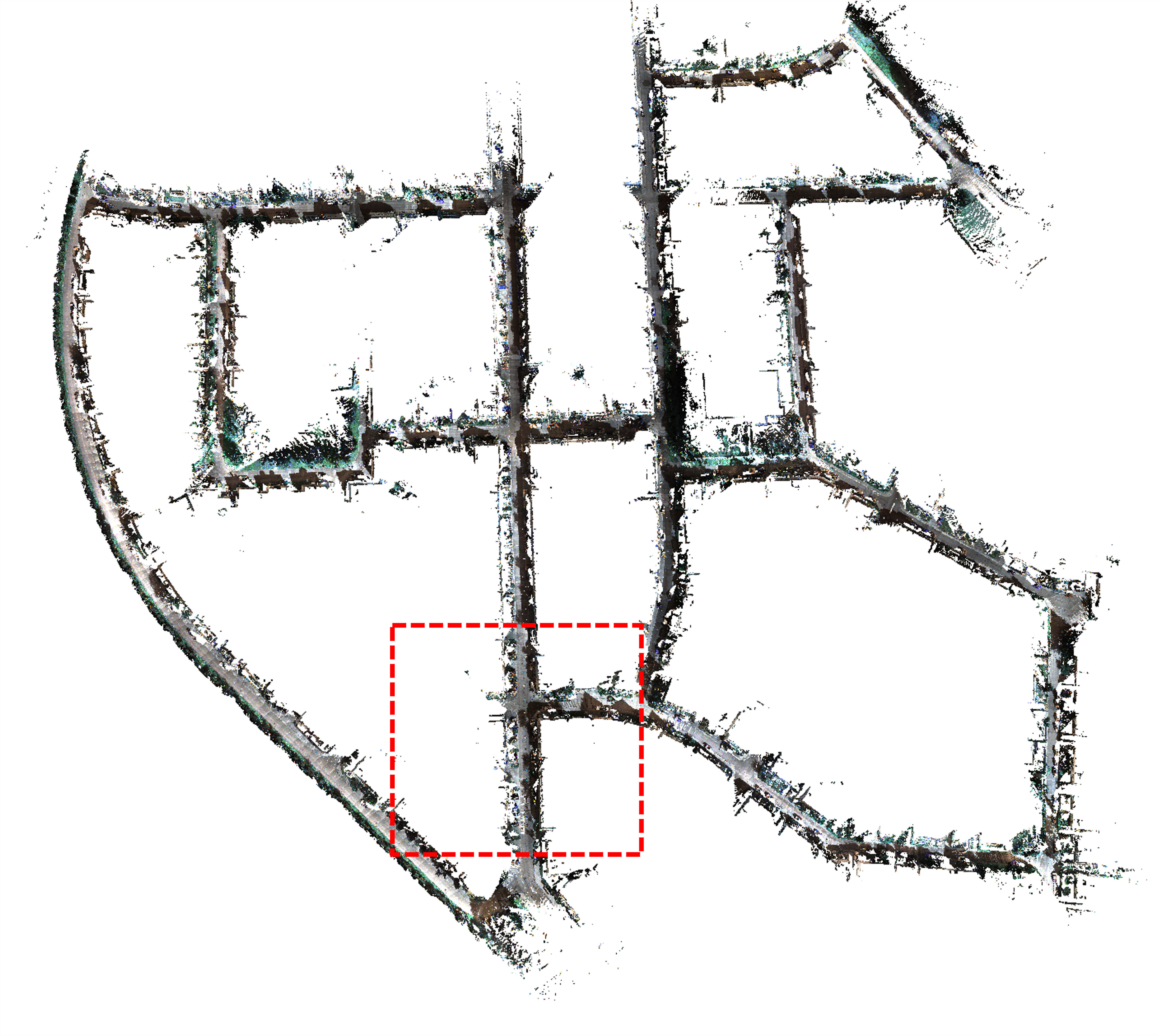}
		\caption{KITTI 00}
	\end{subfigure}
	\begin{subfigure}{0.48\textwidth} 
		\includegraphics[width=\textwidth, interpolate]{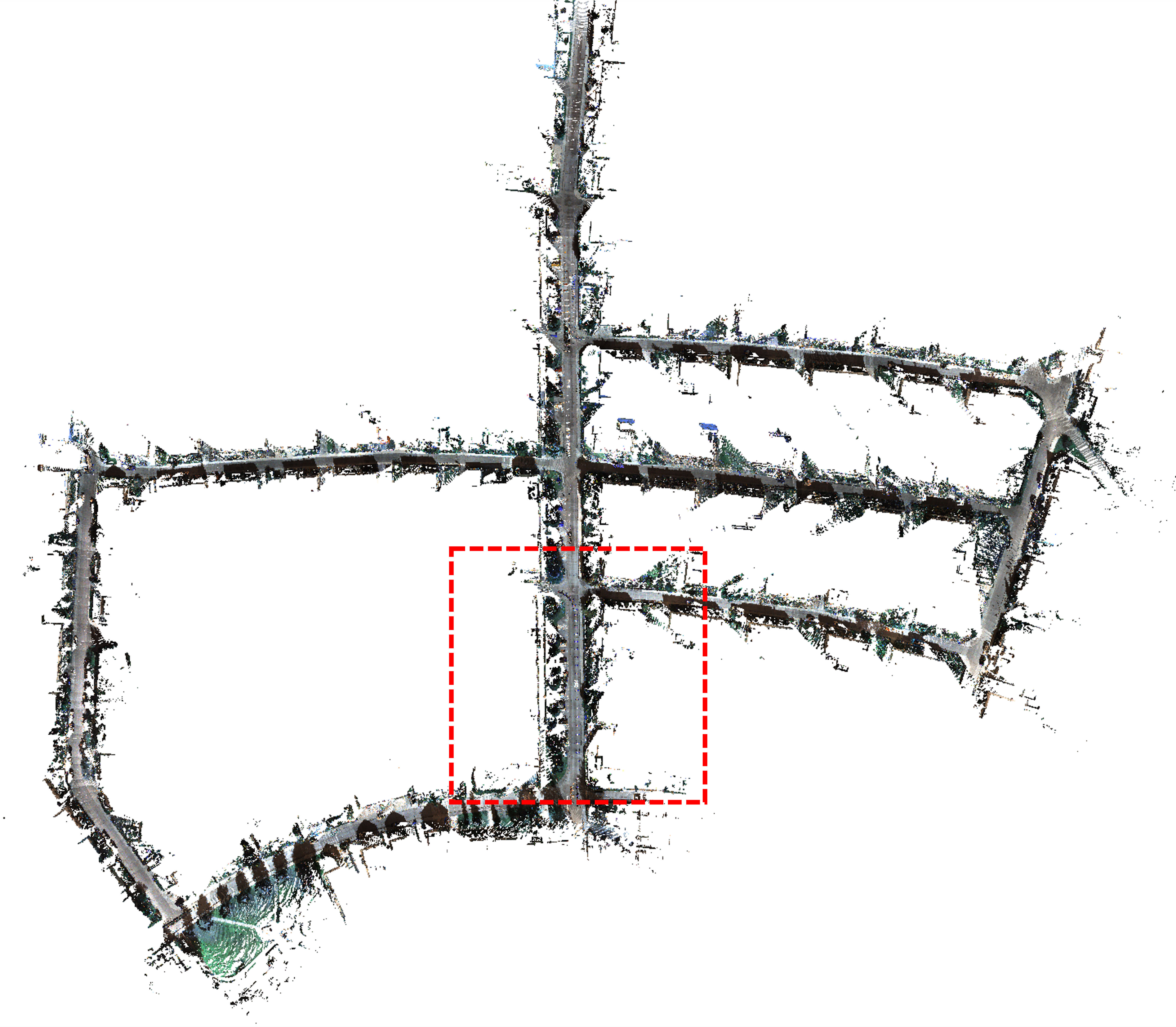}
		\caption{KITTI 05}
	\end{subfigure}
	\begin{subfigure}{0.48\textwidth} 
		\centering
		\includegraphics[width=\textwidth, interpolate]{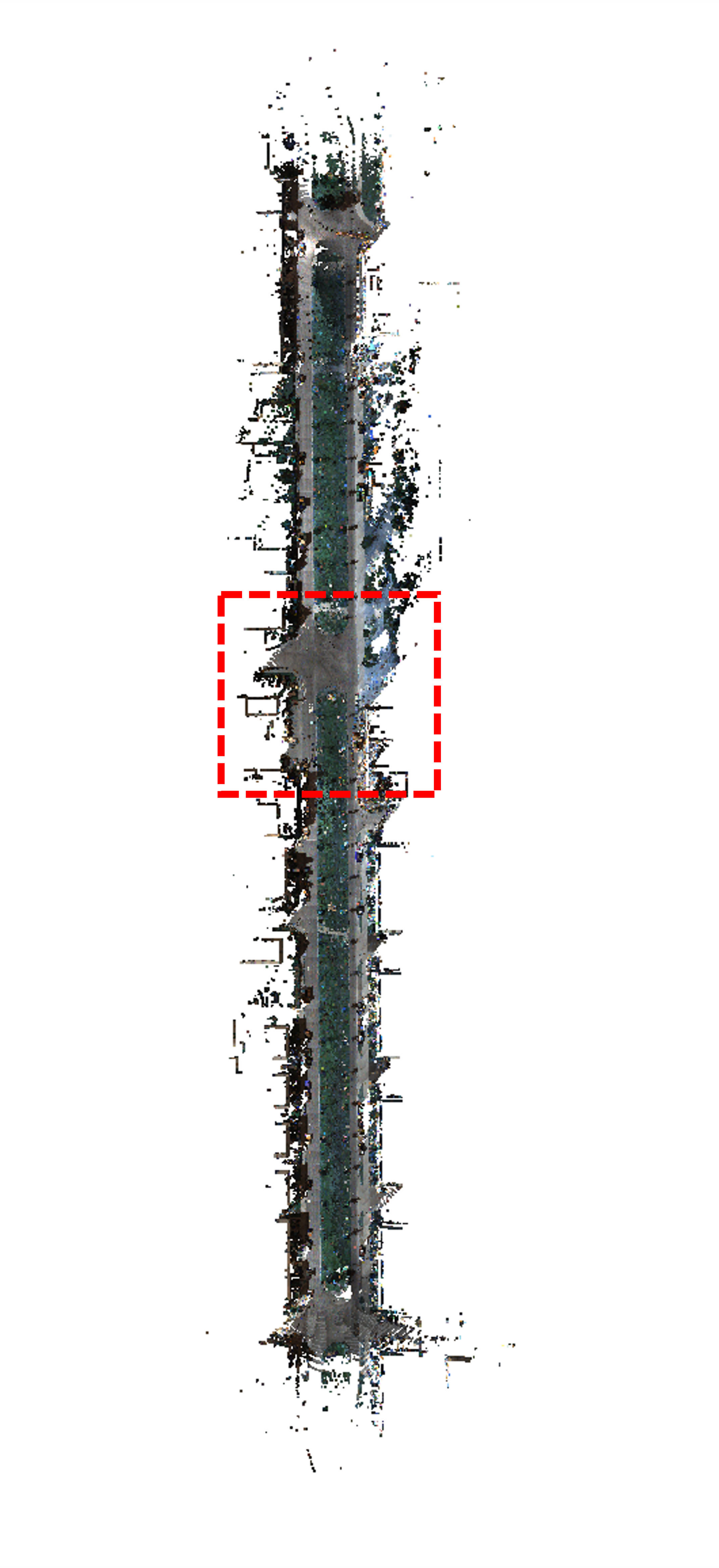}
		\caption{KITTI 06}
	\end{subfigure}
	\begin{subfigure}{0.48\textwidth} 
		\includegraphics[width=\textwidth, interpolate]{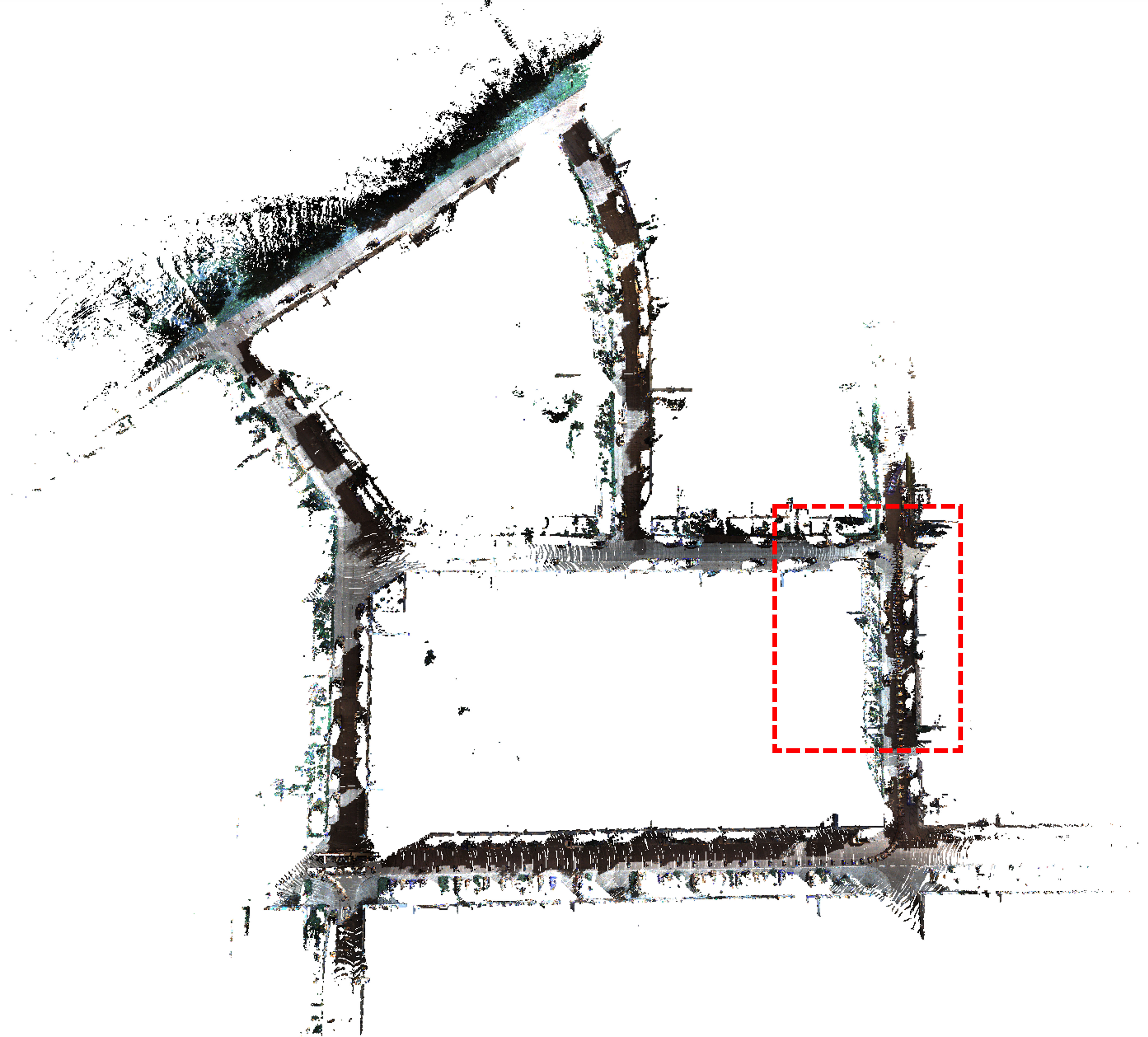}
		\caption{KITTI 07}
	\end{subfigure}
	\caption{
	Mapping result of JVLDLoc on KITTI sequences.
	}
	\label{fig:lidarmapBEVbox}
\end{figure}

\begin{figure}[t]
	\centering
	\begin{subfigure}{0.86\textwidth} 
		\includegraphics[width=\textwidth, interpolate]{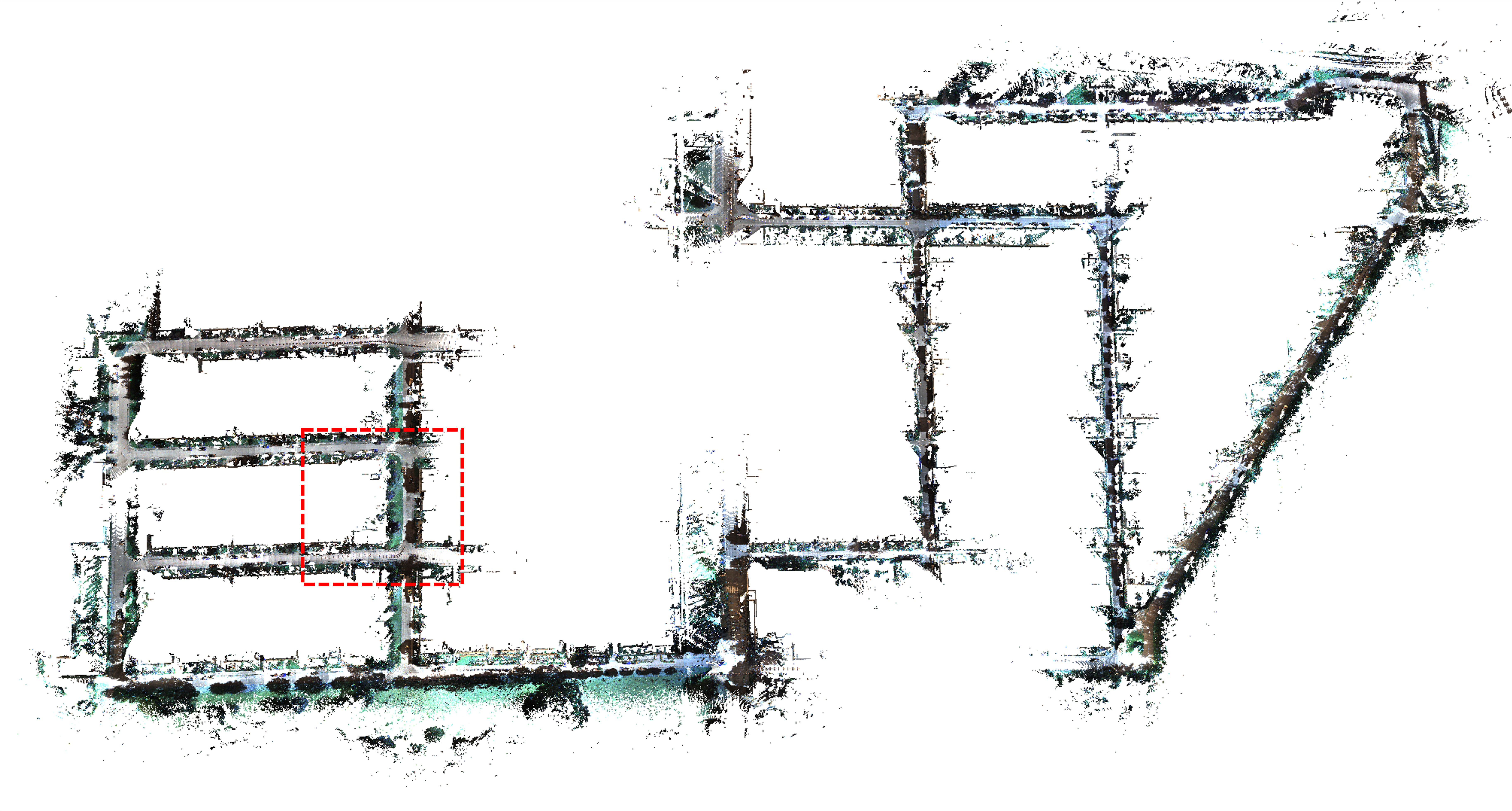}
	\end{subfigure}
	\caption{
	Mapping result of JVLDLoc on KITTI 08.
	}
	\label{fig:lidarmapBEVbox2}
\end{figure}

\begin{figure}[t]
	\centering
	\begin{subfigure}{0.32\textwidth} 
		\includegraphics[width=\textwidth, interpolate]{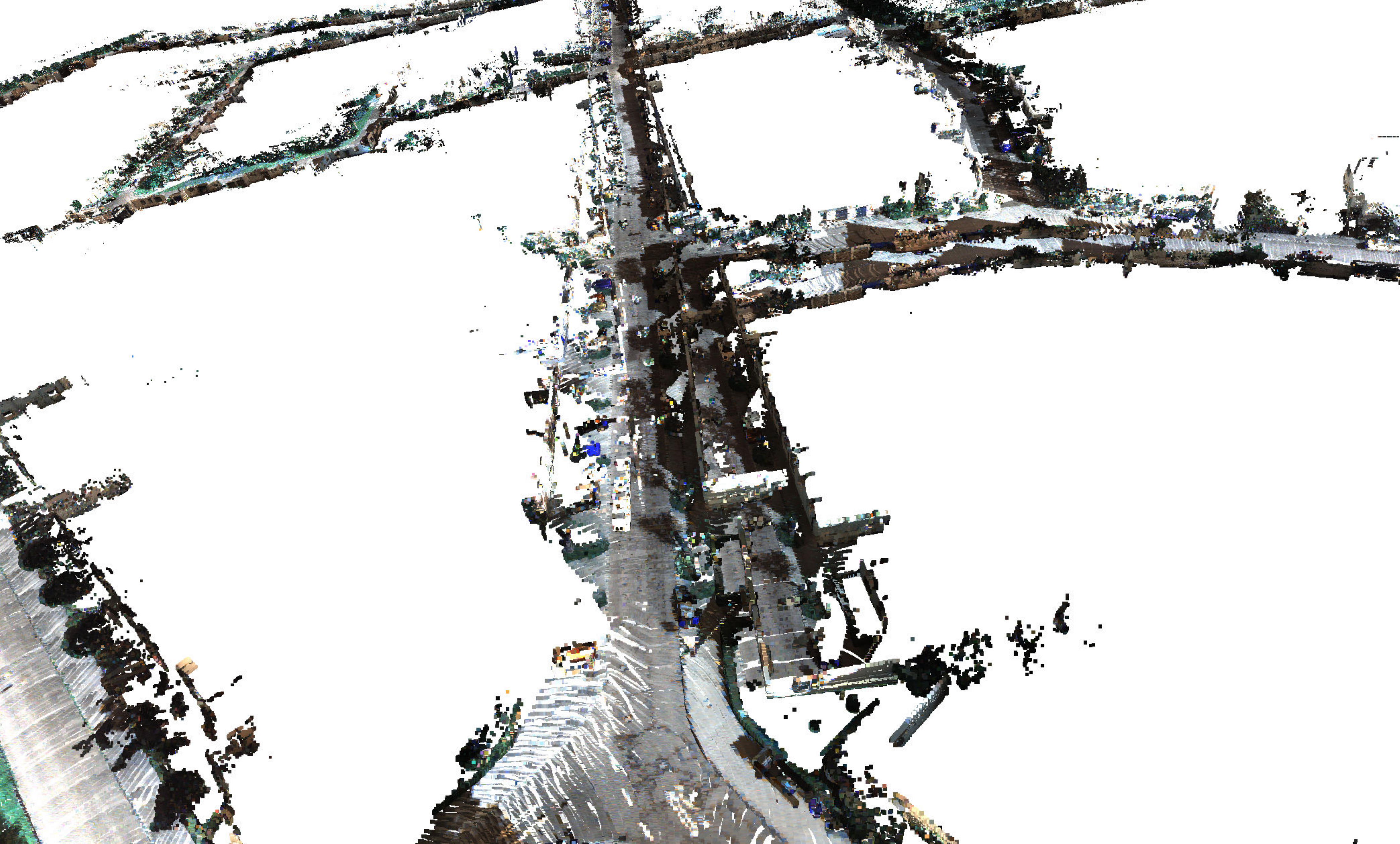}
	\end{subfigure}
	\begin{subfigure}{0.32\textwidth}
		\includegraphics[width=\textwidth, interpolate]{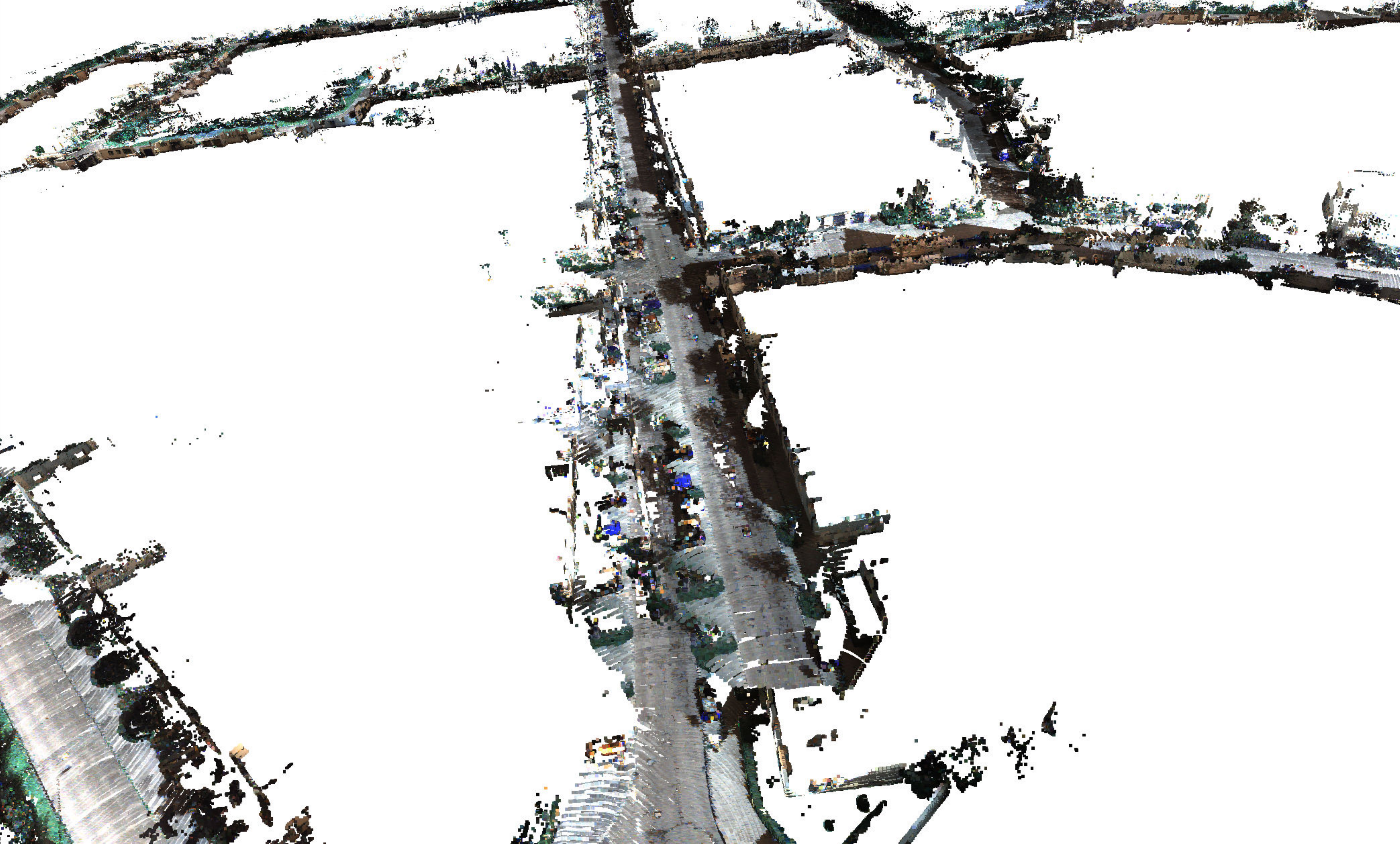}
	\end{subfigure}
	\begin{subfigure}{0.32\textwidth} 
		\centering
		\includegraphics[width=\textwidth, interpolate]{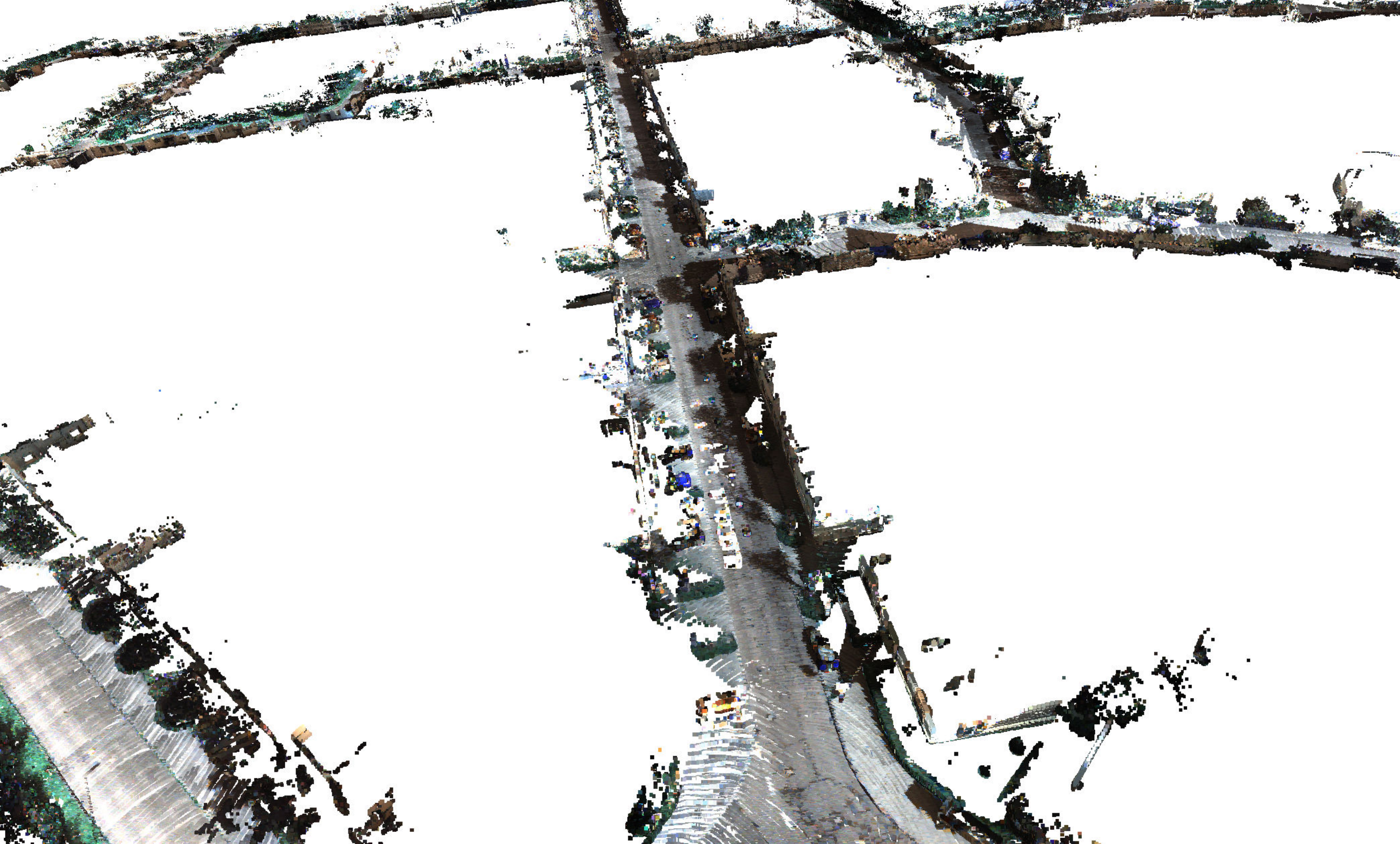}
	\end{subfigure}
	\begin{subfigure}{0.32\textwidth} 
		\centering
		\includegraphics[width=\textwidth, interpolate]{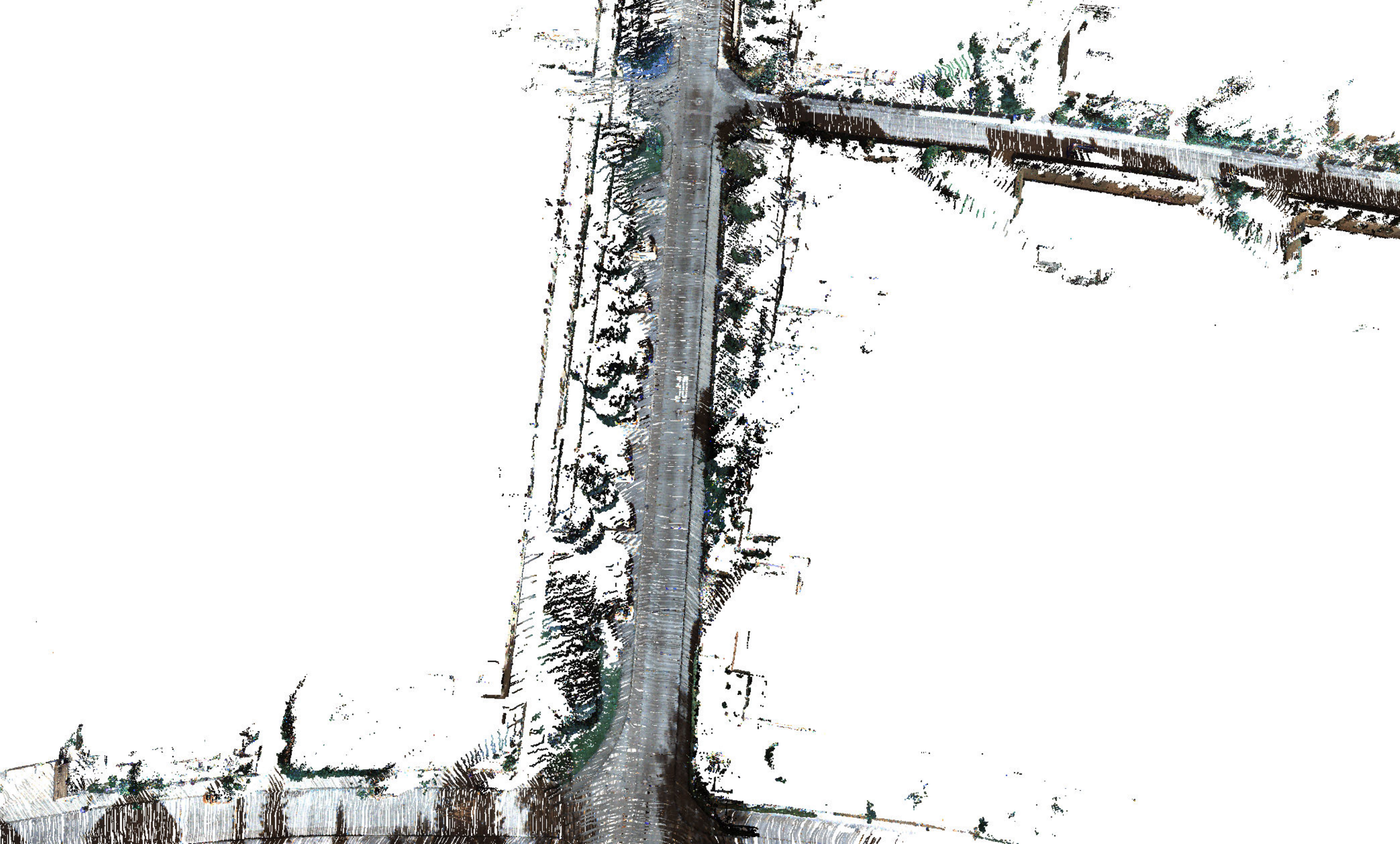}
	\end{subfigure}
	\begin{subfigure}{0.32\textwidth} 
		\includegraphics[width=\textwidth, interpolate]{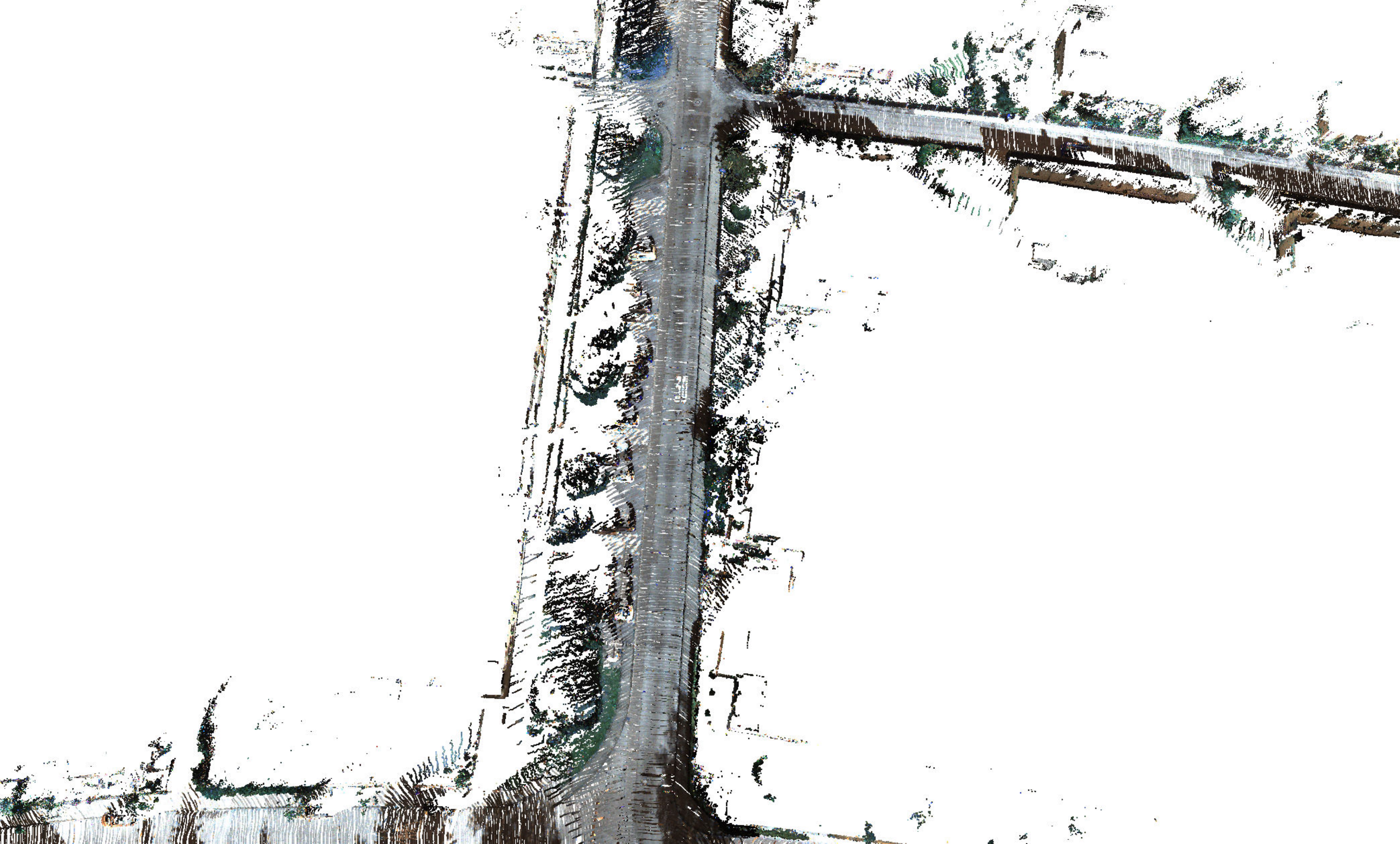}
	\end{subfigure}
	\begin{subfigure}{0.32\textwidth}
		\includegraphics[width=\textwidth, interpolate]{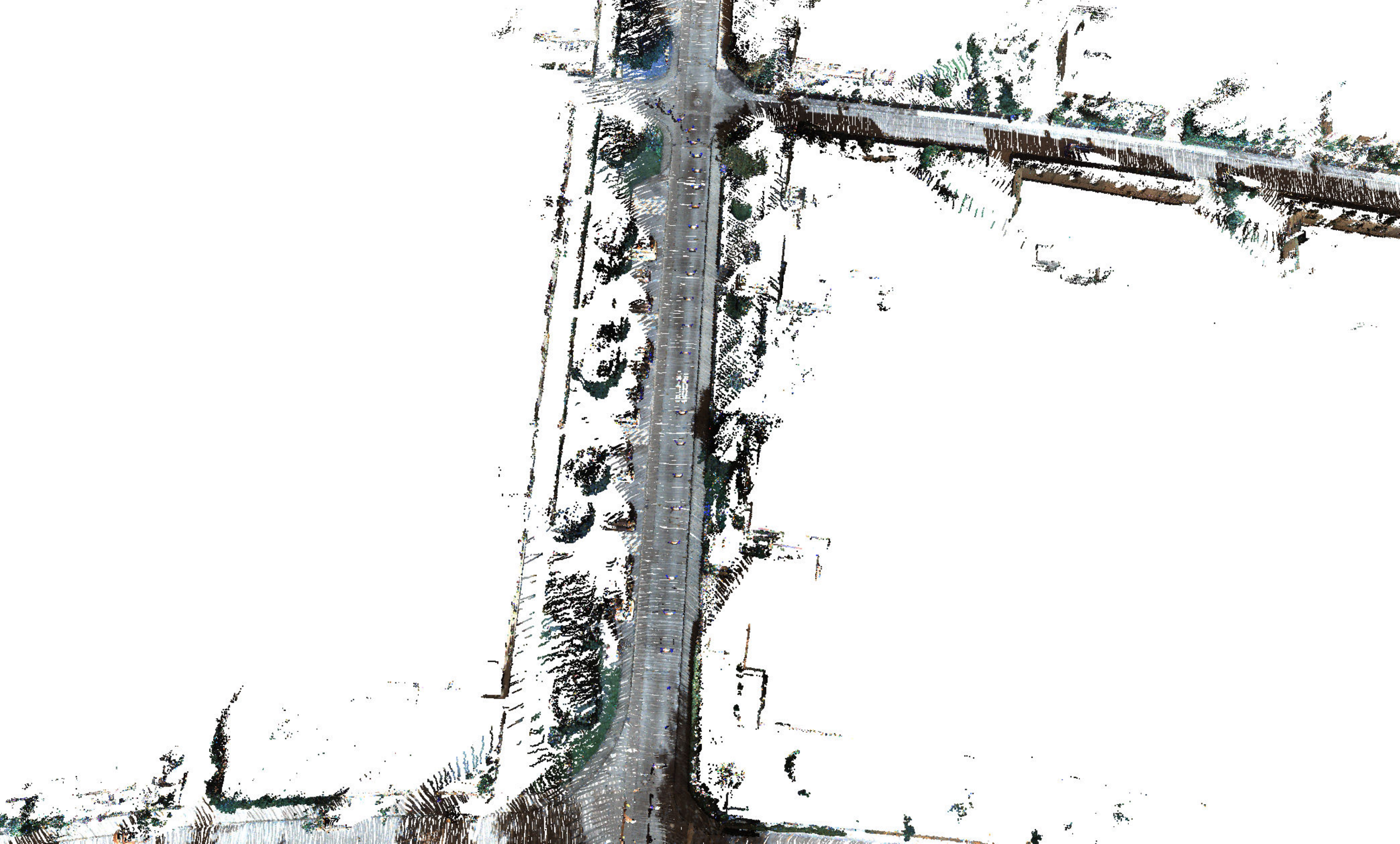}
	\end{subfigure}
	\begin{subfigure}{0.32\textwidth} 
		\centering
		\includegraphics[width=\textwidth, interpolate]{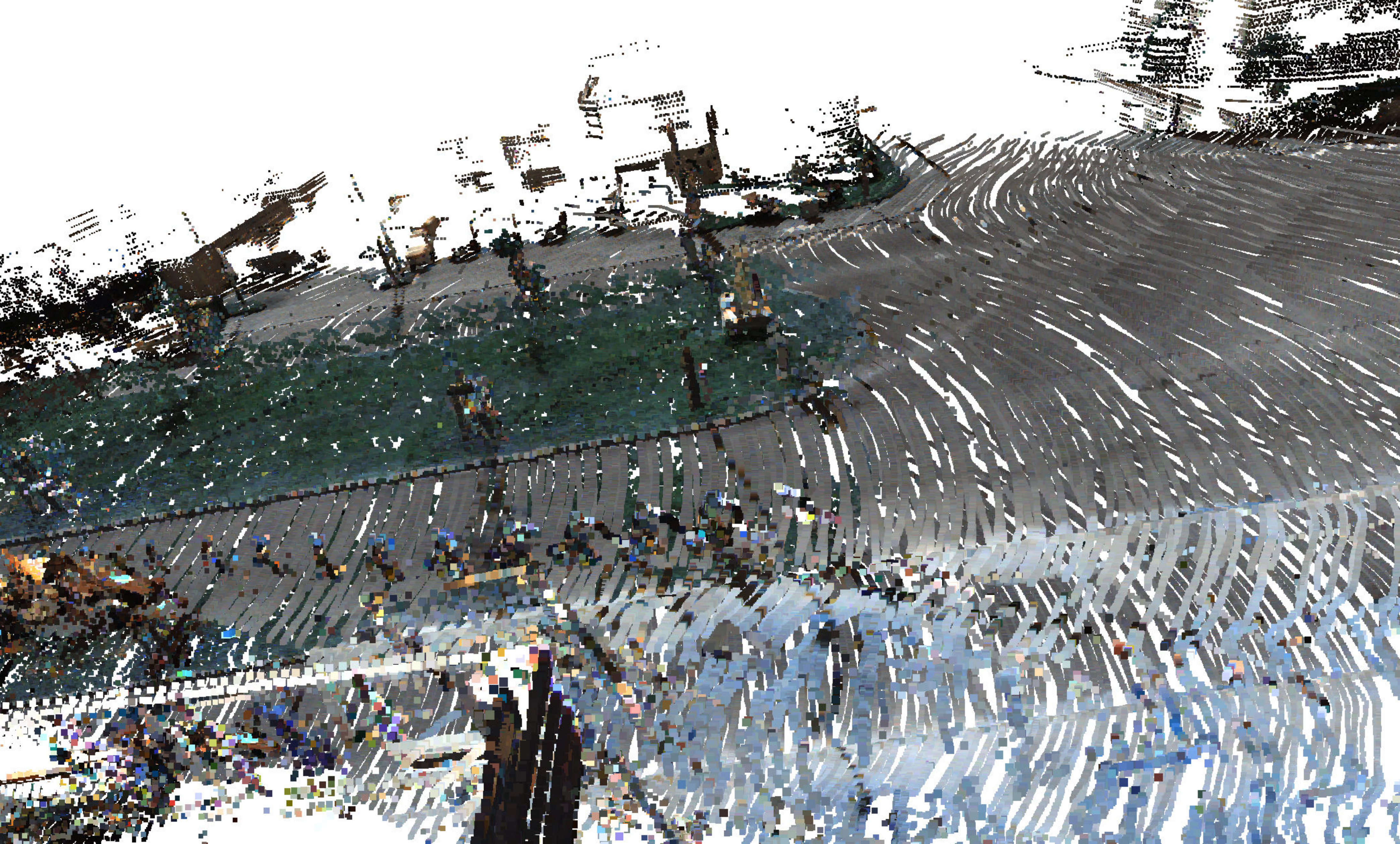}
		\caption{LiDAR odometry $\textbf{N}=1$}
	\end{subfigure}
	\begin{subfigure}{0.32\textwidth} 
		\centering
		\includegraphics[width=\textwidth, interpolate]{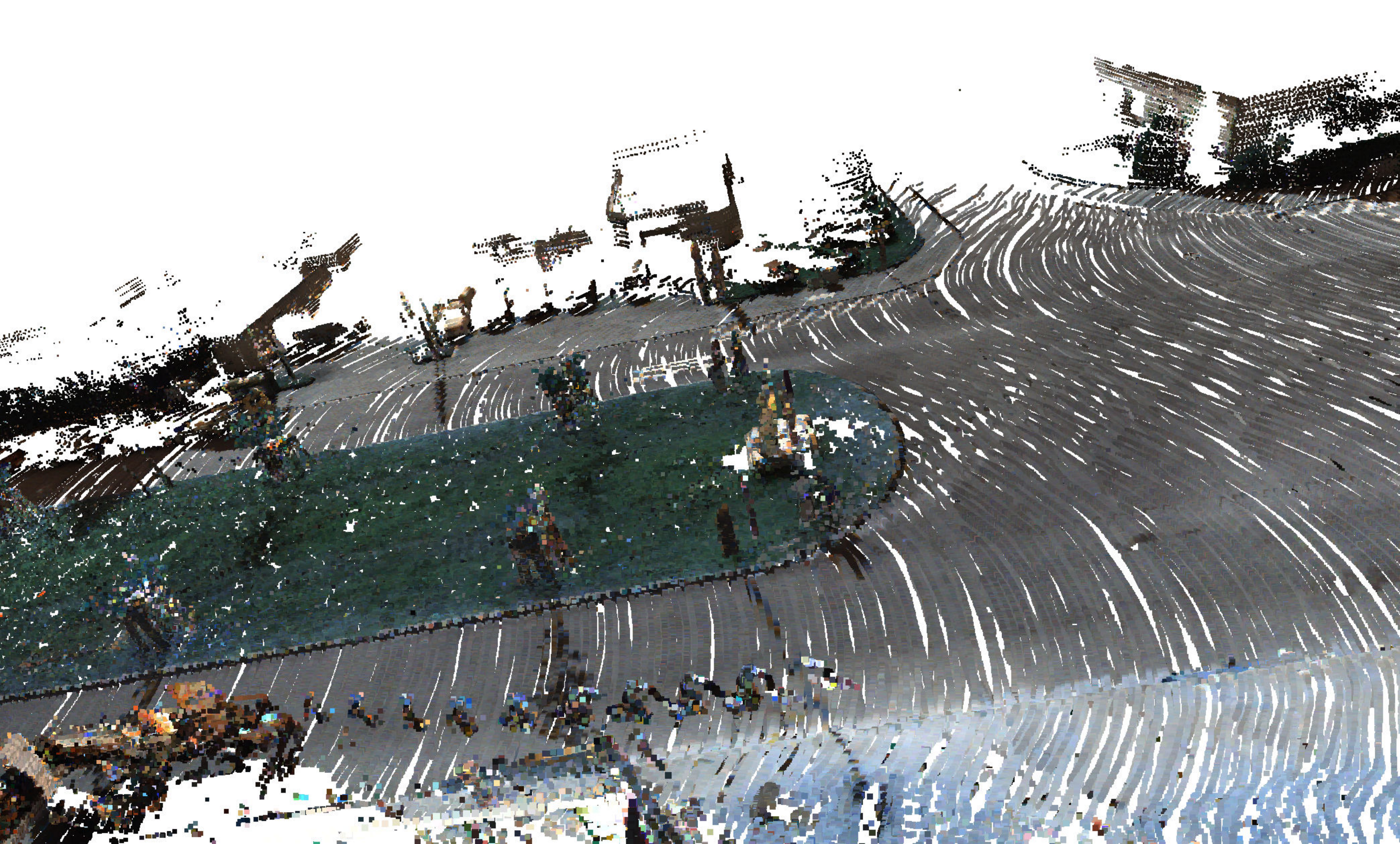}
		\caption{IMLS-SLAM}
	\end{subfigure}
	\begin{subfigure}{0.32\textwidth} 
		\centering
		\includegraphics[width=\textwidth, interpolate]{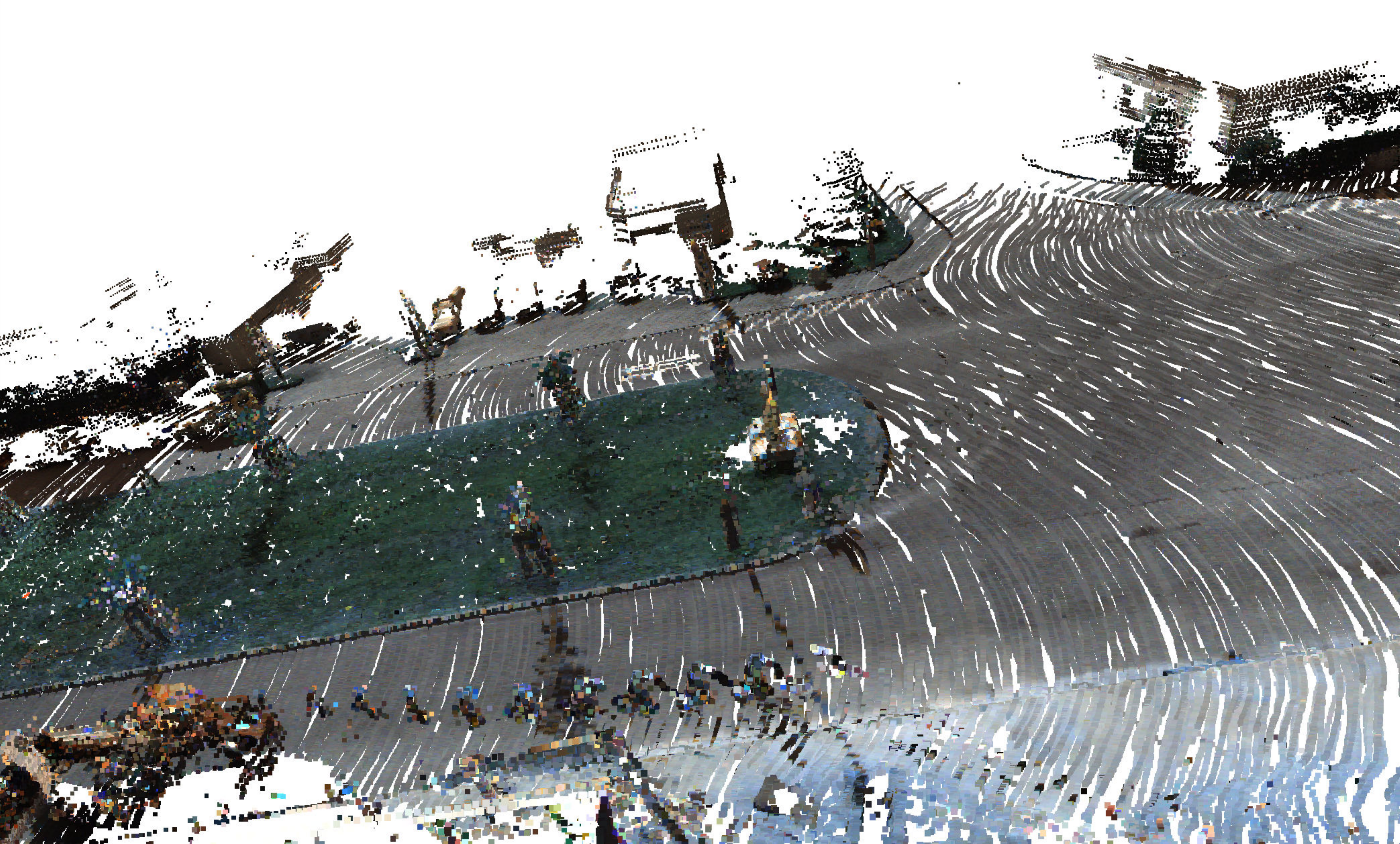}
		\caption{JVLDLoc}
	\end{subfigure}
	\caption{
	Magnified view of mapping result on KITTI 00, 05, 06 from top row to bottom row.
	Each column is the result of diffent methods.
	All above figures are the enlargement of the red box region of corresponding sequence in \cref{fig:lidarmapBEVbox}.
	}
	\label{fig:lidarmapzoom1}
\end{figure}

\begin{figure}[t]
	\centering
	\begin{subfigure}{0.48\textwidth} 
		\centering
		\includegraphics[width=\textwidth, interpolate]{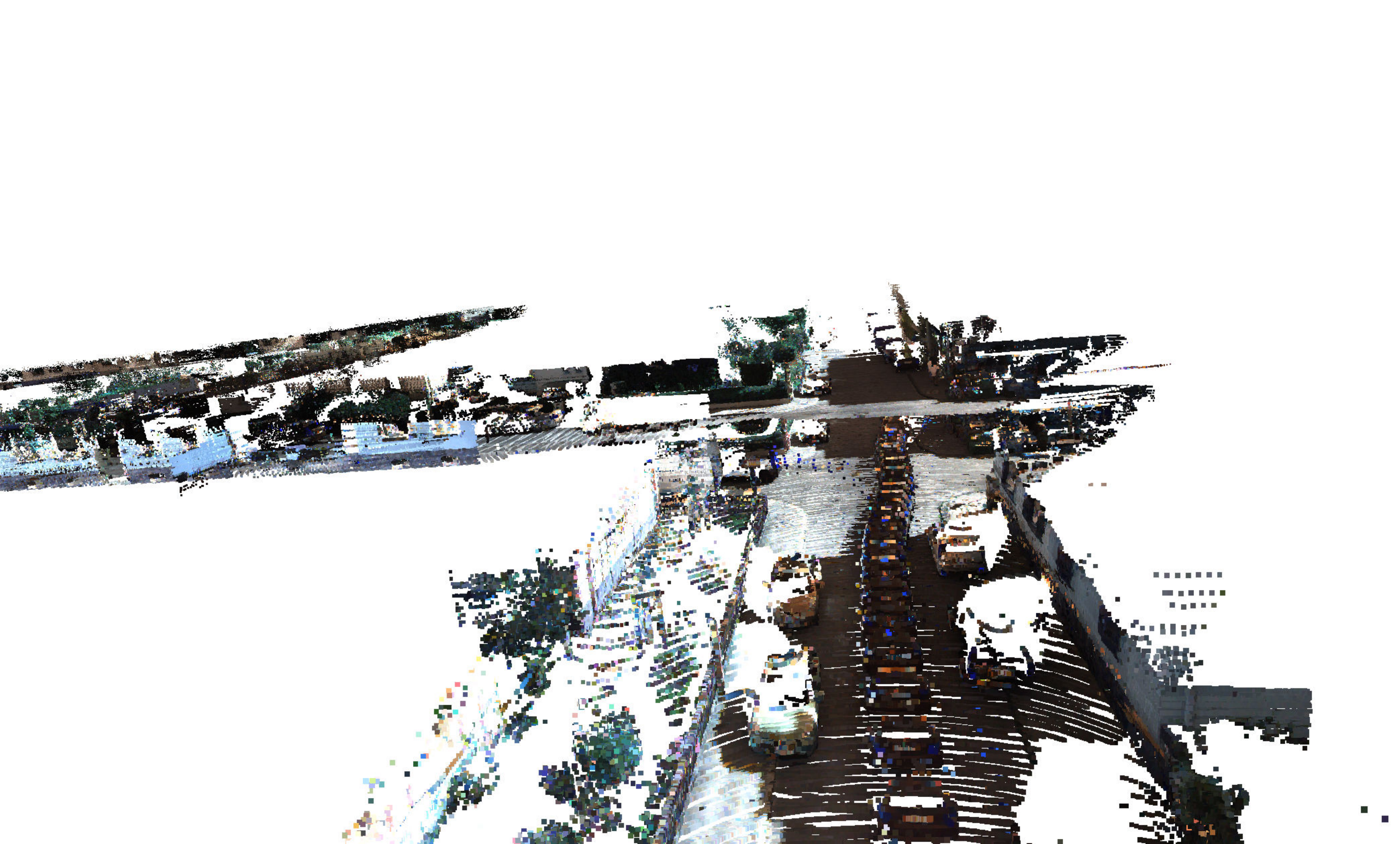}
		\caption{JVLDLoc w.o. direction}
	\end{subfigure}
	\begin{subfigure}{0.48\textwidth} 
		\centering
		\includegraphics[width=\textwidth, interpolate]{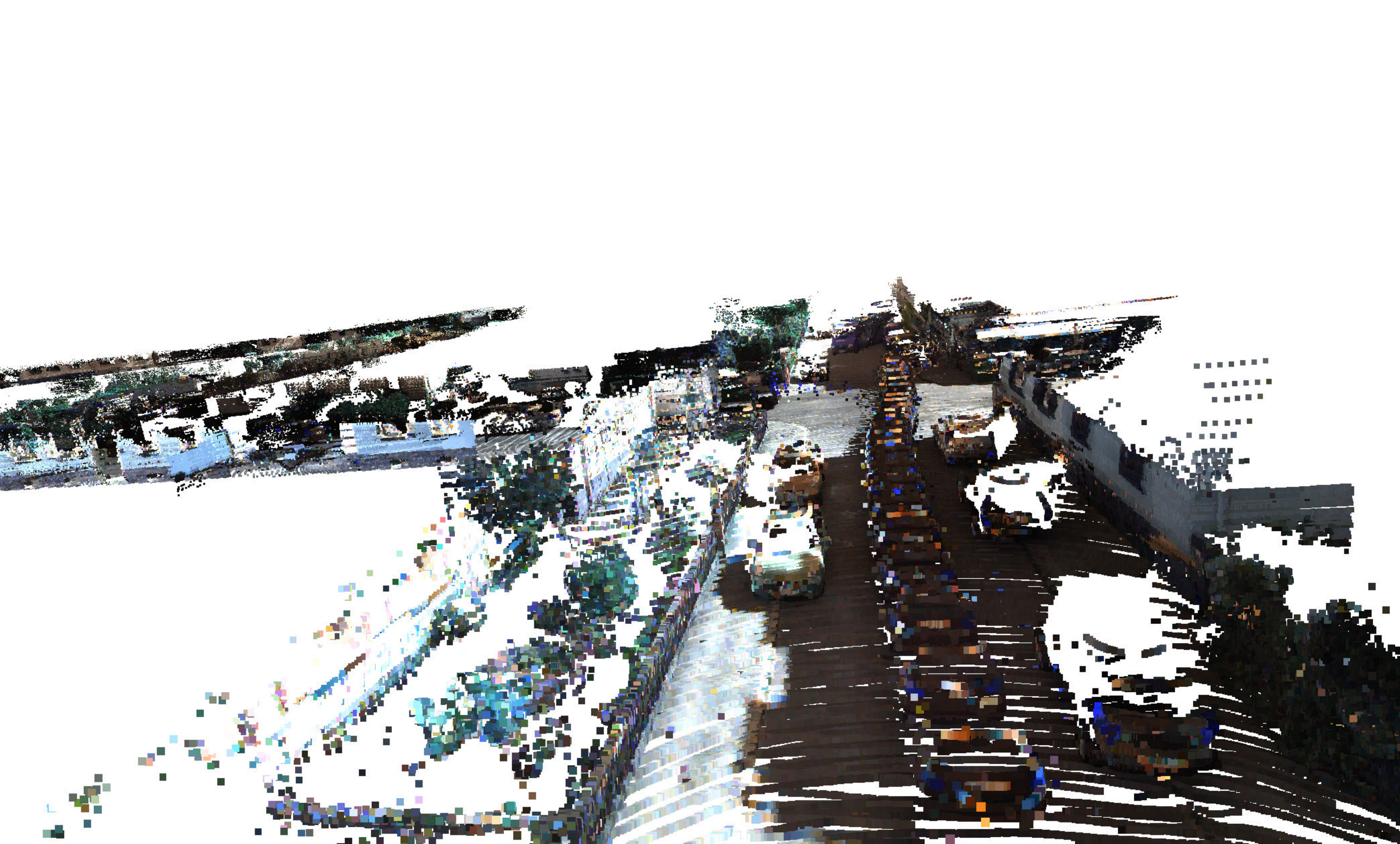}
		\caption{JVLDLoc}
	\end{subfigure}
	\caption{
		Magnified view of mapping result on KITTI 07.
		Each column is the result of diffent methods.
		All above figures are the enlargement of the red box region of corresponding sequence in \cref{fig:lidarmapBEVbox}.
	}
	\label{fig:lidarmapzoom2}
\end{figure}

\begin{figure}[t]
	\centering
	\begin{subfigure}{0.48\textwidth} 
		\centering
		\includegraphics[width=\textwidth, interpolate]{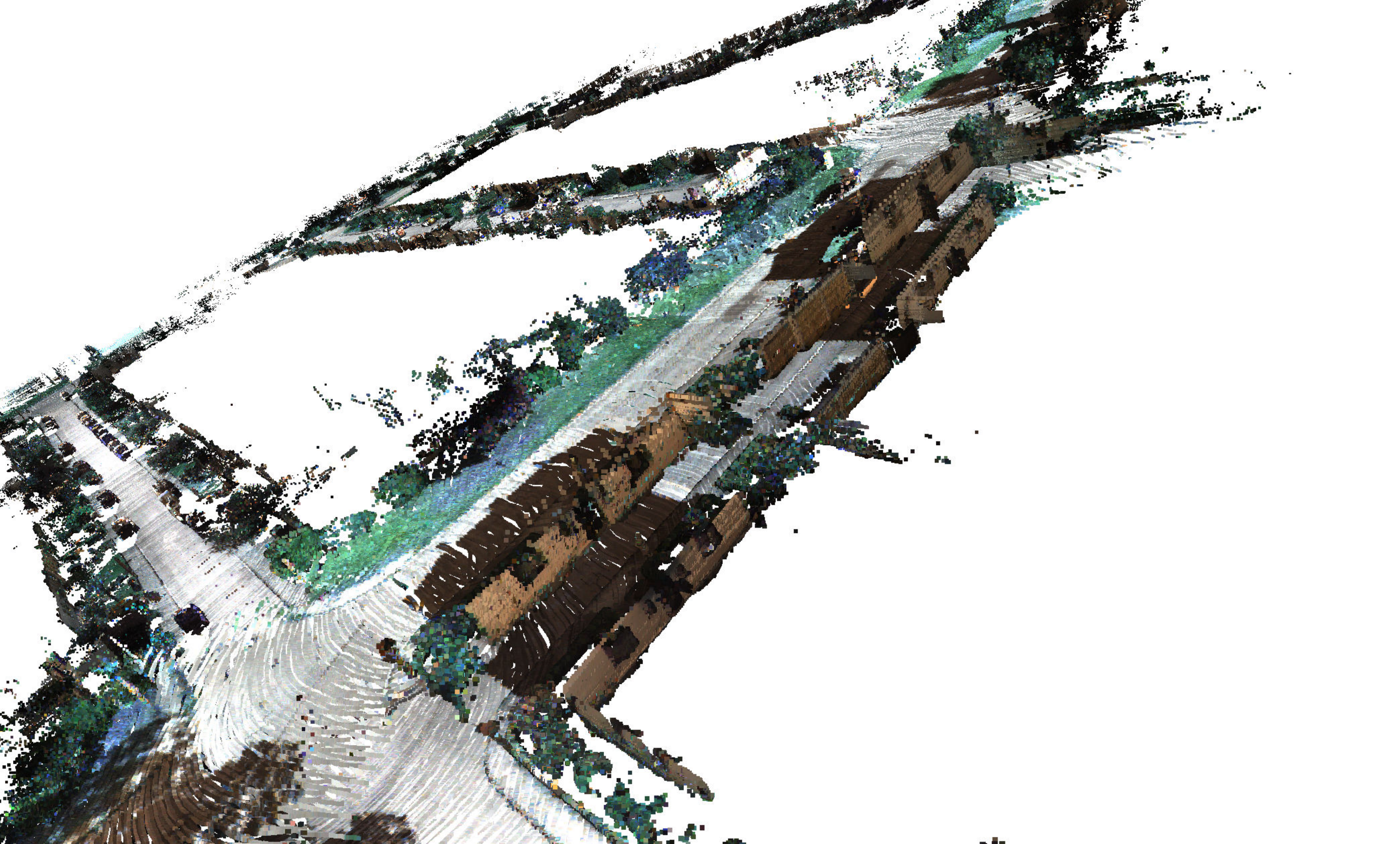}
		\caption{LiDAR odometry $\textbf{N}=1$}
	\end{subfigure}
	\begin{subfigure}{0.48\textwidth} 
		\centering
		\includegraphics[width=\textwidth, interpolate]{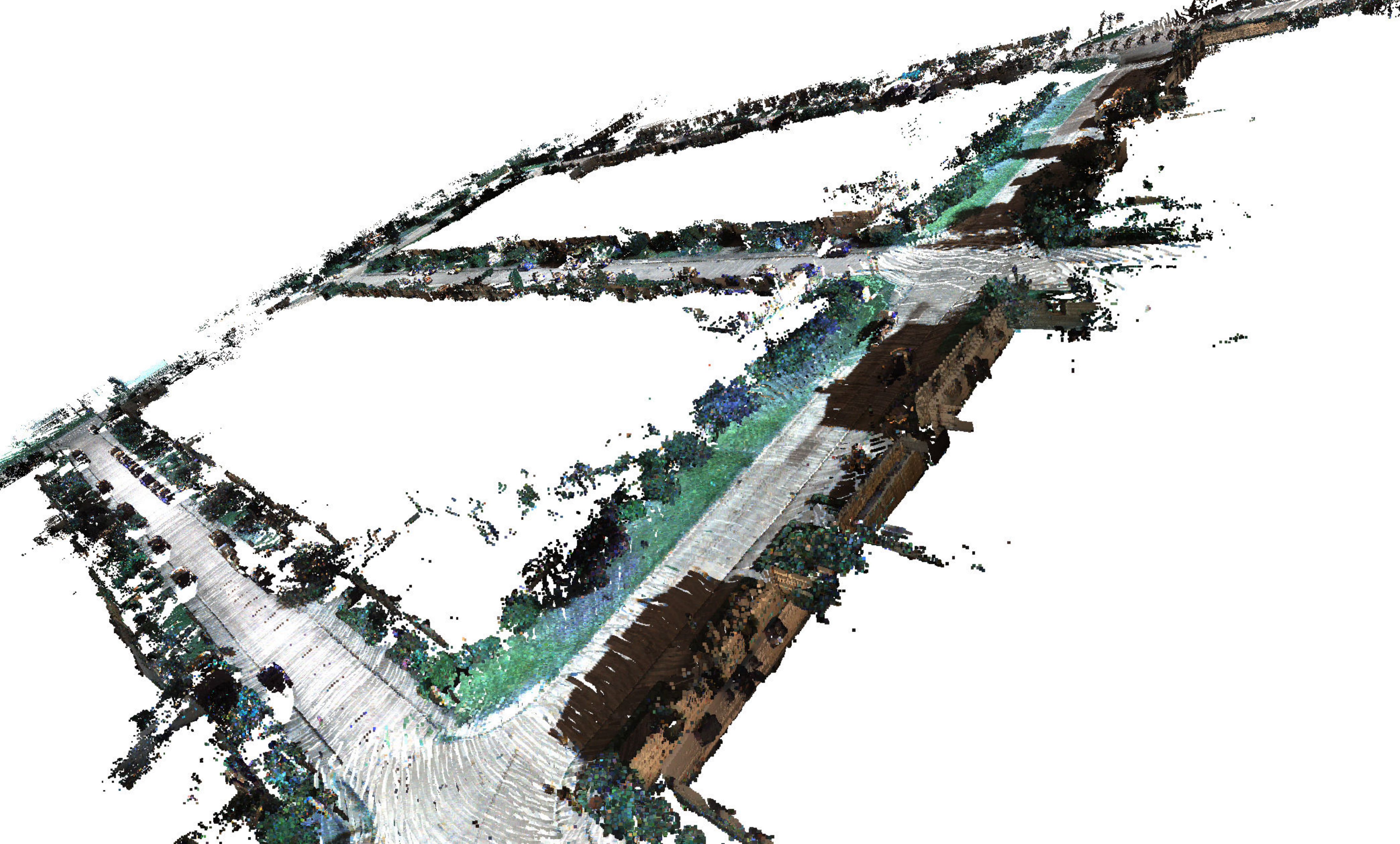}
		\caption{JVLDLoc w.o. direction}
	\end{subfigure}
	\begin{subfigure}{0.96\textwidth} 
		\centering
		\includegraphics[width=\textwidth, interpolate]{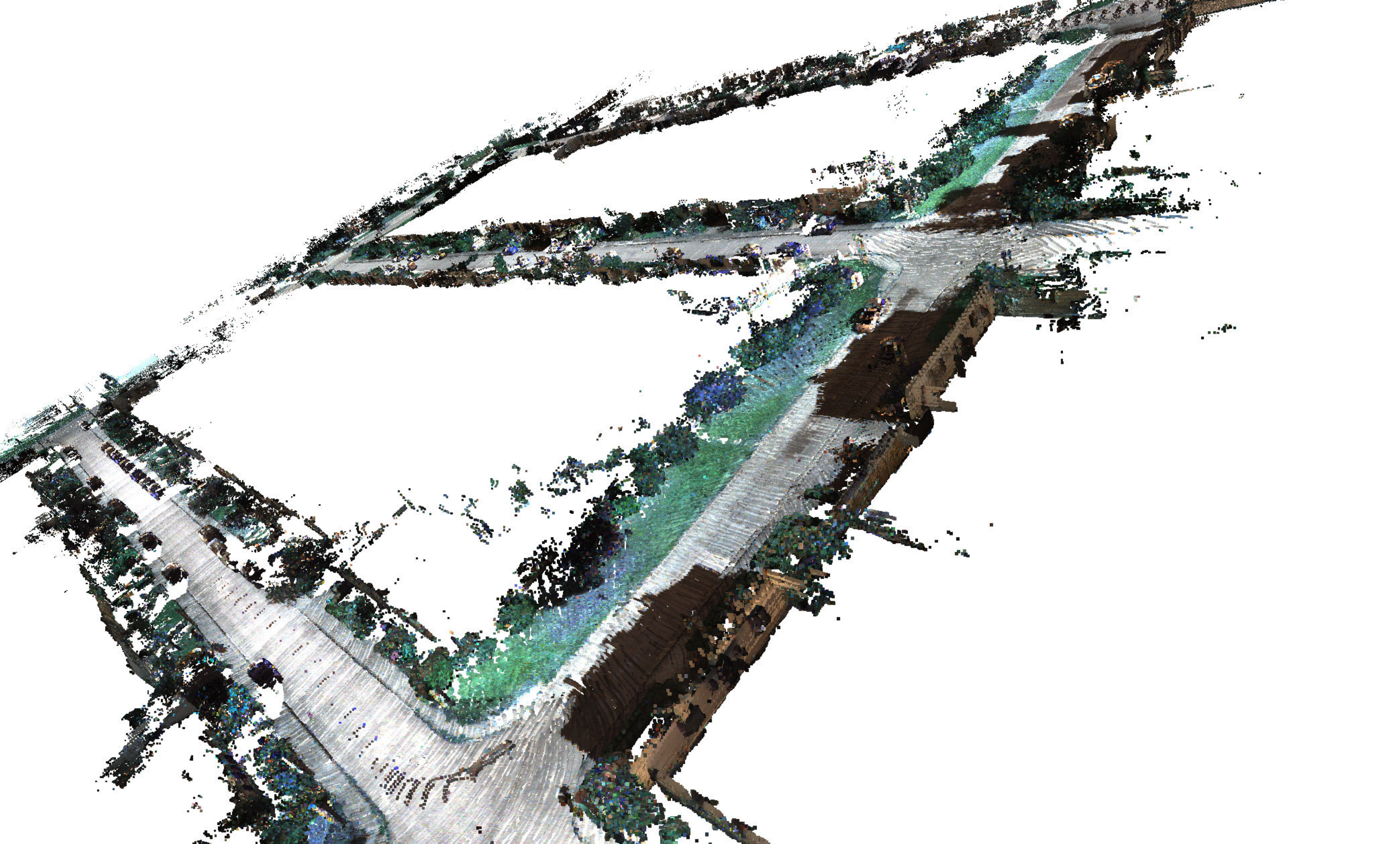}
		\caption{JVLDLoc}
	\end{subfigure}
	\caption{
		Magnified view of mapping result on KITTI 08 of diffent methods.
		All above figures are the enlargement of the red box region in \cref{fig:lidarmapBEVbox2}.
	}
	\label{fig:lidarmapzoom3}
\end{figure}

Here we present more mapping comparison on KITTI sequences in \cref{fig:lidarmapBEVbox}, \cref{fig:lidarmapBEVbox2}, \cref{fig:lidarmapzoom1}
, \cref{fig:lidarmapzoom2}, \cref{fig:lidarmapzoom3}.
These mapping results qualitatively prove that our method can indeed alleviate the global drift.

\subsectionBefore
\subsection{Effects of Direction Priors}
\subsectionAfter

Besides Fig. 6 in our submission,\cref{fig:trajcompwod1} presents another intuitive trajectory comparisons on KITTI 01.
Apparently, we can get less drift after long distance under direction priors.

\begin{table}[t] 
	\centering
	\caption{Comparison to other methods on KITTI 00-10.
	$t_{rel}$ and $r_{rel}$ are relative translational error (\%) and rotational error ($^{\circ}$/100m).
	(NA : Results are Not Available from paper).}	
	\label{tab:compwithsota}
		\resizebox{1.0\textwidth}{!}
		{
			\begin{tabular}{l||ccc|ccc|ccc|ccc|ccc|ccc}
				\toprule
				&  \multicolumn{3}{c|}{00}  &\multicolumn{3}{c|}{01}      & \multicolumn{3}{c|}{02} & \multicolumn{3}{c|}{03} &  \multicolumn{3}{c|}{04} & \multicolumn{3}{c}{05} \\ 
				
				\multirow{-2}{*}{\begin{tabular}[c]{@{}c@{}}Method \end{tabular}}
				&  $t_{rel}$  & $r_{rel}$ & $APE$  & $t_{rel}$ & $r_{rel}$ & $APE$  & $t_{rel}$ & $r_{rel}$   & $APE$   & $t_{rel}$ & $r_{rel}$ & $APE$   & $t_{rel}$  & $r_{rel}$ & $APE$  & $t_{rel}$ & $r_{rel}$ & $APE$ \\
				\midrule
				\hline
				\noalign{\smallskip}

				IMLS-SLAM \cite{deschaud2018imls}    
                &0.50 &0.18 &	3.90
                &0.82 &0.10 &	2.41
                &0.53 & 0.14 &7.16
                &0.68 &0.22 &	0.65
                &0.33 & 0.12 &0.18
                &0.32 &0.13 &	1.79
                \\ 
				LIMO-PL \cite{Huang2020LidarMonocularVO} 
				&0.99  &NA&	NA	  
				& 1.87 &NA&	NA
				& 1.38 &NA&	NA 
				& 0.65 &NA&	NA	  
				& 0.42 &NA&	NA
				& 0.72 &NA&	NA
				\\ 
				DVL-SLAM \cite{Shin2020DVLSLAMSD}    
				&0.93  &NA&	NA	  
				& 1.47 &NA&	NA
				& 1.11 &NA&	NA 
				& 0.92 &NA&	NA	  
				& 0.67 &NA&	NA
				& 0.82 &NA&	NA
				\\ 
				TVL-SLAM-No calib \cite{chou2021efficient}    
				&0.59  &NA&	0.84	  
				& 1.08 &NA	&	6.56
				& 0.74 & NA&	2.16 
				& 0.71 &NA&	0.75	  
				& 0.49 &NA&	0.18
				& 0.32 &  NA&	0.41
				\\ 
				TVL-SLAM-VL calib \cite{chou2021efficient}     
				&0.57  &NA&	0.88	  
				& 0.86 &NA	&	4.40
				& 0.67 & NA&	1.87 
				& 0.71 &NA&	0.74	  
				& 0.45 &NA&	0.22
				& 0.31 &  NA&	0.42
				\\ 
				LIO-SAM    \cite{shan2020lio} 
				&253  &31.0&	956	  
				& 2.94 &0.60	&	19.8
				& 268 & 87.6&	212
				& NA &NA&	NA	  
				& 0.92 &0.47&	0.30
				& 0.69 & 0.39&	1.81
				\\
				JVLDLoc input IMLS    
				&0.63  &0.15&	1.31	  
				& 0.88 &0.06	&	1.77 
				& 0.65 & 0.11&	2.62 
				& 0.70 &0.14&	0.53	  
				& 0.29 &0.10&	0.15
				& 0.36 &  0.09&	0.71
				\\ \bottomrule
				\toprule
				&  \multicolumn{3}{c|}{06}  &\multicolumn{3}{c|}{07}      & \multicolumn{3}{c|}{08} & \multicolumn{3}{c|}{09} &  \multicolumn{3}{c|}{10} & \multicolumn{3}{c}{AVG on 00-10} \\ 
				\multirow{-2}{*}{\begin{tabular}[c]{@{}c@{}}Method \end{tabular}}
				&  $t_{rel}$  & $r_{rel}$ & $APE$  & $t_{rel}$ & $r_{rel}$ & $APE$  & $t_{rel}$ & $r_{rel}$   & $APE$   & $t_{rel}$ & $r_{rel}$ & $APE$   & $t_{rel}$  & $r_{rel}$ & $APE$  & $t_{rel}$ & $r_{rel}$ & $APE$ \\
				\midrule
				\hline
				\noalign{\smallskip}
				IMLS-SLAM \cite{deschaud2018imls}    
                &0.33 & 0.08 & 0.46
                &0.33 & 0.14 &0.46
                &0.80 & 0.18 &2.43
                &0.55 & 0.12 &1.39
                &0.52 & 0.17 &0.77
                & 0.52 & 0.14 &1.96
                \\ 
				LIMO-PL \cite{Huang2020LidarMonocularVO} 
				& 0.61 &NA&	NA
				& 0.56 &NA&	NA
				& 1.27 &NA&	NA
				& 1.06 &NA&	NA
				& 0.83 &NA&	NA
				& 0.94 &NA&NA
				\\ 
				DVL-SLAM \cite{Shin2020DVLSLAMSD}    
				& 0.92 &NA&	NA
				& 1.26 &NA&	NA
				& 1.32&NA&	NA
				& 0.66 &NA&	NA
				& 0.70 &NA&	NA
				& 0.98 &NA&NA
				\\ 
				TVL-SLAM-No calib \cite{chou2021efficient}    
				& 0.32 &  NA&	0.32
				& 0.36 &  NA&	0.37
				& 0.88 &  NA &	2.52
				& 0.64 & NA&	1.32
				& 0.68 &  NA&	1.05
				& 0.62 & NA  &	1.50
				\\ 
				TVL-SLAM-VL calib \cite{chou2021efficient}     
				& 0.31&  NA&	0.28
				& 0.35 &  NA&	0.35
				& 0.86 &  NA &	2.14
				& 0.61 & NA&	1.18
				& 0.53 &  NA&	0.63
				& 0.57 & NA  &	1.19
				\\ 
				LIO-SAM    \cite{shan2020lio} 
				&0.59  &0.30&	0.73	  
				& 0.48 &0.29	&	0.48 
				& 83.4 & 32.4&	292
				& 1.12 &0.41&	4.76	  
				& 0.82 &0.33&	1.18
				& 1.08 &  0.40&	4.16
				\\
				JVLDLoc input IMLS    
				& 0.33 &  0.07&	0.30
				& 0.33 &  0.13&	0.33
				& 0.81 &  0.18 &	2.52
				&0.47 & 0.12&	1.38
				& 0.51 &  0.15&	0.78
				& 0.54 & 0.12   &	1.13
				\\ \bottomrule
			\end{tabular}
		}
\end{table}

\subsectionBefore
\subsection{Comparison to Other Methods on KITTI Odometry Dataset}
\subsectionAfter
As a complete to Table 3 in main paper, \cref{tab:compwithsota} shows comparison of localization results on full KITTI 00-10.

\begin{figure}[t]
	\centering
	\begin{subfigure}{0.3\textwidth}
		\centering
		\includegraphics[width=\textwidth, interpolate]{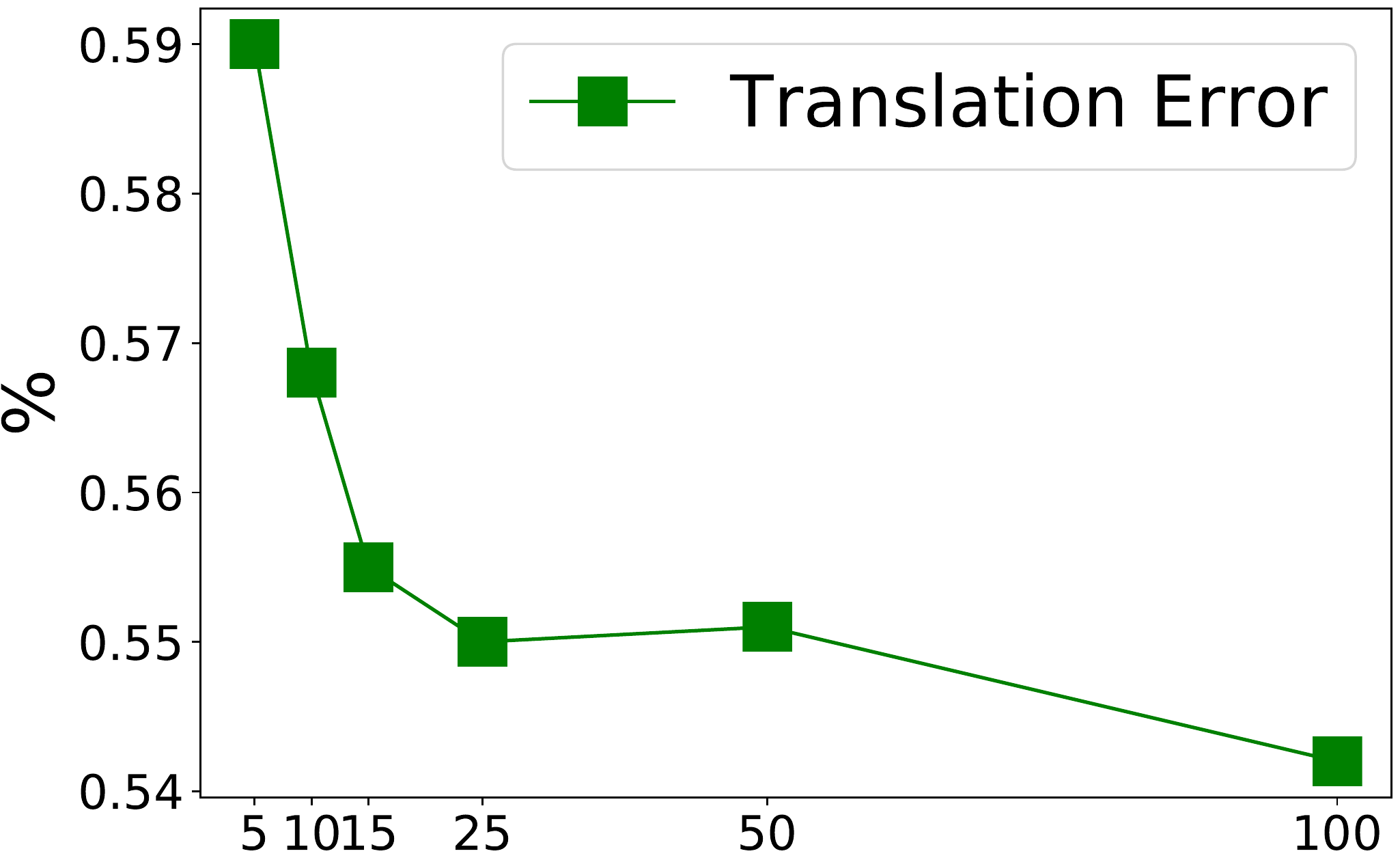}
	\end{subfigure}
	\begin{subfigure}{0.3\textwidth}
		\includegraphics[width=\textwidth, interpolate]{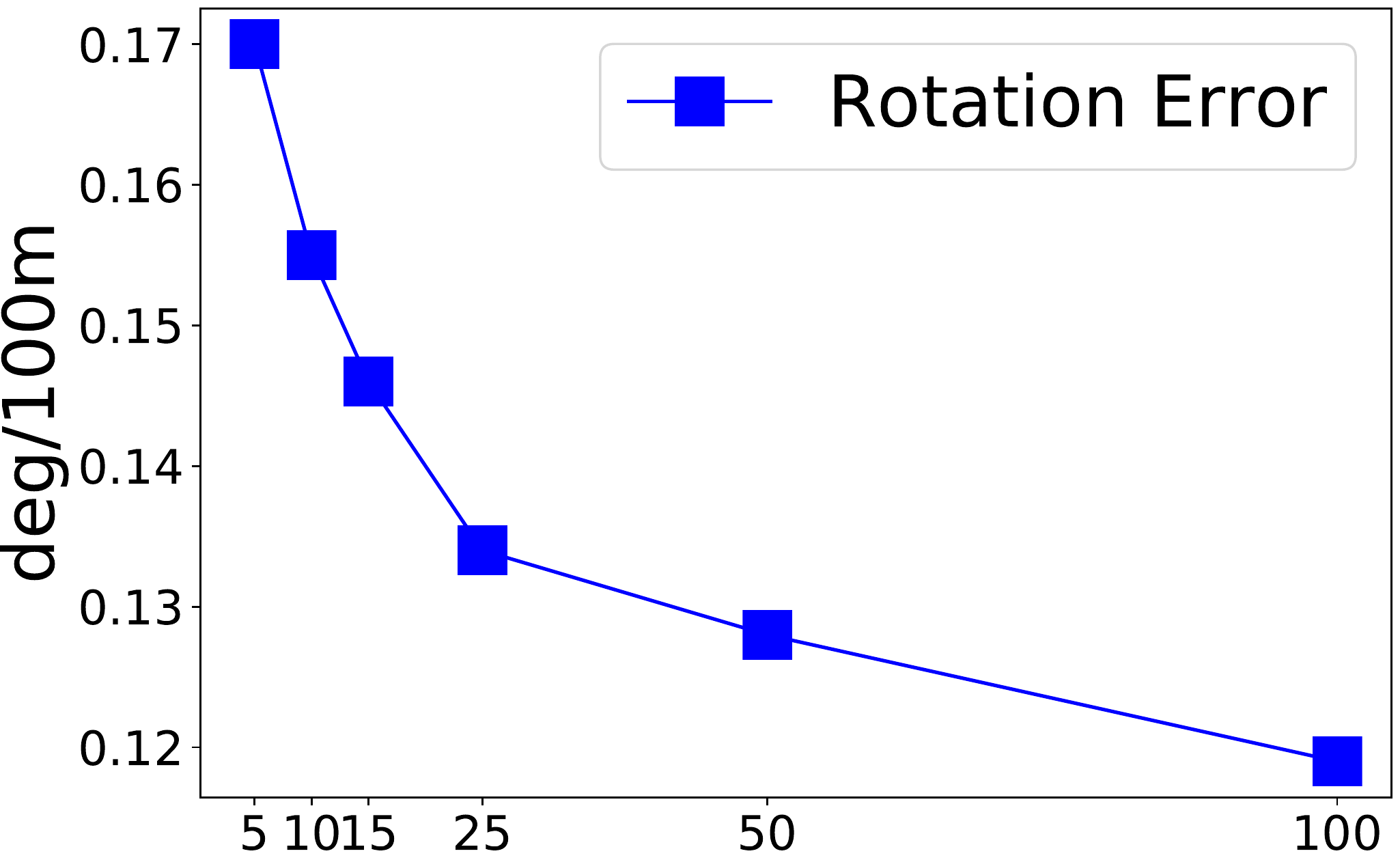}
	\end{subfigure}
	\begin{subfigure}{0.3\textwidth}
		\includegraphics[width=\textwidth, interpolate]{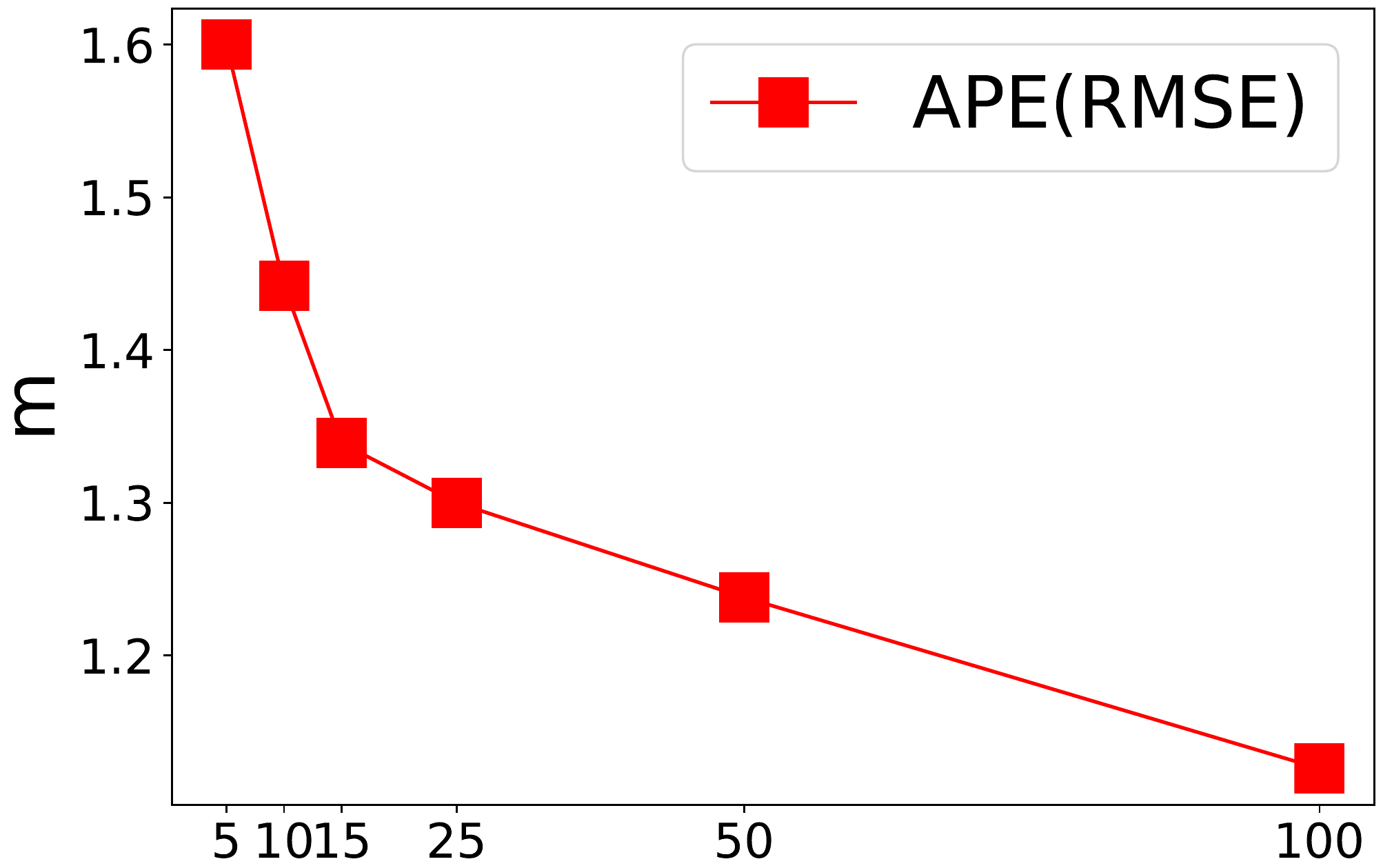}
	\end{subfigure}
	\caption{The average result under differnent scan map size of keyframe on KITTI 00-10.
	From left to right is relative translation error, rotation error and APE respectively.
	The x label is values of different $\textbf{N}$.}
	\label{fig:differentN}
\end{figure}

\subsectionBefore
\subsection{Ablation Study}
\subsectionAfter

\begin{figure}[t]
	\centering
	\begin{subfigure}{0.48\textwidth}
		\centering
		\includegraphics[width=\textwidth, interpolate]{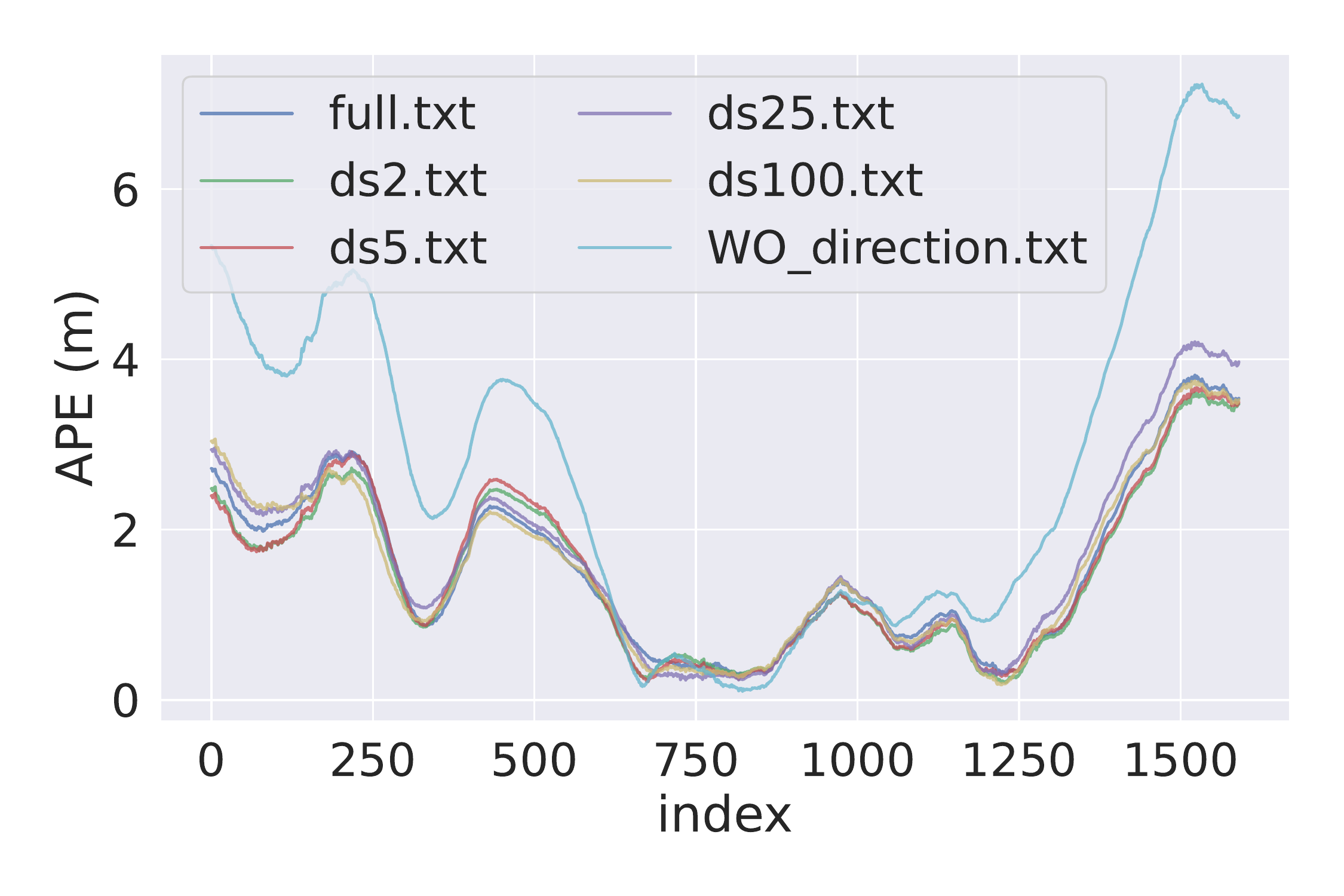}
	\end{subfigure}
	\begin{subfigure}{0.48\textwidth}
		\includegraphics[width=\textwidth, interpolate]{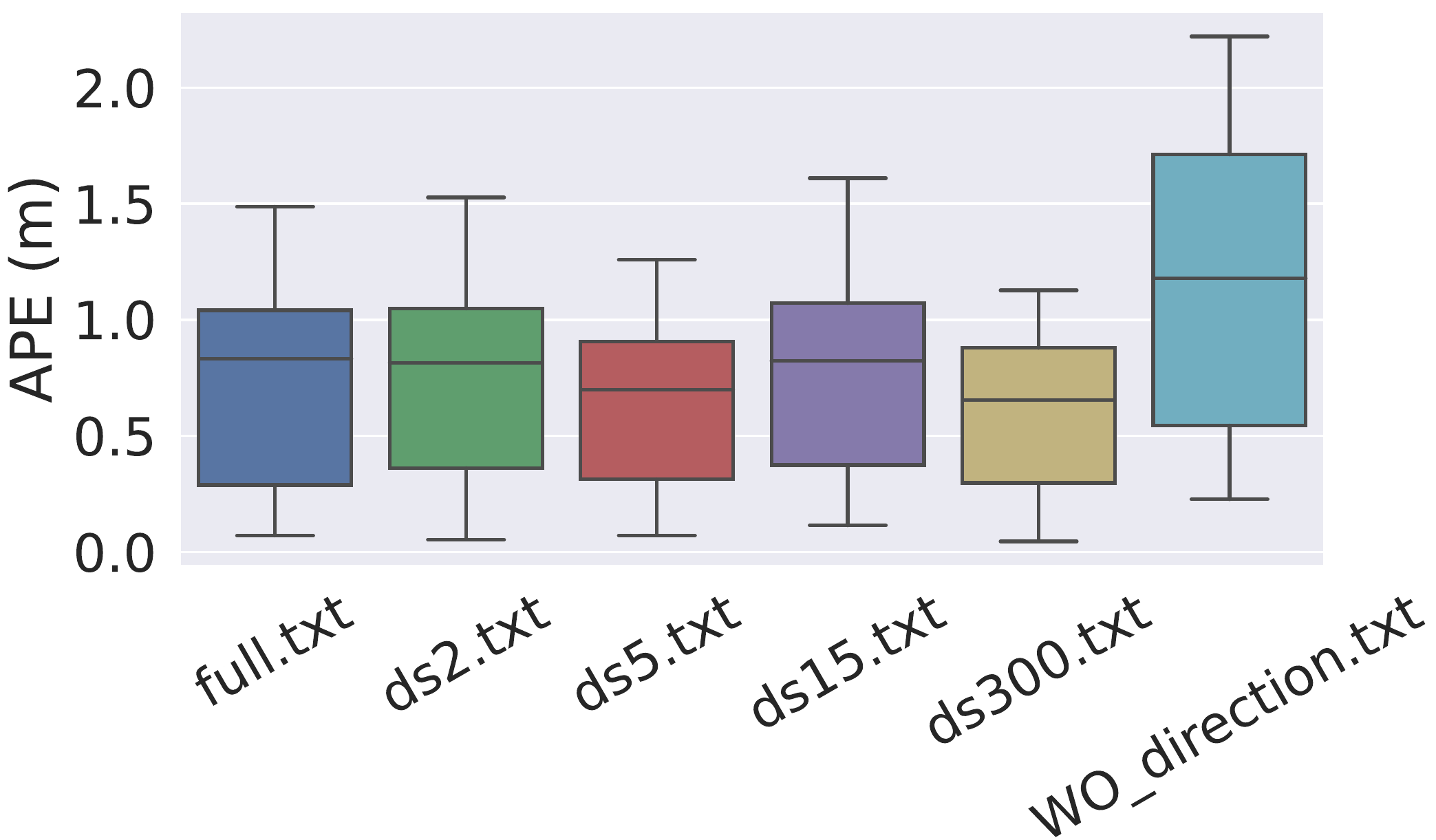}
	\end{subfigure} 
	\caption{
		left: Comparison of APE on KITTI 09.The x-coordinate is frame index. 
		right: Box plots of APE on KITTI 05. 
		$full$ refers to no downsampling, $WO direction$ refers to no direction priors.
		Other lines respectively corresponds to different $V_{interval}$.
		Take $ds2.txt$ as an example, it is the APE when $V_{interval}=2$.
	}
	\label{fig:aperawdsVP}
\end{figure}


\subsubsection{Different Number of Direction Priors}
%
For each KITTI sequence's relative rotation error and APE RMSE under differnent $V_{interval}$, see \cref{tab:changevpnumber}.
Note that $V_{interval}=0$ means no reduction to direction priors.
Besides Fig. 7 in main paper, \cref{fig:aperawdsVP} shows similiar results in line with the conclusion in our paper.  
In this experiment, the inpout prior is LiDAR odometry when $\textbf{N}=1$.

\begin{table}[t] 
	\caption{The comparison of different number of direction priors on KITTI 00-10.
	each value in table is $r_{rel}(^{\circ}/100m)/APE(RMSE)(m)$. }	 
	\label{tab:changevpnumber}
	\centering
		\resizebox{1.0\textwidth}{!}
		{
			\begin{tabular}{l||c|c|c|c|c|c}
				\toprule
				{$V_{interval}$}& {00} & {01} & {02} & {03} & {04} & {05} \\
				\midrule
				\hline 
				\noalign{\smallskip}
				
				0    
				&0.199/1.616
				&0.137/3.097
				&0.135/3.015
				&0.161/0.585
				&0.125/0.175
				&0.133/0.802
				\\
				5    
				&0.185/1.555
				&0.139/3.071
				&0.139/3.116
				&0.176/0.593
				&0.126/0.182
				&0.131/0.733
				\\
				10
				&0.189/1.659
				&0.143/3.242
				&0.140/3.164
				&0.174/0.590
				&0.127/0.182
				&0.139/0.798
				\\
				15
				&0.203/1.528	  
				&0.143/3.116 
				&0.150/3.073 
				&0.174/0.569	  
				&0.131/0.181
				&0.135/0.867
				\\ 
				25
				&0.215/1.790	  
				&0.135/2.953
				&0.144/3.387 
				&0.173/0.579	  
				&0.131/0.187
				&0.144/0.724
				\\
				50
				&0.215/1.678	  
				&0.138/3.099
				&0.141/2.987 
				&0.184/0.559	  
				&0.128/0.174
				&0.156/0.941
				\\ 
				100    
				&0.225/1.737	  
				&0.136/3.020 
				&0.141/2.730 
				&0.187/0.559	  
				&0.123/0.193
				&0.138/0.888
				\\ \bottomrule
				\noalign{\smallskip}
				{}& {06} & {07} & {08} & {09} & {10} & {AVG} \\
				\midrule
				\hline 
				\noalign{\smallskip}
				0    
				&0.071/0.302
				&0.202/0.521
				&0.230/3.131
				&0.217/1.828
				&0.250/0.933
				& 0.169/1.455
				\\
				5    
				&0.074/0.295
				&0.188/0.475
				&0.253/3.194
				&0.234/1.792
				&0.227/0.891
				& 0.170/1.445
				\\
				10
				&0.074/0.302
				&0.186/0.423
				&0.266/3.399
				&0.226/2.035
				&0.252/0.970
				& 0.174/1.524
				\\ 
				15 
				&0.076/0.307
				&0.178/0.478
				&0.277/3.571
				&0.225/1.990
				&0.242/0.938
				& 0.176/1.511
				\\ 
				25 
				&0.077/0.313
				&0.191/0.510
				&0.279/3.711
				&0.233/1.970
				&0.255/0.956
				&0.180/1.553
				\\
				50
				&0.074/0.310
				&0.175/0.437
				&0.265/3.298
				&0.229/1.748
				&0.256/0.961
				&0.178/1.472
				\\ 
				100    
				&0.074/0.311
				&0.199/0.553
				&0.273/3.412
				&0.224/1.820
				&0.219/0.901
				&0.176/1.466
				\\ \bottomrule
			\end{tabular}
		}
\end{table}

\subsubsectionBefore
\subsubsection{Different Scan Map Size of Keyframe}
This part we explore the influence of different scan map size $\textbf{N}$ of each keyframe
when it constructs point to IMLS surface error in local mapping.
In this experiment, the inpout prior is IMLS-SLAM\cite{deschaud2018imls}.
\cref{tab:changescanmap} presents relative rotation error and RMSE on each KITTI sequence. 
\cref{fig:differentN} also plots average relative translation error, rotation error and EMSE on KITTI 00-10. 
It is reasonable that the more scan kept in the local map, the bettrer the localization result are. 
Because each current point cloud will have longer and stronger constraints on geometric consistency to its pose.
However, when the size of local map is too large, the gain in accuracy won't be as noticeable when $\textbf{N}$ is small.

\begin{table}[t] 
	\centering
	\caption{The comparesion of different scan map size on KITTI 00-10.
	each value in table is $r_{rel}(^{\circ}/100m)/APE(RMSE)(m)$. }	
	\label{tab:changescanmap}
		\resizebox{1.0\textwidth}{!}
		{
			\begin{tabular}{l||c|c|c|c|c|c}
				\toprule
				{\textbf{N}}& {00} & {01} & {02} & {03} & {04} & {05} \\
				\midrule
				\hline 
				\noalign{\smallskip}

				5    
				&0.241/2.372
				&0.102/2.881
				&0.235/3.381
				&0.158/0.652
				&0.109/0.122
				&0.101/0.684
				\\
				10
				&0.177/1.462
				&0.069/2.187
				&0.182/3.119
				&0.174/0.582
				&0.121/0.143
				&0.103/0.866
				\\
				15
				&0.157/1.372	  
				&0.075/2.398 
				&0.156/2.957 
				&0.169/0.565	  
				&0.115/0.149
				&0.101/0.684
				\\ 
				25
				&0.150/1.305	  
				&0.064/2.034
				&0.134/2.845 
				&0.164/0.552	  
				&0.106/0.152
				&0.094/0.683
				\\
				50
				&0.147/1.295	  
				&0.062/1.835
				&0.120/2.771 
				&0.184/0.567	  
				&0.105/0.145
				&0.089/0.707
				\\ 
				100    
				&0.149/1.307	  
				&0.063/1.766 
				&0.111/2.622 
				&0.144/0.534	  
				&0.096/0.145
				&0.090/0.706
				\\ \bottomrule
				\noalign{\smallskip}
				{}& {06} & {07} & {08} & {09} & {10} & {AVG} \\
				\midrule
				\hline 
				\noalign{\smallskip}
				5    
				&0.075/0.318
				&0.222/0.462
				&0.256/3.989
				&0.141/1.780
				&0.230/0.959
				&0.165/1.591
				\\
				10
				&0.077/0.311
				&0.191/0.450
				&0.248/3.540
				&0.163/2.319
				&0.205/0.881
				& 0.153/1.434
				\\ 
				15 
				&0.079/0.342
				&0.163/0.323
				&0.218/3.196
				&0.141/1.780
				&0.230/0.959
				&0.146/1.343
				\\ 
				25  
				&0.079/0.305
				&0.141/0.425
				&0.212/3.155
				&0.133/1.922
				&0.192/0.920
				&0.134/1.300
				\\
				50
				&0.078/0.303
				&0.138/0.383
				&0.194/2.934
				&0.128/1.808
				&0.164/0.868
				&0.128/1.238
				\\ 
				100    
				&0.071/0.295
				&0.131/0.332
				&0.181/2.520
				&0.118/1.383
				&0.154/0.775
				&0.119/1.126
				\\ \bottomrule
			\end{tabular}
		}
\end{table}

\clearpage 
\bibliographystyle{splncs04} 
\bibliography{mybibliography}